%% file: GLKM.tex
\documentclass[lettersize,journal]{IEEEtran}
\usepackage{amsmath,amsfonts,amssymb}
\usepackage{algorithmic}
\usepackage{algorithm}
\usepackage{array}
\usepackage{booktabs}
\usepackage[caption=false,font=normalsize,labelfont=sf,textfont=sf]{subfig}
\usepackage{makecell}
\usepackage{multirow}
\usepackage{textcomp}
\usepackage{stfloats}
\usepackage{url}
\usepackage{verbatim}
\usepackage{graphicx}
\usepackage{cite}
\usepackage[colorlinks,linkcolor=red,anchorcolor=blue,citecolor=green,CJKbookmarks=True]{hyperref}
\newcolumntype{C}{>{\centering\arraybackslash}p}

\begin{document}

\title{Generative Latent Kernel Modeling for Blind Motion Deblurring}

\author{Chenhao Ding, Jiangtao Zhang, Zongsheng Yue, Hui Wang, Qian Zhao, and Deyu Meng
\IEEEcompsocitemizethanks{\IEEEcompsocthanksitem Chenhao Ding, Jiangtao Zhang, Zongsheng Yue, Hui Wang, Qian Zhao are with the School of Mathematics and Statistics, Xi'an Jiaotong University, Xi'an, China. Deyu Meng is with the School of Mathematics and Statistics, Xi'an Jiaotong University, Xi'an, China, and also with Pengcheng Laboratory, Shenzhen, China. E-mail: \{dch0319,zhangjt2021,huiwang2019\}@stu.xjtu.edu.cn, \{zsyue, timmy.zhaoqian, dymeng\}@xjtu.edu.cn.}
\thanks{C. Ding and J. Zhang contributed equally to this work. Q. Zhao is the corresponding author.}
}

\markboth{Journal of \LaTeX\ Class Files,~Vol.~14, No.~8, August~2021}%
{Shell \MakeLowercase{\textit{et al.}}: A Sample Article Using IEEEtran.cls for IEEE Journals}


\maketitle

\begin{abstract}
Deep prior-based approaches have demonstrated remarkable success in blind motion deblurring (BMD) recently. These methods, however, are often limited by the high non-convexity of the underlying optimization process in BMD, which leads to extreme sensitivity to the initial blur kernel. To address this issue, we propose a novel framework for BMD that leverages a deep generative model to encode the kernel prior and induce a better initialization for the blur kernel. Specifically, we pre-train a kernel generator based on a generative adversarial network (GAN) to aptly characterize the kernel's prior distribution, as well as a kernel initializer to provide a well-informed and high-quality starting point for kernel estimation. By combining these two components, we constrain the BMD solution within a compact latent kernel manifold, thus alleviating the aforementioned sensitivity for kernel initialization. Notably, the kernel generator and initializer are designed to be easily integrated with existing BMD methods in a plug-and-play manner, enhancing their overall performance. Furthermore, we extend our approach to tackle blind non-uniform motion deblurring without the need for additional priors, achieving state-of-the-art performance on challenging benchmark datasets. 
The source code is available at \url{https://github.com/dch0319/GLKM-Deblur}.
\end{abstract}


\section{Introduction}
\IEEEPARstart{M}{otion} blur is a common degradation in digital imaging. It occurs when either the camera or objects in the scene move during the exposure time, resulting in a streaked or smeared appearance in the captured image. This phenomenon significantly degrades image quality and visual perception. Blind motion deblurring (BMD) addresses this issue by recovering the latent sharp image from its blurry observation without pre-specifying the blurring process. The motion blur can be categorized as uniform when the scene depth is constant and the camera moves parallel to the image plane, or non-uniform when these conditions are not met. By assuming the motion blur is uniform and spatially invariant, the blurring process can be mathematically formulated as 
\begin{equation}
    \boldsymbol{y}= \boldsymbol{k} \otimes \boldsymbol{x}+\boldsymbol{n},
    \label{eq_model}
\end{equation}
where $\boldsymbol{y}$ is the blurry image, $\boldsymbol{x}$ the underlying sharp image, $\boldsymbol{k}$ the blur kernel, $\boldsymbol{n}$ usually additive white Gaussian noise (AWGN), and $\otimes$ denotes the 2D convolution operator.
Given the necessity to estimate both the latent sharp image and the blur simultaneously, BMD is a severely ill-posed problem, thereby attracting considerable research attention.

Over the past decades, plenty of methods have been explored within a Bayesian framework. Most of them employ the maximum a posteriori (MAP) technique \cite{krishnan2009fast, krishnan2011blind, tog2008_shan_high_quality, xu2013unnatural,pan2014deblurring, pan2016l_0, michaeli2014blind, pan2017deblurring}, focusing on finding the most probable latent image through optimization. In contrast, variational inference (VI) approaches \cite{fergus2006removing, tip2015_zhou_variational_dirichlet_deblur, book2017_bishop_review,yue2024deep} aim to approximate the complete posterior distribution of the desirable sharp image, characterizing the uncertainty in the estimation. The fundamental limitation of these approaches stem from the manually designed priors for both the latent sharp image \cite{krishnan2009fast, xu2013unnatural,pan2014deblurring, pan2016l_0, michaeli2014blind, pan2017deblurring, cho2009fast, chan1998total, yan2017image} and the blur kernel \cite{icassp1997_molina_dirichlet_deblur, tip2015_zhou_variational_dirichlet_deblur}, which often struggle to accurately characterize the underlying true distributions, particularly in complex real-world scenarios, thereby hindering their performance. Moreover, the optimization process usually exhibits a highly non-convex nature due to complicated priors' construction, entrapping in suboptimal local minima~\cite{levin2009understanding}.

Deep learning-based BMD approaches have dominated the research area due to the powerful fitting capability of deep neural networks (DNNs). These approaches can be broadly categorized into two primary paradigms: data-driven methods and model-driven methods.
Data-driven~\cite{nah2017deep, tao2018scale, kupyn2018deblurgan, kupyn2019deblurgan, zamir2021multi, zamir2022restormer} directly learn a mapping from blurry images to their latent sharp counterparts via end-to-end training on a large-scale dataset. Albeit their impressive performance across multiple benchmarks, these approaches often suffer from the overfitting issue, resulting in poor generalization on unseen data, particularly in scenarios with large and complex motion blur kernels. In contrast, model-driven methods have shown superior generalization capabilities, mainly focusing on the design of more effective priors for both the image and the blur kernel within an energy minimization framework. A typical example is deep image prior (DIP)~\cite{ulyanov2018deep}, which models the image prior as a DNN with random inputs for image restoration and has been successfully extended to BMD tasks~\cite{ren2020neural, huo2023blind, li2023self}. In recent years,  diffusion models~\cite{ho2020denoising, song2021scorebased}, known for their exceptional generative capabilities, have been introduced into BMD tasks~\cite{chung2023parallel, chihaoui2024blind} to model the image priors, leading to notable improvements in performance. However, these methods do not fully exploit the statistical properties of the blur kernels and ignore the sensitivity issue regarding the kernel initialization, which arises from the non-convex nature of BMD. 

To address the aforementioned challenges of model-driven approaches, we propose a new method to depict and initialize the blur kernel using Deep Generative Prior (DGP)~\cite{pan2021exploiting, chan2021glean, yang2021gan, wang2021towards}. Specifically for BMD, we train a kernel generator based on a generative adversarial network (GAN)~\cite{goodfellow2014generative} as kernel prior, leveraging the GAN's powerful capability to model complex kernel distributions. The learned generator effectively captures the blur kernel in a low-dimensional latent space, facilitating a more compact representation compared to the original kernel space. Subsequently, we train a mapping from the blurry image to the latent code of the kernel generator, acting as a kernel initializer for the BMD task. This initialization step simplifies the learning process, as the latent kernel space offers a more tractable domain for the model. To solve the BMD problem, we first acquire a coarse kernel initialization from the blurry image via the kernel initializer, and then jointly fine-tune both the image generator and the kernel DGP to refine the final outputs. 
As the kernel generator sufficiently fits the kernel distribution and the kernel initializer provides an improved initial estimation, our method significantly enhances BMD performance, accelerating convergence, particularly for large blur kernels. 

This paper is an extension of our previous work~\cite{zhang2024blind}. In the previous work, we only explored the proposed generative latent kernel modeling technique within the BMD framework employing DIP as image prior. This present work extends this approach to BMD frameworks incorporating more general image priors, including the DIP~\cite{ren2020neural}, variational DIP (VDIP)~\cite{huo2023blind}, and diffusion model prior~\cite{chihaoui2024blind,chung2023parallel}. Furthermore, we advance the generative latent kernel modeling technique to deal with non-uniform deblurring scenarios.

The main contributions of this work are summarized as follows:
\begin{itemize}
    \item We construct a GAN-based blur kernel generator to better characterize the kernel structures. This generator compresses the blur kernel onto a more compact latent space and could be used as an effective kernel prior for BMD.
    \item We propose to learn a kernel initializer that maps from the blurry image to the latent code of the corresponding kernel. Attributed to the compactness of the latent kernel space, the proposed initializer can provide a more accurate kernel initialization for the subsequent BMD process.	
    \item By combining the designed kernel prior and initializer, our method serves as a plug-and-play generative latent kernel prior that can enhance the performance of various BMD approaches utilizing different image priors.
    \item Our method naturally extends to blind non-uniform motion deblurring without requiring additional priors, achieving state-of-the-art (SotA) performance on the challenging benchmark dataset \cite{lai2016comparative}.
\end{itemize}


\section{Related Works}
In this section, we review the related works from two aspects: (A) blind motion deblurring methods, which provide the context of our research problem, and (B) deep prior techniques for image processing, which form the foundation of our proposed approach.

\subsection{Blind Motion Deblurring}
BMD methods can be broadly categorized into two main approaches: optimization-based methods and deep learning-based methods.

\subsubsection{Optimization-based BMD methods}
Due to the inherently ill-posed nature of BMD, appropriate constraints on both the clean image and blur kernel solution spaces are essential for obtaining viable solutions. Traditional optimization-based methods have extensively focused on developing effective priors for this purpose. For clean images, various priors have been proposed, including total variation \cite{chan1998total}, hyper-Laplacian prior \cite{krishnan2009fast}, $l_1/l_2$-norm \cite{krishnan2011blind}, and transform-specific $l_1$-norm \cite{cai2009blind, xu2013unnatural} in the gradient domain, while patch-based priors \cite{michaeli2014blind, sun2013edge}, low-rank prior \cite{ren2016image}, and dark/bright channel prior \cite{pan2017deblurring, yan2017image} have been applied directly in the image domain. For blur kernels, beyond the fundamental non-negative and normalization constraints, researchers have introduced sparsity prior \cite{pan2017deblurring} and spectral prior \cite{liu2014blind} to better characterize kernel properties. Additionally, algorithmic refinements such as delayed normalization \cite{perrone2014total} and multi-scale implementation \cite{zuo2016learning} have been developed to enhance kernel estimation. Nevertheless, these approaches predominantly rely on handcrafted priors that often fail to accurately capture the intrinsic statistical properties of natural images and blur kernels, limiting their competitiveness in the current deep learning era.

\subsubsection{Deep learning-based BMD methods}
Leveraging their remarkable success across diverse domains, deep learning techniques have been increasingly adopted for the BMD task. Early approaches strategically integrated DNNs into traditional optimization frameworks, capitalizing on their representational flexibility. For instance, several researchers proposed neural network-based kernel predictors \cite{sun2015learning, chakrabarti2016neural, gong2017motion, yue2024deep} to enhance the accuracy of blur estimation. With the advancement of computational resources, the paradigm shifted toward end-to-end supervised learning, where sophisticated neural architectures directly map blurred images to their sharp counterparts using extensive paired training data \cite{nah2017deep, kupyn2018deblurgan, kupyn2019deblurgan, pan2020physics, cai2020dark, cho2021rethinking, zamir2021multi, chen2022simple,zamir2022restormer, kong2023efficient, pan2023cascaded, tang2024residual,pan2025learning,tang2025degradation}. While these approaches have achieved state-of-the-art performance on benchmarks such as GoPro \cite{nah2017deep} and RealBlur \cite{rim2020real}, they often struggle with generalization to images containing large complex blur kernels that fall outside the distribution of the pre-defined training sets.

Recently, a novel category of deep learning approaches for BMD that leverages deep priors (which will be elaborated in the next subsection) has attracted increasing attention. In this domain, Ren \MakeLowercase{\textit{et al.}} \cite{ren2020neural} pioneered the SelfDeblur method, which innovatively employed DIP to parameterize both the latent sharp image and blur kernel. Following this seminal work, numerous variants and extensions have been proposed, predominantly focusing on enhancing the image prior to improve performance \cite{huo2023blind,li2023self,tang2023uncertainty}. These methods achieved promising results in certain cases, notably surpassing fully supervised deep learning approaches on the challenging benchmark by Lai \MakeLowercase{\textit{et al.}} \cite{lai2016comparative}. Nevertheless, the intrinsic properties of blur kernels were not thoroughly explored, resulting in performance instability when dealing with large kernels due to the inherent non-convexity of the optimization problem. 
Alternatively, another direction leverages pre-trained DGPs for both the image and kernel. Asim \textit{et al.} \cite{asim2020blind} pioneered this by fine-tuning pre-trained generators, while Chung \textit{et al.} \cite{chung2023parallel} employed powerful pre-trained diffusion models as DGPs to perform simultaneous sampling of both components.

\subsection{Deep Prior for Image Processing}
In recent years, the flexibility of DNNs has enabled extensive application in characterizing image priors. Two predominant categories of such deep priors have emerged: Deep Image Prior (DIP) and Deep Generative Prior (DGP), which we briefly review below.

\subsubsection{Deep image prior} 
DIP was originally proposed by Ulyanov \MakeLowercase{\textit{et al.}} \cite{ulyanov2018deep} who demonstrated that a deep neural network could effectively approximate a target image (maybe up to a transformation) using random noise as input. The inherent complexity of the network architecture and its associated operations enables the DNN to accurately characterize the manifold of natural images, thereby functioning analogously to traditional priors or regularizers in image processing frameworks. Since its inception, DIP has garnered significant attention and has been successfully applied to various image processing tasks (beyond the aforementioned BMD tasks), including natural image denoising \cite{ulyanov2018deep}, super-resolution \cite{liang2021flow,yue2022blind}, inpainting \cite{ulyanov2018deep}, image decomposition \cite{gandelsman2019double}, low-light enhancement \cite{zhao2021retinexdip}, PET image reconstruction \cite{gong2018pet} and hyperspectral image denoising \cite{luo2021hyperspectral}.

\subsubsection{Deep generative prior} 
Attributed to their powerful generation capabilities, pre-trained deep generators, such as GANs~\cite{goodfellow2014generative}, effectively approximate the distributions of natural images, thereby providing statistically reasonable
priors. These priors, referred to as DGPs~\cite{pan2021exploiting}, offer a sophisticated alternative to traditional handcrafted regularizers. Following a comprehensive pre-training phase, DGP can be employed as an estimator to approximate target images via fine-tuning, similarly to DIP. Fine-tuning a DGP is typically achieved via two principal strategies. 
The first strategy maintains fixed generator parameters and optimizes only the latent input: Menon \MakeLowercase{\textit{et al.}} \cite{menon2020pulse} demonstrate this approach for photo upsampling, and Chihaoui \MakeLowercase{\textit{et al.}} \cite{chihaoui2024blind,chihaoui2025diffusionimageprior}, Murata \MakeLowercase{\textit{et al.}} \cite{murata2023gibbsddrm} apply it to blind image restoration; notably, this paradigm is closely related to GAN inversion \cite{xia2022gan}. The second strategy jointly fine-tunes both the latent code and the pre-trained generator's weights \cite{pan2021exploiting,wang2025navigating,shah2025join}, which can more precisely identify the optimal estimate within the model's learned image manifold. 

In addition to directly modeling the images, DGP has been effectively extended to characterize various degradation operators, particularly blur kernels. For example, Asim \MakeLowercase{\textit{et al.}} \cite{asim2020blind} pioneered this approach by pre-training a variational auto-encoder (VAE) \cite{kingma2013auto} as a DGP for motion blur kernels in BMD tasks. Subsequently, Liang \MakeLowercase{\textit{et al.}} \cite{liang2021flow} employed normalizing flow (NF) \cite{dinh2014nice, dinh2017density, kingma2018glow} as a DGP for Gaussian blur kernels in blind super-resolution applications. More recently, Chung \MakeLowercase{\textit{et al.}} \cite{chung2023parallel} leveraged pre-trained diffusion models as DGPs for both Gaussian and motion blur kernels. These seminal works highlight the effectiveness of DGP for blur kernels and partially inspire our work.

While the aforementioned deep priors have shown promising results, they still face challenges in effectively characterizing motion blur kernels. Our work addresses this gap by introducing a GAN-based kernel generator that better captures the statistical properties of blur kernels, combined with a novel kernel initializer that operates in the latent space. Unlike previous approaches that either rely on untrained networks or require manual regularization, our method provides a more stable and accurate initialization for the BMD process. Furthermore, our approach serves as a plug-and-play kernel prior that can enhance various BMD frameworks utilizing different image priors, demonstrating its versatility and effectiveness across diverse deblurring scenarios.

\section{Preliminaries}
In this section, we analyze the methodological foundations and inherent limitations of existing model-driven approaches for BMD, particularly those utilizing the DIP prior.

The seminal work~\cite{ren2020neural} pioneered a dual-network architecture where the latent sharp image $\boldsymbol{x}$ is parameterized as an encoder-decoder convolutional network, while the blur kernel $\boldsymbol{k}$ is estimated via a single-hidden-layer fully-connected network. The corresponding optimization problem is as follows:
\begin{equation}
    \underset{\theta_k,\theta_x}{\min}\left\|G_k(\boldsymbol{z}_k;\theta_k)\otimes G_x(\boldsymbol{z}_x;\theta_x)-\boldsymbol{y}\right\|^2,
    \label{eq_dip_bid_basic}
\end{equation}
where $G_k(\cdot;\theta_k)$ and $G_x(\cdot;\theta_x)$ denote kernel and image generation networks, respectively, driven by random noise inputs $\boldsymbol{z}_k$ and $\boldsymbol{z}_x$. Notably, explicit physical constraints on $\boldsymbol{k}$ and $\boldsymbol{x}$ are omitted under the assumption that proper network architectures could implicitly enforce these constraints through their inductive biases. Consequently, the problem of recovering $\boldsymbol{k}$ and $\boldsymbol{x}$ is reformulated as an optimization over the network parameters $\theta_k^{\ast}$ and $\theta_x^{\ast}$.

\begin{figure}[t]
    \begin{minipage}{1\columnwidth}
        \centering
        \includegraphics[width=1\textwidth]{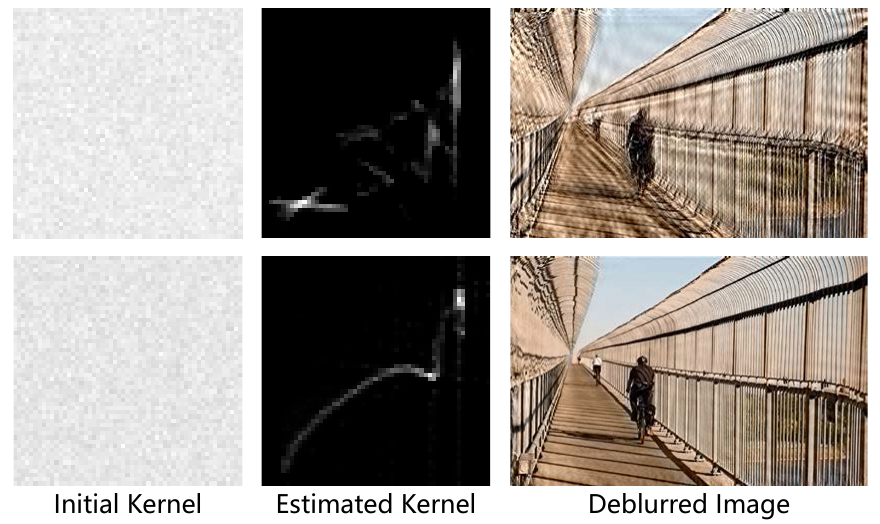}
    \end{minipage}
    \begin{minipage}{1\columnwidth}
        \centering
        \caption{Illustration of the initialization effect of the DIP-based BMD. The two rows correspond to two independent runs of SelfDeblur \cite{ren2020neural}. From left to right: the randomly initialized kernel, the finally estimated kernel, and the deblurred image.}\label{fig_intro}
    \end{minipage}
\end{figure}

Benefiting from the large capacity of DNNs, this framework demonstrated competitive performance on existing deblurring benchmarks~\cite{lai2016comparative}. Building upon this foundation, Huo \MakeLowercase{\textit{et al.}}\cite{huo2023blind} further integrated DIP into a variational Bayesian inference framework, introducing probabilistic modeling to improve optimization stability. Nevertheless, these DIP-based approaches did not sufficiently consider blur kernels' statistical properties or prior distributions, thereby constraining their overall efficacy.

A more critical issue lies in the non-convexity during optimization. Specifically, the random initialization of the latent variable $\boldsymbol{z}_k$ and the network parameters $\theta_k^{\ast}$ often results in unstable solutions, particularly for large blur kernels. As illustrated in Fig.~\ref{fig_intro}, two different random initializations, though drawn from the same distribution, can yield drastically different kernel estimations, leading to substantially divergent deblurring results. This instability becomes more problematic when extending DIP-based methods to non-uniform motion blur scenarios, where blur is assumed to vary across different image patches. In such cases, the estimated blur kernels for different image regions can exhibit pixel shifts, causing misalignments between adjacent patches. These misalignments give rise to visible artifacts along patch boundaries in the deblurred image.

These identified limitations motivate the pursuit of more advanced blur kernel modeling techniques. In this study, we investigate the statistical structures of blur kernels through DGP and develop a kernel initializer that operates within the latent space. This strategy provides more accurate and stable initializations and thus improves the optimization process. By addressing key challenges related to kernel representation and initialization, our approach aims to enhance both the performance and stability of deep prior-based BMD methods.

\section{Proposed Method: Overview and Pre-training}\label{sec:method}
\begin{figure*}[t]
    \centering
    \includegraphics[width=0.9\textwidth]{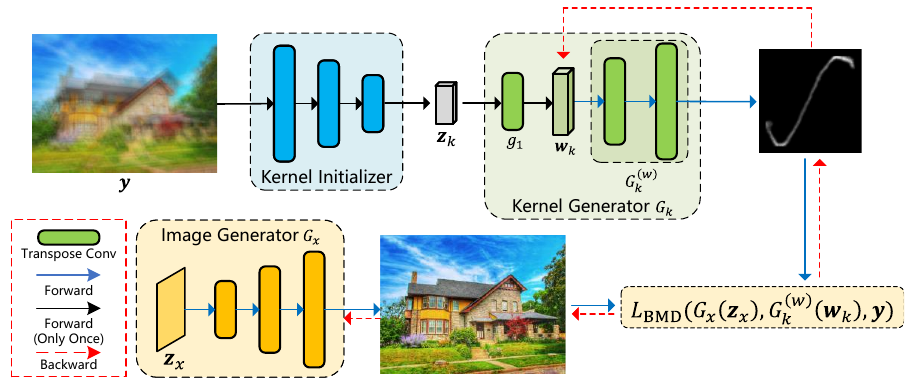}
    \caption{Overview of the Proposed BMD Framework. The framework incorporates a pre-trained kernel generator and initializer, where the image generator $G_x$ can accommodate various generative priors (detailed discussion in Sec.~\ref{sec:bmd_process}).}
    \label{fig_framework}
\end{figure*}

\subsection{Overview of the Proposed Method}
The proposed method consists of two stages. The first stage aims to pre-train a \emph{kernel generator} and a \emph{kernel initializer}, while the second stage solves the BMD problem with the learned generator and initializer.

In the first stage, we first train a kernel generator $G_k(\cdot;\theta_k^{\ast})$ using GAN~\cite{goodfellow2014generative}, where $\theta_k^{\ast}$ represents the optimized weight parameter. This generator builds up a map between blur kernels and random latent vectors, and thus can be plugged into the BMD problem as a DGP for the blur kernel.
Subsequently, by fixing this kernel generator, we train an encoder $E(\cdot;\theta_E^{\ast})$ that transforms a blurry image $\boldsymbol{y}$ into its corresponding kernel latent code through GAN-inversion techniques \cite{xia2022gan}. This encoder serves as an effective kernel initializer and facilitates a precise kernel prediction in the latent space. 
Based on this kernel initialization, we optimize both the image generator and the kernel latent code. Note that the image generator can be instantiated using various generative models, including DIP, diffusion models, and other deep generative architectures, demonstrating the versatility of the proposed kernel optimization strategy. The overall framework of our method is illustrated in Fig. \ref{fig_framework}.

In Sec. \ref{sec:kg_learning} and Sec. \ref{sec:ki_learning}, we provide details for learning the kernel generator and kernel initializer, respectively. Then in Sec. \ref{sec:bmd_process}, we discuss how to apply the pre-trained generator and initializer to the BMD process.

\subsection{Pre-training the Kernel Generator}\label{sec:kg_learning}
To pre-train the kernel generator, we first synthesize a substantial collection of motion blur kernels according to the physical generation mechanisms proposed in \cite{kupyn2018deblurgan} or \cite{lai2016comparative}. The kernel generator $G_k(\cdot;\theta_k)$, consisting of multiple convolutional layers, is then trained on these synthesized blur kernels following DCGAN~\cite{radford2015unsupervised}. Consistent with the DCGAN framework, the pre-training employs the Binary Cross-Entropy loss. Attributed to the powerful fitting capabilities of GAN, this straightforward kernel generator effectively captures the statistical structures of blur kernels, thus serving as a kernel DGP in the subsequent BMD task. Notably, the learned kernel generator provides a crucial low-dimensional latent space representation, facilitating the kernel optimization in BMD, as elaborated in Sec. \ref{sec:ki_learning}. 

\noindent{\textbf{Remark}}. Several prior studies have investigated deep generative priors for blur kernels using various generative models, including VAE~\cite {asim2020blind}, normalizing flow~\cite{liang2021flow}, and diffusion models~\cite{chung2023parallel}. 
In this work, we propose to use a GAN-based architecture, mainly motivated by three key technical considerations. First, in contrast to the vanilla VAE, which favors blurry outputs \cite{bredell2023explicitly}, GANs are known to generate samples with superior sharpness. 
Second, unlike normalizing flows, which require strict invertibility between the latent code and the blur kernel, GANs enable a more efficient kernel representation in a lower-dimensional latent space. This feature is particularly beneficial for modeling motion blur kernels with large dimensions, such as those reaching $75\times75$ pixels in the Lai dataset \cite{lai2016comparative}. 
Third, while diffusion-based models, such as BlindDPS~\cite{chung2023parallel}, model blur kernels via diffusion priors, they often necessitate the inclusion of additional manual priors (e.g., $\ell_0$ or $\ell_1$ regularization) to address sparse kernel structures.  This reliance on hand-crafted priors introduces additional complexity, as it requires careful tuning of regularization parameters for each dataset to achieve optimal performance. 

\subsection{Learning the Kernel Initializer}\label{sec:ki_learning}
\input{algorithm/initializer.tex}

To alleviate the sensitivity issue caused by kernel initialization, we propose to initialize the blur kernel in the latent space through an additional encoder $E(\cdot;\theta_E^{\ast})$, which maps the blurry image $\boldsymbol{y}$ to the corresponding latent code $\boldsymbol{z}$ of our learned kernel DGP. This design aims to provide a more stable and accurate starting point by leveraging the kernel encoder, thus facilitating the BMD process. Intuitively, the encoder can be directly trained by solving the following optimization problem:
\begin{equation}
    \theta_E^{\ast} = \arg\mathop{\min}_{\theta_E} \sum\nolimits_n \ell(G_k(E(\boldsymbol{y}_n;\theta_E);\theta_k^{\ast}),\boldsymbol{k}_n),
    \label{eq:encoder1}
\end{equation}
where $\boldsymbol{y}_n$ is the $n$-th blurry image, $\boldsymbol{k}_n$ is the corresponding blur kernel, $\theta_k^{\ast}$ denotes the pre-trained parameters of the kernel generator, and $\ell(\cdot,\cdot)$ is a suitable loss function.

However, training the encoder directly through supervision in the kernel space, as formulated in Eq.~\eqref{eq:encoder1}, proves challenging due to the inherently complex structures of blur kernels. To address this issue, we develop a collaborative learning strategy inspired by \cite{guan2020collaborative} to ease the training process. This strategy introduces an auxiliary variable $\boldsymbol{z}_n$, representing the latent code of the kernel $\boldsymbol{k}_n$ through the learned kernel generator, to split Eq.~\eqref{eq:encoder1} as follows:
\begin{equation}
    \begin{aligned}
    \min_{\theta_E} \sum_n \Bigl\{ & \|G_k(E(\boldsymbol{y}_n; \theta_E); \theta_k^{\ast}) - G_k(\boldsymbol{z}_n; \theta_k^{\ast})\|_1 \\
    & + \lambda \|E(\boldsymbol{y}_n; \theta_E) - \boldsymbol{z}_n\|_2^2 \Bigr\}, \\
    \text{s.t. } & \boldsymbol{z}_n = \arg\min_{\boldsymbol{z}} \|G_k(\boldsymbol{z}; \theta_k^{\ast}) - \boldsymbol{k}_n\|_1,
    \end{aligned}
    \label{eq:encoder2}
\end{equation}
where $\lambda$ is a tuning parameter set to $0.1$ in this work. By introducing such an auxiliary variable, we move the optimization to the latent space, which is more tractable owing to its typically compactness compared to the original kernel space. For convenience, we denote the objective functions of the outer and inner optimizations as $\mathcal{L}_E(\theta_E; \{\boldsymbol{z}_n\})$ and $\mathcal{L}_z^n(\boldsymbol{z})$, respectively, in the following.

The optimization of Eq.~\eqref{eq:encoder2} involves a sub-task of solving for $\boldsymbol{z}_n$, which is inherently a GAN-inversion problem~\cite{xia2022gan}. 
To solve this problem, we alternatively optimize $\boldsymbol{z}_n$ and $\theta_E$ in each iteration. Initially, $\boldsymbol{z}_n$ is set as $E(\boldsymbol{y}_n;\theta_E)$, and then updated through several gradient descent steps by minimizing the lower-level objective $\mathcal{L}_z^n(\boldsymbol{z})$. Subsequently, based on the updated $\boldsymbol{z}_n$, $\theta_E$ is optimized according to the upper-level objective $\mathcal{L}_E(\theta_E; \{\boldsymbol{z}_n\})$, also by gradient descent steps. The whole procedure is summarized in Algorithm \ref{algo_ki}.

Once trained, the encoder $E(\cdot;\theta_E^{\ast})$ can be regarded as an effective kernel initializer, providing promising predictions of blur kernels directly from blurry images. 
Although the prediction is not perfect, it closely resembles the ground truth kernel, and thus is expected to serve as a good starting point for subsequent BMD processing. 

\section{Proposed Method: Deblurring Process}\label{sec:bmd_process}
In this section, we provide a detailed description of the application of our pre-trained kernel generator and initializer to the BMD process. We first present our framework for uniform blind motion deblurring, demonstrating its effectiveness when integrated with various image priors, including DIP~\cite{ren2020neural}, VDIP~\cite{huo2023blind}, BIRD~\cite{chihaoui2024blind}, and BlindDPS~\cite{chung2023parallel}. Then, we extend our approach to handle non-uniform blind motion deblurring, where multiple spatially varying blur kernels are required to be estimated simultaneously. 

\subsection{Uniform Blind Motion Deblurring Framework}
\input{algorithm/uniform_dip.tex}

Given the pre-trained kernel generator $G_k(\cdot;\theta_k^{\ast})$ and kernel initializer $E(\cdot;\theta_E^{\ast})$, we can formulate the uniform BMD problem as follows:
\begin{equation}
    \min_{\boldsymbol{z}_k,\theta_x} \mathcal{L}_{\text{BMD}} \left(G_k(\boldsymbol{z}_k;\theta_k^{\ast}), G_x(\boldsymbol{z}_x;\theta_x), \boldsymbol{y}\right),
    \label{eq:bid_ours_naive_1}
\end{equation}
where $G_x(\boldsymbol{z}_x;\theta_x)$ is an image generator with input $\boldsymbol{z}_x$ and parameters $\theta_x$, $\mathcal{L}_\text{BMD}$ is the loss function.
$\boldsymbol{z}_k$ represents the latent code for the kernel, which is initialized by our pre-trained kernel initializer as $\boldsymbol{z}_k^{0}=E(\boldsymbol{y};\theta_E^{\ast})$. After optimization, the desirable sharp image and blur kernel can be obtained via $\hat{\boldsymbol{x}}=G_x(\boldsymbol{z}_x;\theta_x^{\ast})$ and $\hat{\boldsymbol{k}}=G_k(\boldsymbol{z}_k^{\ast};\theta_k^{\ast})$, respectively, where $\theta_x^{\ast}$ and $\boldsymbol{z}_k^{\ast}$ are the optimal solution of \eqref{eq:bid_ours_naive_1}. However, such a straightforward solution often yields suboptimal results, particularly when dealing with large blur kernels. This is mainly attributed to the dimensional disparity between the low-dimensional latent code $\boldsymbol{z}_k$ and the higher-dimensional blur kernel, which imposes overly restrictive constraints on the solution space and consequently increases the challenges of the optimization problem.

To address this issue, we propose to implement the optimization on the feature map of the kernel generator's first layer rather than its input. This strategy is inspired by StyleGAN ~\cite{karras2019style}, which introduces a mapping network that transforms the latent code to a style vector to enhance the generation control. The higher-dimensional feature map in the first layer of the kernel generator offers a richer optimization space than the constrained latent code. By denoting the first layer of $G_k(\cdot;\theta_k^{\ast})$ as $g_1(\cdot)$ and the truncated generator without $g_1(\cdot)$ as $G_k^{(w)}(\cdot)$, the BMD problem of Eq.~\eqref{eq:bid_ours_naive_1} is reformulated as
\begin{equation}    
\min_{\boldsymbol{w}_k,{\theta}_x} \mathcal{L}_\text{BMD} \left(G_k^{(w)}(\boldsymbol{w}_k),G_x(\boldsymbol{z}_x;{\theta}_x),\boldsymbol{y} \right),
    \label{eq:bid_ours_1}
\end{equation}
where $\boldsymbol{w}_k$ is initialized as $g_1\left(E(\boldsymbol{y};\theta_E^{\ast})\right)$. 

It is worth noting that fine-tuning the whole parameters of the generator has been explored in previous studies~\cite{pan2021exploiting}.  However, due to the highly non-convex nature of the BMD optimization, optimizing too many parameters simultaneously can potentially degrade performance rather than enhance it. Therefore, in our approach, we strategically limit the optimization to only the latent representation $\boldsymbol{w}_k$ for the blur kernel. 

\input{algorithm/uniform_bird.tex}

Next, we extend our method to various BMD frameworks with two representative generation priors as follows:

\subsubsection{DIP}
DIP~\cite{ren2020neural} explores the implicit prior embedded in convolutional neural networks (CNNs) for image restoration tasks. A typical DIP-based method for BMD is SelfDeblur \cite{ren2020neural}, which models the sharp image and the blur kernel as a CNN and a multi-layer perception (MLP), respectively. Similarly to SelfDeblur, it is easy to apply the proposed latent kernel modeling technique to DIP, \textit{i.e.},
\begin{equation}
    \min_{\boldsymbol{w}_k,\theta_x} \left\|G_k^{(w)}(\boldsymbol{w}_k) \otimes G_x(\boldsymbol{z}_x; \theta_x) - \boldsymbol{y}\right\|_2^2.
\end{equation}
The optimization process alternates between updating $\boldsymbol{w}_k$ and $\theta_x$ using gradient descent, as outlined in Algorithm \ref{algo:bmd_dip}. 

VDIP~\cite{huo2023blind} is an improved version of DIP that solves the problem using variational Bayesian inference. Naturally, our proposed latent kernel modeling method can also be applied within the VDIP framework, with detailed implementation provided in the supplementary material.

\begin{figure}[t]
    \begin{minipage}{1\columnwidth}
        \centering
        \includegraphics[width=1\textwidth]{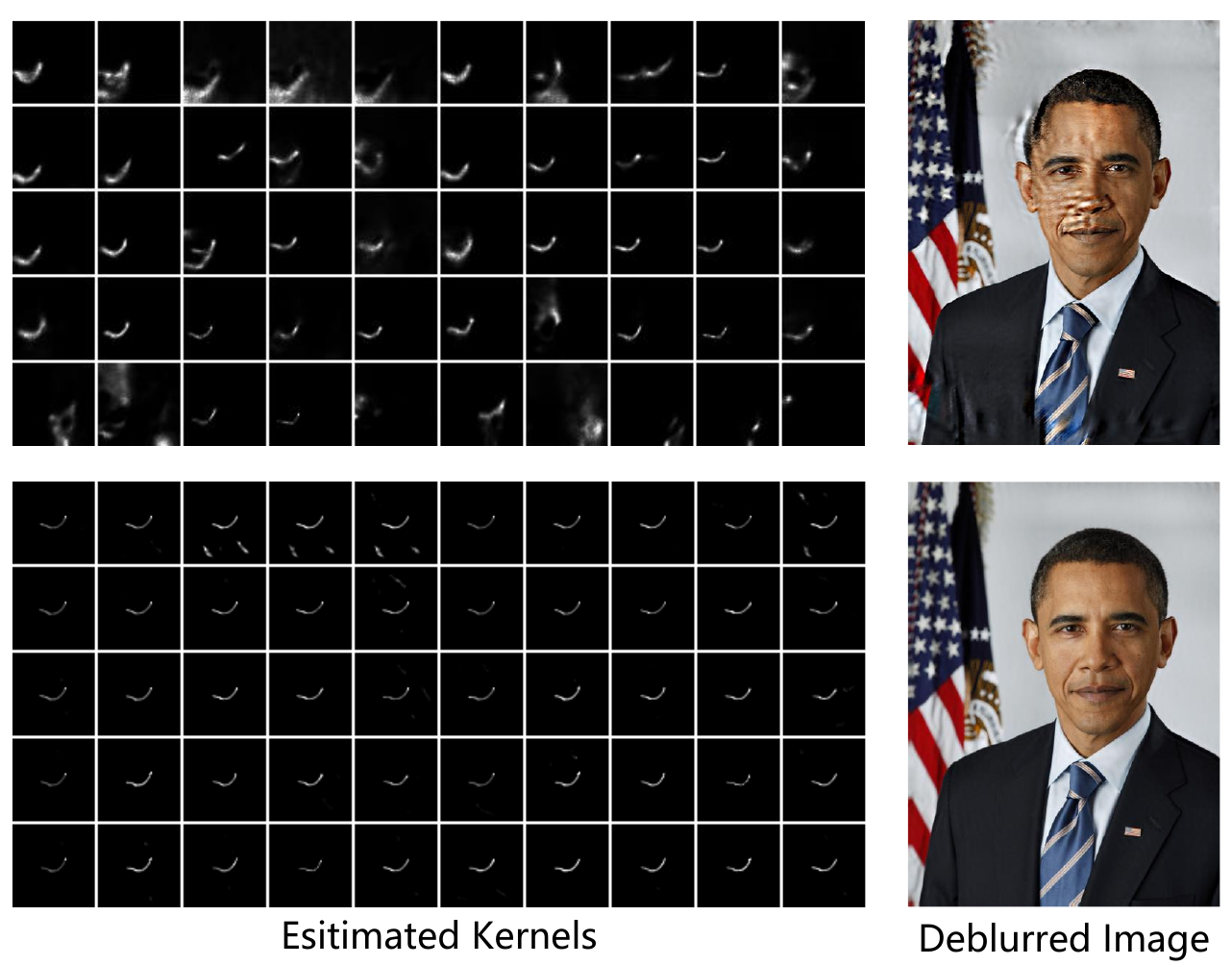}
    \end{minipage}
    \begin{minipage}{1\columnwidth}
        \centering
        \caption{Illustration of our method's effectiveness. The first row shows results from VDIP-Std \cite{huo2023blind}, while the second row shows results from our VDIP-Std-GLKM. From left to right: the finally estimated kernels and the corresponding deblurred images.}
        \label{fig:nonuniform}
    \end{minipage}
\end{figure}

\subsubsection{Diffusion models}
DDPM \cite{ho2020denoising} pioneers a newly powerful generative paradigm, which models the image generation process as a Markovian chain. 
Subsequently, Song \MakeLowercase{\textit{et al.}} ~\cite{song2021denoising} proposed a non-Markovian diffusion model, namely DDIM, which renders a deterministic sampling process as follows: 
\begin{equation}
    \boldsymbol{x}_{t- {\delta t}} = \sqrt{\bar{\alpha}_{t-{\delta t}}}  \hat{\boldsymbol{x}}_{0|t} + \sqrt{1 - \bar{\alpha}_{t-{\delta t}}} . \frac{\boldsymbol{x}_t - \sqrt{\bar{\alpha}_{t}} \hat{\boldsymbol{x}}_{0|t}}{\sqrt{1 - \bar{\alpha}_{t}}},
    \label{eq:ddim_inf}
\end{equation}
where 
\begin{align}
    \hat{\boldsymbol{x}}_{0|t} = \frac{(\boldsymbol{x}_t - \sqrt{1 - \bar{\alpha}_{t}} \epsilon_{\theta} (\boldsymbol{x}_t, t))}{\sqrt{\bar{\alpha}_{t}}},
    \label{eq:ddim_inf_one} 
\end{align}
$\epsilon_{\theta} (\boldsymbol{x}_t, t)$ denotes the diffusion UNet with a noise-prediction mode, and $\bar{\alpha}_{t}$ is a pre-defined hyper-parameter controlling the noise schedule. By iteratively applying Eq.~\eqref{eq:ddim_inf}, we can easily obtain a deterministic and differentiable transition function from $\boldsymbol{x}_T$ to $\boldsymbol{x}_0$, denoted as $\text{DDIMReverse}( \boldsymbol{x}_T, \delta t)$.

BIRD~\cite{chihaoui2024blind} employed this transition function as an image prior to solve the BMD problem. Our proposed latent kernel modeling technique can be seamlessly embedded into BIRD:
\begin{equation}
    \min_{\boldsymbol{w}_k,\boldsymbol{x}_T} \left\| G_k^{(w)}(\boldsymbol{w}_k) \otimes \text{DDIMReverse}( \boldsymbol{x}_T, \delta t) - \boldsymbol{y}\right\|_2^2.
\end{equation}
The detailed optimization process is listed in Algorithm~\ref{algo:bmd_bird}, where $\boldsymbol{w}_k$ and $\boldsymbol{x}_T$ are iteratively updated based on their respective gradients. Note that $\boldsymbol{x}_T$ is normalized after each update step following the official implementation of BIRD.

Beyond BIRD, our proposed latent kernel modeling can also be integrated into other diffusion model-based BMD methods, such as BlindDPS~\cite{chung2023parallel}. The detailed implementation is presented in the supplementary material.

\subsection{Non-uniform Blind Motion Deblurring Framework}\label{sec:nonuniformbmd}

\input{algorithm/nonuniform.tex}

After introducing our framework for uniform BMD, we now extend it to handle more challenging non-uniform cases. To model the non-uniform motion blur of static scenes, we utilize the space-variant overlap-add (SVOLA) formulation~\cite{harmeling2010space,hirsch2010efficient}, which describes the relationship between the blurry image $\boldsymbol{y}$ and its underlying sharp counterpart $\boldsymbol{x}$ as follows:
\begin{equation}
    \boldsymbol{y} = \mathcal{F} \Bigl(\boldsymbol{k}, \boldsymbol{x} \Bigr) + \boldsymbol{n} = \mathcal{F} \Bigl(\{\boldsymbol{k}_i\}_{i=1}^P, \boldsymbol{x}\Bigr) + \boldsymbol{n},
    \label{eq:svola}
\end{equation}
Specifically, the value at pixel position $\boldsymbol{p}=(x,y)$ is given by
\begin{equation}
    \boldsymbol{y}(\boldsymbol{p}) = \sum_{i=1}^{P} \Bigl( \boldsymbol{k}_{i} \otimes \bigl( \boldsymbol{w}(\boldsymbol{p}-{c}_{i}) \odot \mathcal{P}_{i} \boldsymbol{x}(\boldsymbol{p}) \bigr) \Bigr) + \boldsymbol{n}(\boldsymbol{p}),
\end{equation}
where $\odot$ denotes entry-wise multiplication, $\otimes$ denotes convolution, $\mathcal{P}_i$ is a mask operator that extracts the $i$-th patch from the image, $\boldsymbol{k}_i$ is the $i$-th kernel, and $\boldsymbol{w}(\cdot - c_i)$ is a window function that is translated to align with the center $c_i$ of the $i$-th image patch. 

Unlike uniform cases, the non-uniform BMD problem requires simultaneous estimation of both the latent sharp image $\boldsymbol{x}$ and multiple region-specific blur kernels $\boldsymbol{k}=\{\boldsymbol{k}_i\}_{i=1}^P$. This presents two key challenges: the difficulty in kernel estimation for regions with insufficient distinctive features, and the probability for randomly initialized kernels to deviate from the true kernel manifold. These challenges can lead to boundary artifacts in the restored image, as illustrated through a representative example in Fig.~\ref{fig:nonuniform}.

Based on these observations, we propose to introduce a prior on the kernel set to effectively estimate kernels across all image regions using our latent kernel modeling strategy. In real-world scenarios, blur kernels are constrained by the camera's motion trajectory and scene depth, sharing the same 3D camera motion that forms a lower-dimensional manifold \cite{joshi2010image, carbajal2023blind}. To leverage this implicit relationship among kernels, we adopt a two-stage coarse-to-fine approach: first estimating a global kernel $\boldsymbol{k}$ as coherent initialization, then finetuning local kernels through joint optimization while maintaining their relationship to the underlying camera motion. 

Specifically, taking image generation prior DIP as an example, we initialize the latent representations $\boldsymbol{w}_{k}=\{\boldsymbol{w}_{ki}\}_{i=1}^P$ using our pre-trained kernel generator and initializer by setting $\boldsymbol{w}_{ki}^{0} \gets g_1(E(\boldsymbol{y};\theta_E^{\ast})),~i = 1, \ldots, P$. We then jointly optimize the latent kernel representations $\boldsymbol{w}_{k}$ and the parameters $\theta_x$ of the image generator DIP, \textit{i.e.},
\begin{equation}
    \min_{\boldsymbol{w}_{k},\theta_x} \left\|\mathcal{F}\left({G_{k}^{(w)}\left(\boldsymbol{w}_{k}\right)},G_{x}\left(\boldsymbol{z}_{x} ; \theta_{x}\right)\right)-\boldsymbol{y}\right\|_2^{2}.
\end{equation}
The complete non-uniform BMD process is summarized in Algorithm~\ref{algo_bmd_nonuniform}. Note that our non-uniform BMD method can also be naturally extended to various BMD frameworks like VDIP and diffusion models.

\section{Experiments on Uniform BMD}\label{sec:exp_uniform}
\input{table/uniform_synthetic.tex}
In this section, we evaluate the effectiveness of our proposed \textbf{G}enerative \textbf{L}atent \textbf{K}ernel \textbf{M}odeling (GLKM) technique in uniform deblurring scenarios based on two DIP-based methods, namely DIP~\cite{ren2020neural} and VDIP~\cite{huo2023blind}, as well as two diffusion model-based methods, namely BIRD~\cite{chihaoui2024blind} and BlindDPS~\cite{chung2023parallel}. For VDIP, we consider three variants: VDIP-Std, VDIP-Extreme, and VDIP-Sparse, corresponding to the original VDIP, the extreme channel prior version, and the sparse image prior version, respectively. For clarity in the subsequent analysis, we adopt the notation ``[Baseline]-GLKM'' (e.g., DIP-GLKM and BIRD-GLKM) to indicate the performance of our approach applied to the four baseline methods above. For all comparison methods, we follow the officially default settings of these methods or tune them ourselves for the best performance.

\noindent \textbf{Pre-training Setup.} In the pre-training phase, we first train the kernel generator $G_k(\cdot;\theta_k^{\ast})$ following DCGAN~\cite{radford2015unsupervised}, with kernels synthesized as described in \cite{kupyn2018deblurgan} and \cite{chung2023parallel}. 
Then we freeze the $G_k(\cdot;\theta_k^{\ast})$ and train the kernel initializer $E(\cdot;\theta_E^{\ast})$ that adopts the ResNet-18 \cite{he2016deep} architecture. In training the kernel initializer, we randomly crop $256\times256$ patches from the source images in OpenImages \cite{kuznetsova2020open} as sharp images to generate the blurry ones. The Adam optimizer \cite{kingma2014adam} with its default configuration is employed, with an initial learning rate of $1\times10^{-4}$, which is reduced to $1\times10^{-5}$ once the loss stabilizes, and further to $1\times10^{-6}$ as training progresses.

\noindent \textbf{Evaluation Metrics.} For datasets with ground truth sharp images, we employ three reference-based metrics: Peak Signal-to-Noise Ratio (PSNR), Structural Similarity Index Measure (SSIM), and Learned Perceptual Image Patch Similarity (LPIPS)~\cite{zhang2018unreasonable}. For real-world datasets without ground truth, we utilize three no-reference metrics: NIQE~\cite{mittal2012making}, BRISQUE~\cite{mittal2011blind}, and PIQE~\cite{venkatanath2015blind} to evaluate the restored image quality.

\subsection{BMD with DIP-based Methods}

\input{table/uniform_lai.tex}
\noindent \textbf{Testing Datasets.} 
We evaluate the proposed GLKM equipped with DIP-based image generators on a synthetic dataset and the well-known BMD benchmark established by Lai \MakeLowercase{\textit{et al.}}~\cite{lai2016comparative}. For the synthetic dataset, we randomly selected 80 images from MSCOCO~\cite{lin2014microsoft} to synthesize the blurry images following~\cite{kupyn2018deblurgan} or \cite{lai2016comparative}. The Lai dataset consists of 100 synthetic blurry images and several real-world blurry images. The synthetic images were obtained by applying 4 blur kernels, with sizes ranging from $31\times31$ to $75\times75$, to 25 clean images. These 25 clean images are categorized into 5 groups, namely \emph{Manmade}, \emph{Natural}, \emph{People}, \emph{Saturated}, and \emph{Text}.

\noindent \textbf{Implementation Details.} 
For both DIP and VDIP as image prior, the learning rate of $\boldsymbol{w}_k$ and $\theta_x$ are set as $5\times10^{-4}$ and $1\times10^{-2}$, and the number of optimization steps $T$ is 5,000. We follow the settings in the original paper for the architecture of $G_x(\cdot;\theta_x)$. 

\noindent \textbf{Comparison Methods.} 
To verify the effectiveness of the proposed GLKM, we compare it against four traditional model-based methods (Xu \MakeLowercase{\textit{et al.}} \cite{xu2013unnatural}, Dong \MakeLowercase{\textit{et al.}} \cite{dong2017blind}, Pan et al \cite{pan2016l_0,pan2017deblurring}), five supervised deep learning methods (Kupyn \MakeLowercase{\textit{et al.}} \cite{kupyn2019deblurgan}, Kaufman \MakeLowercase{\textit{et al.}} \cite{kaufman2020deblurring}, Cho \MakeLowercase{\textit{et al.}} \cite{cho2021rethinking}, Zamir \MakeLowercase{\textit{et al.}} \cite{zamir2021multi,zamir2022restormer}), and three DIP-based methods (Ren \MakeLowercase{\textit{et al.}} \cite{ren2020neural}, Huo \MakeLowercase{\textit{et al.}} \cite{huo2023blind}, Zhang \MakeLowercase{\textit{et al.}} \cite{zhang2024cross}).

\noindent \textbf{Experimental results.} 
Tab.~\ref{tab:uniform_synthetic} and \ref{tab:uniform_lai} report the quantitative results of all competing methods on the synthetic datasets. Across all datasets and image categories, our proposed GLKM applied to DIP-based methods consistently outperforms the corresponding baseline approaches in all metrics. Notably, GLKM yields the most substantial enhancement for the original DIP approach (34.2\% average PSNR improvement), followed by VDIP-Std (25.8\% average PSNR improvement), while providing more modest gains for VDIP-Sparse and VDIP-Extreme. This demonstrates that the carefully designed GLKM technique can provide sufficient regularization for the BMD process, eliminating the need for additional handcrafted priors.

As observed in Fig.~\ref{fig:uniform_synthetic} and \ref{fig:uniform_lai}, our GLKM demonstrates significant advantages in visual quality compared to existing approaches. 
While traditional model-based methods achieve relatively accurate blur kernel estimations, their final deblurring results remain suboptimal due to the inherent limitations of handcrafted image priors. Supervised deep learning methods, despite their sophisticated architectures, tend to produce over-smoothed results, primarily due to their insufficient exploitation of the underlying physical blur model. Even though achieving relatively better results, the existing deep prior-based methods still have some limitations. Specifically, Ren \MakeLowercase{\textit{et al.}}'s~\cite{ren2020neural} approach yields unsatisfactory results, while both Huo \MakeLowercase{\textit{et al.}}~\cite{huo2023blind} and Zhang \MakeLowercase{\textit{et al.}}'s~\cite{zhang2024cross} methods struggle to preserve sharp structural details in the recovered images. In contrast, our proposed GLKM demonstrates dual advantages: it not only achieves more accurate blur kernel estimation but also recovers finer image details with significantly reduced artifacts and distortions.

Tab.~\ref{tab:real_lai} shows the quantitative results on Lai \MakeLowercase{\textit{et al.}}'s real blurred dataset. As can be seen, GLKM applied to DIP-based methods can generate images of the highest quality based on NIQE, BRISQUE and PIQE among all compared methods. As shown in Fig.~\ref{fig_real}, our method has relatively higher or at least comparable visual quality, showing its potential in dealing with blurry images that are with unknown complex blur kernels.

\input{table/real_lai.tex}

\input{fig/uniform_synthetic.tex}

\input{fig/uniform_lai.tex}

\input{fig/real_lai.tex}

\subsection{BMD with Diffusion Model-based Methods}
\noindent \textbf{Evaluation Datasets.} To assess the performance of our GLKM with diffusion-based image generation methods, we employ three widely-used benchmark datasets at 256$\times$256 resolution: AFHQ-dog~\cite{choi2020stargan}, CelebAHQ~\cite{liu2018large}, and ImageNet~\cite{deng2009imagenet}. Specifically, we utilize validation subsets containing 500 images from both AFHQ-dog and CelebAHQ, along with 1,000 test images from ImageNet. Following~\cite{chung2023parallel}, we adopt their pre-trained diffusion models in our experiments. The evaluation protocol involves applying synthetic motion blur with an intensity parameter of 0.5, consistent with the setup in~\cite{chung2023parallel}.

\noindent \textbf{Implementation Details.} 
For BIRD as image generator, we use $\delta t = 100$ and iteration $N = 100$ following~\cite{chihaoui2024blind}. The learning rate of $\boldsymbol{w}_k$ and $\boldsymbol{x}_T$ are set as $2\times10^{-4}$ and $3\times10^{-3}$, respectively. For BlindDPS as image generator, the learning rate of $\boldsymbol{w}_k$ and $\boldsymbol{x}_t$ are set as $5\times10^{-3}$ and $0.3$, respectively. 

\noindent \textbf{Comparison Methods.} Our experimental evaluation includes five representative deblurring approaches: Pan \MakeLowercase{\textit{et al.}} \cite{pan2017deblurring} representing traditional optimization methods, MPRNet \cite{zamir2021multi} showcasing supervised deep learning techniques, Ren \MakeLowercase{\textit{et al.}} \cite{ren2020neural} exemplifying DIP-based approaches, alongside two recent diffusion-based solutions by Chihaoui \MakeLowercase{\textit{et al.}}\cite{chihaoui2024blind} and Chung \MakeLowercase{\textit{et al.}}\cite{chung2023parallel}.

\noindent \textbf{Experimental results.} Tab.~\ref{tab:uniform_diffusion} presents a comprehensive quantitative comparison of all competing methods. The results clearly demonstrate that our proposed GLKM consistently outperforms existing diffusion model-based approaches across all evaluation datasets and metrics.
GLKM significantly improves upon BIRD~\cite{chihaoui2024blind}, whose kernel estimation relies on a simple MLP network. More notably, GLKM also enhances BlindDPS~\cite{chung2023parallel}, despite the latter's use of pre-trained diffusion models with handcrafted kernel priors. This result highlights the superior robustness of our approach for BMD.

Fig.~\ref{fig:afhq}, \ref{fig:celebahq}, and \ref{fig:imagenet} provide a visual comparison of the results. The kernels estimated by GLKM exhibit clearer motion trajectories and fewer noise artifacts than competing methods. Consequently, our restorations achieve higher fidelity and better preserve intricate details.

\input{table/uniform_diffusion.tex}

\input{fig/uniform_diffusion.tex}

\section{Experiments on Non-uniform BMD}
Having demonstrated the effectiveness of GLKM in uniform deblurring cases, we now extend it to non-uniform scenarios. Since the pre-trained diffusion models are limited to generating images at specific resolutions, we only evaluate our proposed GLKM technique based on DIP-based methods.

\noindent \textbf{Pre-training Setup.} It is worth noting that the kernel generator $G_k(\cdot;\theta_k^{\ast})$ and kernel initializer $E(\cdot;\theta_E^{\ast})$ pre-trained for uniform BMD cases can be directly applied to non-uniform BMD scenarios without additional retraining. This transferability demonstrates the robustness of our approach and its ability to generalize across different blur types. 

\noindent \textbf{Evaluation Metrics.} We employ three reference-based metrics: Peak Signal-to-Noise Ratio (PSNR), Structural Similarity Index Measure (SSIM), and Learned Perceptual Image Patch Similarity (LPIPS)~\cite{zhang2018unreasonable}.

\subsection{BMD with DIP-based Methods}
\input{table/nonuniform_synthetic.tex}

\input{table/nonuniform_lai.tex}

\input{fig/nonuniform_synthetic.tex}

\input{fig/nonuniform_lai.tex}
\noindent \textbf{Testing Datasets.} For evaluating our method on non-uniform BMD, we conduct experiments on two datasets: a synthetic dataset and the benchmark dataset from Lai \MakeLowercase{\textit{et al.}} \cite{lai2016comparative}. For our synthetic dataset, we randomly selected 80 images from MSCOCO \cite{lin2014microsoft} and applied non-uniform motion blur synthesized using motion trajectories as described in \cite{lai2016comparative}. The Lai \MakeLowercase{\textit{et al.}} \cite{lai2016comparative}'s dataset consists of 100 images created by recording real camera motions with inertial sensors. These motions were used to construct spatially varying blur kernels (assuming constant scene depth), which were then applied to sharp images with 1\% Gaussian noise added. The dataset is organized into 5 categories of 20 images each, featuring complex scenes with challenging non-uniform blur patterns.

\noindent \textbf{Implementation Details.} 
When implementing DIP-based image priors, we optimize the parameters with the following configuration: the optimization process runs for 5,000 iterations, with learning rates of $5\times10^{-4}$ and $1\times10^{-2}$ applied to $\boldsymbol{w}_k$ and $\theta_x$ respectively. To ensure fair experimental comparison, we maintain identical network architectures as the original DIP~\cite{ren2020neural} and VDIP~\cite{huo2023blind}. Additionally, we configure the grid size $P$ in Eq.~\ref{eq:svola} as $5\times 10$.

\noindent \textbf{Comparison Methods.} We verify the effectiveness of the proposed method on our synthetic dataset and Lai \MakeLowercase{\textit{et al.}}'s non-uniform blurred dataset. 12 comparison methods are considered, including three traditional model-based methods (Xu \MakeLowercase{\textit{et al.}} \cite{xu2013unnatural}, Whyte \MakeLowercase{\textit{et al.}} \cite{whyte2012non}, Vasu \MakeLowercase{\textit{et al.}} \cite{vasu2017local}), four supervised deep learning methods (Cho \MakeLowercase{\textit{et al.}} \cite{cho2021rethinking}, Kupyn \MakeLowercase{\textit{et al.}} \cite{kupyn2019deblurgan}, MPRNet \cite{zamir2021multi}, Restormer \cite{zamir2022restormer}), and five unsupervised methods (Zhang \MakeLowercase{\textit{et al.}} \cite{zhang2023event}, Fang \MakeLowercase{\textit{et al.}} \cite{fang2023self}, Li \MakeLowercase{\textit{et al.}} \cite{li2023self}, Ren \MakeLowercase{\textit{et al.}} \cite{ren2020neural},Huo \MakeLowercase{\textit{et al.}} \cite{huo2023blind}). Since DIP~\cite{ren2020neural} and VDIP~\cite{huo2023blind} were originally designed for uniform BMD, we adapt them for non-uniform cases by integrating the SVOLA formulation (Eq.~\ref{eq:svola}) into their respective frameworks.

\noindent \textbf{Experimental Results.} The quantitative results of all competing methods are summarized in Tab.~\ref{tab:nonuniform_synthetic} and \ref{tab:nonuniform_lai}. As evidenced by these comprehensive evaluations, DIP with GLKM achieves the best average performance on Lai \MakeLowercase{\textit{et al.}}'s non-uniform dataset with the highest PSNR (22.11), SSIM (0.744), and lowest LPIPS (0.221). 
Particularly noteworthy is GLKM's remarkable performance on the challenging Text category, where DIP-GLKM is the only approach to achieve a PSNR exceeding 20 and an SSIM surpassing 0.8. 

Fig.~\ref{fig:nonuniform_synthetic} and \ref{fig:nonuniform_lai} demonstrate the superior visual quality of our method. As can be observed, compared with competing methods, GLKM effectively preserves sharper edges and finer details while avoiding ringing artifacts. This leads to outputs with clearer textures and more natural-looking features, particularly noticeable in challenging areas containing text or complex patterns.

\section{Conclusion}
In this paper, we have proposed a new framework, GLKM, for the BMD task. Within this framework, we first pre-train a kernel generator as a DGP for blur kernels and a kernel initializer that can offer a well-initialized kernel. Then, during the BMD process, the blur kernel is initialized in the latent space and jointly optimized with various image generators. We also extend GLKM to non-uniform BMD scenarios without the need for extra priors. Comprehensive experiments have demonstrated the effectiveness of GLKM, showing that it serves as a plug-and-play kernel prior that can be easily integrated with existing BMD methods to significantly enhance their restoration performance. In the future, we will make further efforts to address the limitations discussed in the supplementary material.

\section*{Acknowledgments}
This work was supported in part by the NSFC Projects (No. 12226004, 62076196, 62331028, 62272375). The authors thank Prof. Jiangxin Dong for providing the code for generating the blur kernels with the type of Lai \MakeLowercase{\textit{et al.}}'s dataset.

 
%
\bibliographystyle{IEEEtran}
\bibliography{ref}

\end{document}

%% file: algorithm/initializer.tex
\begin{algorithm}[t]
    \caption{Kernel Initializer Learning}
    \label{algo_ki}
    \begin{algorithmic}[1]
        \footnotesize
        \REQUIRE Pre-trained kernel generator $G_k(\cdot;\theta_k^{\ast})$, blurry image-kernel pairs $\{\boldsymbol{y}_n,\boldsymbol{k}_n\}$, step size $\alpha$
        \STATE Initialize $\theta_E^{0}$
        \FOR{$t=1$ \TO $T$}
            \FOR{each $n$}
            \STATE $\boldsymbol{z}^{0}=E\big(\boldsymbol{y}_n;\theta_E^{t-1}\big)$
                \FOR{$s=1$ \TO $S$}
                \STATE $\boldsymbol{z}^{s}=\boldsymbol{z}^{s-1}-\alpha\nabla_{\boldsymbol{z}}\mathcal{L}_z^{n}(\boldsymbol{z})|_{\boldsymbol{z}=\boldsymbol{z}^{s-1}}$
                \ENDFOR
            \STATE $\boldsymbol{z}_n^{t}=\boldsymbol{z}^{S}$
            \ENDFOR
            \STATE $\theta^{0}=\theta_E^{t-1}$
            \FOR{$l=1$ \TO $L$}
                \STATE $\theta^{l}=\theta^{l-1}-\alpha\nabla_\theta\mathcal{L}_E(\theta; \{\boldsymbol{z}_n^{t}\})|_{\theta=\theta^{l-1}}$
            \ENDFOR
            \STATE $\theta_E^{t}=\theta^{L}$
        \ENDFOR
        \ENSURE Kernel initializer $E(\cdot;\theta_E^{\ast})$, where $\theta_E^{\ast}=\theta_E^{T}$
    \end{algorithmic}
\end{algorithm}

%% file: algorithm/uniform_dip.tex
\begin{algorithm}[t]
    \caption{Uniform BMD with DIP}
    \label{algo:bmd_dip}
    \begin{algorithmic}[1]
        \footnotesize
        \REQUIRE Pre-trained kernel generator $G_k(\cdot;\theta_k^{\ast})=G_k^{(w)}(g_1(\cdot))$, pre-trained kernel initializer $E(\cdot;\theta_E^{\ast})$, blurry image $\boldsymbol{y}$, DIP network $G_x(\cdot;\theta_x)$ with its input $\boldsymbol{z}_x$, learning rate $\alpha$
        \STATE Initialize $\boldsymbol{w}_k^{0} \gets g_1(E(\boldsymbol{y};\theta_E^{\ast}))$
        \FOR{$t = 1$ \TO $T$}
            \STATE $\boldsymbol{w}_{k}^{t} \gets \boldsymbol{w}_{k}^{t-1} - \alpha\nabla_{\boldsymbol{w}_{k}}\left\|G_k^{(w)}(\boldsymbol{w}_k) \otimes G_x(\boldsymbol{z}_x; \theta_x) - \boldsymbol{y}\right\|_2^2$
            \STATE $\theta_x^{t} \gets \theta_x^{t-1} - \alpha\nabla_{\theta_x}\left\|G_k^{(w)}(\boldsymbol{w}_k) \otimes G_x(\boldsymbol{z}_x; \theta_x) - \boldsymbol{y}\right\|_2^2$
        \ENDFOR
        \STATE Set final parameters: $\boldsymbol{w}_k^{\ast} \gets \boldsymbol{w}_k^{T}$, ${\theta}_x^{\ast} \gets {\theta}_x^{T}$
        \ENSURE Estimated sharp image $\hat{\boldsymbol{x}}=G_x(\boldsymbol{z}_x,{\theta}_x^{\ast})$, blur kernel $\hat{\boldsymbol{k}} = G_k^{(w)}(\boldsymbol{w}_k^{\ast})$
        \normalsize
    \end{algorithmic}
\end{algorithm}

%% file: algorithm/uniform_bird.tex
\begin{algorithm}[t]
    \caption{Uniform BMD with BIRD}
    \label{algo:bmd_bird}
    \begin{algorithmic}[1]
        \footnotesize
        \REQUIRE Pre-trained kernel generator $G_k(\cdot;\theta_k^{\ast})=G_k^{(w)}(g_1(\cdot))$, pre-trained kernel initializer $E(\cdot;\theta_E^{\ast})$, blurry image $\boldsymbol{y}$, learning rate $\alpha$, step size $\delta t$, the dimension of $\boldsymbol{y}$ $d=N_x \times N_y$
        \STATE Initialize $\boldsymbol{w}_k^{0} \gets g_1(E(\boldsymbol{y};\theta_E^{\ast}))$,  $\boldsymbol{x}_T \sim \mathcal{N}(0,\mathbf{I})$
        \FOR{$t = 1$ \TO $T$}
            \STATE $\boldsymbol{w}_{k}^{t} \gets \boldsymbol{w}_{k}^{t-1} - \alpha\nabla_{\boldsymbol{w}_{k}}\left\|G_k^{(w)}(\boldsymbol{w}_k) \otimes \text{DDIMReverse}( \boldsymbol{x}_T, \delta t) - \boldsymbol{y}\right\|_2^2$
            \STATE $\boldsymbol{x}_T^t \gets \boldsymbol{x}_T^{t-1} - \alpha\nabla_{\boldsymbol{x}_T}\left\|G_k^{(w)}(\boldsymbol{w}_k) \otimes \text{DDIMReverse}( \boldsymbol{x}_T, \delta t) - \boldsymbol{y}\right\|_2^2$
            \STATE $\boldsymbol{x}_T^t \gets \frac{\boldsymbol{x}_T^t}{\left \| \boldsymbol{x}_T^t \right \| } \sqrt{d} $
        \ENDFOR
        \STATE Set final parameters: $\boldsymbol{w}_k^{\ast} \gets \boldsymbol{w}_k^{T}$, $\boldsymbol{x}_T^{\ast} \gets \boldsymbol{x}_T^{T}$
        \ENSURE Estimated sharp image $\hat{\boldsymbol{x}}= \text{DDIMReverse}(\boldsymbol{x}_T^{\ast}, \delta t)$, blur kernel $\hat{\boldsymbol{k}} = G_k^{(w)}(\boldsymbol{w}_k^{\ast})$
        \normalsize
    \end{algorithmic}
\end{algorithm}

%% file: algorithm/nonuniform.tex
\begin{algorithm}[t]
    \caption{Non-Uniform BMD Framework}
    \label{algo_bmd_nonuniform}
    \begin{algorithmic}[1]
        \footnotesize
        \REQUIRE Pre-trained kernel generator $G_k(\cdot;\theta_k^{\ast})=G_k^{(w)}(g_1(\cdot))$, pre-trained kernel initializer $E(\cdot;\theta_E^{\ast})$, blurry image $\boldsymbol{y}$, DIP network $G_x(\cdot;\theta_x)$ with its input $\boldsymbol{z}_x$, learning rate $\alpha$
        
        \STATE Initialize $\boldsymbol{w}_{ki}^{0} \gets g_1(E(\boldsymbol{y};\theta_E^{\ast})),\quad  i = 1, \ldots, P$
        
        
        \FOR{$t = 1$ \TO $T$}
            \STATE $\boldsymbol{w}_{k}^{t} \gets \boldsymbol{w}_{k}^{t-1} - \alpha\nabla_{\boldsymbol{w}_{k}}\left\|\mathcal{F}\left({G_{k}^{(w)}\left(\boldsymbol{w}_{k}\right)},G_{x}\left(\boldsymbol{z}_{x} ; \theta_{x}\right)\right)-\boldsymbol{y}\right\|_2^{2}$
            \STATE $\theta_x^{t} \gets \theta_x^{t-1} - \alpha\nabla_{\theta_x}\left\|\mathcal{F}\left({G_{k}^{(w)}\left(\boldsymbol{w}_{k}\right)},G_{x}\left(\boldsymbol{z}_{x} ; \theta_{x}\right)\right)-\boldsymbol{y}\right\|_2^{2}$
        \ENDFOR
        
        \STATE $\boldsymbol{w}_{k}^{\ast} \gets \boldsymbol{w}_{k}^{T}$, and $\theta_x^{\ast} \gets \theta_x^{T}$
        
        \ENSURE Estimated sharp image $\hat{\boldsymbol{x}} = G_x(\boldsymbol{z}_x;\theta_x^{\ast})$, estimated blur kernels $\hat{\boldsymbol{k}} = G_k^{(w)}(\boldsymbol{w}_k^{\ast})$
    \end{algorithmic}
\end{algorithm}

%% file: table/uniform_synthetic.tex
\begin{table}[!t]
    \centering
    \caption{Quantitative comparisons of various methods on our synthetic dataset. The best
    and second best results are highlighted in \textbf{bold} and \underline{underline}, respectively.}
    \label{tab:uniform_synthetic}
    \scriptsize
    \begin{tabular}{@{}C{2.5cm}@{}|@{}C{1.2cm}@{}|@{}C{1.2cm}@{}|@{}C{1.2cm}@{}}
        \Xhline{0.8pt}
        Method & PSNR$\uparrow$ & SSIM$\uparrow$ & LPIPS$\downarrow$ \\
        \hline
        Xu\cite{xu2013unnatural} & 16.56 & 0.356 & 0.666 \\
        Dong\cite{dong2017blind} & 18.25 & 0.455 & 0.575 \\
        Pan-DCP\cite{pan2017deblurring} & 21.65 & 0.592 & 0.475 \\
        Pan-$l_0$\cite{pan2016l_0} & 15.45 & 0.328 &  0.711\\
        Kupyn\cite{kupyn2019deblurgan} & 17.31 & 0.409 & 0.694 \\
        Kaufman\cite{kaufman2020deblurring} & 20.29 & 0.523 & 0.598 \\
        Cho\cite{cho2021rethinking} & 16.84 & 0.422 & 0.649 \\
        MPRNet\cite{zamir2021multi} & 16.73 & 0.384 & 0.668 \\
        Restormer\cite{zamir2022restormer} & 16.75 & 0.387 & 0.674 \\
        Zhang\cite{zhang2024cross} & 22.64 & 0.647 & 0.361 \\
        \hline \hline
        DIP\cite{ren2020neural} & 17.44 & 0.406 & 0.608  \\
        VDIP-Std\cite{huo2023blind} & 19.76 & 0.498 & 0.542  \\
        VDIP-
        Extreme\cite{huo2023blind} & 20.52 & 0.546 & 0.503  \\VDIP-Sparse\cite{huo2023blind} & 22.30 & 0.644 & 0.420 \\
        \hline \hline
        DIP-GLKM & \textbf{24.25} & \textbf{0.714} & \textbf{0.263}  \\
        VDIP-Std-GLKM & \underline{23.96} & \underline{0.703} & \underline{0.282}  \\
        VDIP-Extreme-GLKM & 23.75 & 0.692 & 0.306  \\
        VDIP-Sparse-GLKM & 23.29 & 0.667 & 0.318  \\
        \Xhline{0.8pt}
    \end{tabular}
\end{table}

%% file: table/uniform_lai.tex
\begin{table*}[!t]
    \centering
    \caption{Quantitative results of various methods on Lai et al.'s synthetic dataset. The best
    and second best results are highlighted in \textbf{bold} and \underline{underline}, respectively.}
    \label{tab:uniform_lai}
    \scriptsize
    \begin{tabular}{@{}C{2.4cm}@{}|@{}C{0.85cm}@{}|@{}C{0.85cm}@{}|@{}C{0.85cm}@{}|@{}C{0.85cm}@{}|@{}C{0.85cm}@{}|@{}C{0.85cm}@{}|@{}C{0.85cm}@{}|@{}C{0.85cm}@{}|@{}C{0.85cm}@{}|@{}C{0.85cm}@{}|@{}C{0.85cm}@{}|@{}C{0.85cm}@{}|@{}C{0.85cm}@{}|@{}C{0.85cm}@{}|@{}C{0.85cm}@{}|@{}C{0.85cm}@{}|@{}C{0.85cm}@{}|@{}C{0.85cm}@{}}
        \Xhline{0.8pt}
        \multirow{2}{*}{Method} & \multicolumn{3}{c|}{Manmade} & \multicolumn{3}{c|}{Natural} & \multicolumn{3}{c|}{People} & \multicolumn{3}{c|}{Saturated} & \multicolumn{3}{c|}{Text} & \multicolumn{3}{c@{}}{Average} \\ 
        \cline{2-19}
        & PSNR$\uparrow$ & SSIM$\uparrow$ & LPIPS$\downarrow$ & PSNR$\uparrow$ & SSIM$\uparrow$ & LPIPS$\downarrow$ & PSNR$\uparrow$ & SSIM$\uparrow$ & LPIPS$\downarrow$ & PSNR$\uparrow$ & SSIM$\uparrow$ & LPIPS$\downarrow$ & PSNR$\uparrow$ & SSIM$\uparrow$ & LPIPS$\downarrow$ & PSNR$\uparrow$ & SSIM$\uparrow$ & LPIPS$\downarrow$ \\
        \hline
        Xu\cite{xu2013unnatural} & 15.53 & 0.291 & 0.618 & 18.44 & 0.396 & 0.578 & 19.84 & 0.610 & 0.403 & 13.79 & 0.465 & 0.488 & 14.41 & 0.414 & 0.601 & 16.40 & 0.435 & 0.538 \\ 
        Dong\cite{dong2017blind} & 17.80 & 0.463 & 0.477 & 22.33 & 0.611 & 0.438 & 24.50 & 0.750 & 0.309 & 15.77 & 0.597 & 0.381 & 19.81 & 0.715 & 0.316 & 20.04 & 0.627 & 0.384 \\ 
        Pan-DCP\cite{pan2017deblurring} & 19.97 & 0.591 & 0.403 & 22.62 & 0.615 & 0.421 & 25.85 & 0.780 & 0.296 & 15.85 & 0.592 & 0.387 & 18.98 & 0.701 & 0.305 & 20.65 & 0.656 & 0.362 \\ 
        Pan-$l_0$\cite{pan2016l_0} & 13.67 & 0.281 & 0.548 & 16.69 & 0.415 & 0.536 & 18.00 & 0.558 & 0.419 & 12.32 & 0.422 & 0.484 & 15.78 & 0.537 & 0.388 & 15.29 & 0.442 & 0.475 \\ 
        Kupyn\cite{kupyn2019deblurgan} & 15.84 & 0.304 & 0.603 & 18.89 & 0.407 & 0.608 & 21.13 & 0.656 & 0.396 & 13.86 & 0.481 & 0.500 & 14.67 & 0.496 & 0.557 & 16.88 & 0.469 & 0.533 \\ 
        Kaufman\cite{kaufman2020deblurring} & 18.61 & 0.503 & 0.505 & 21.87 & 0.581 & 0.520 & 26.67 & 0.815 & 0.293 & 15.08 & 0.591 & 0.448 & 17.26 & 0.698 & 0.397 & 19.90 & 0.638 & 0.433 \\ 
        Cho\cite{cho2021rethinking} & 15.43 & 0.284 & 0.647 & 18.35 & 0.389 & 0.630 & 19.94 & 0.620 & 0.436 & 13.72 & 0.472 & 0.507 & 14.33 & 0.458 & 0.576 & 16.36 & 0.445 & 0.559 \\ 
        MPRNet\cite{zamir2021multi} & 15.52 & 0.295 & 0.634 & 18.54 & 0.405 & 0.595 & 19.99 & 0.623 & 0.430 & 13.73 & 0.475 & 0.495 & 12.89 & 0.398 & 0.589 & 16.13 & 0.439 & 0.548 \\ 
        Restormer\cite{zamir2022restormer} & 15.57 & 0.296 & 0.619 & 18.51 & 0.395 & 0.606 & 20.16 & 0.627 & 0.419 & 13.78 & 0.473 & 0.505 & 13.49 & 0.450 & 0.584 & 16.30 & 0.448 & 0.547 \\ 
        Zhang\cite{zhang2024cross} & 20.71 & 0.658 & 0.299 & 25.90 & 0.802 & 0.277 & 28.54 & 0.833 & 0.197 & 16.89 & 0.638 & 0.352 & 26.00 & 0.921 & 0.092 & 23.63 & 0.772 & 0.240 \\ 
        \hline\hline
        DIP\cite{ren2020neural} & 18.06 & 0.534 & 0.453 & 19.28 & 0.529 & 0.537 & 24.37 & 0.753 & 0.301 & 16.43 & 0.618 & 0.413 & 18.39 & 0.712 & 0.339 & 19.31 & 0.629 & 0.409 \\ 
        VDIP-Std\cite{huo2023blind} & 16.86 & 0.392 & 0.529 & 19.60 & 0.448 & 0.528 & 25.18 & 0.750 & 0.267 & 14.49 & 0.515 & 0.485 & 19.57 & 0.639 & 0.348 & 19.14 & 0.549 & 0.431 \\ 
        VDIP-Extreme\cite{huo2023blind} & 19.24 & 0.538 & 0.482 & 22.68 & 0.635 & 0.427 & 27.13 & 0.812 & 0.233 & 16.42 & 0.609 & 0.403 & 24.20 & 0.833 & 0.170 & 21.93 & 0.685 & 0.343 \\ 
        VDIP-Sparse\cite{huo2023blind} & 20.82 & 0.664 & 0.324 & 25.06 & 0.760 & 0.325 & 29.36 & 0.858 & 0.186 & 16.61 & 0.611 & 0.394 & 26.04 & 0.898 & 0.113 & 23.58 & 0.758 & 0.268 \\ 
        \hline\hline
        DIP-GLKM & \textbf{23.06} & \textbf{0.773} & \textbf{0.208} & \textbf{26.51} & \textbf{0.819} & \textbf{0.220} & \textbf{31.20} & \textbf{0.904} & \textbf{0.121} & 17.03 & 0.633 & 0.363 & 27.05 & 0.920 & 0.084 & \textbf{24.97} & \textbf{0.810} & \textbf{0.199} \\ 
        VDIP-Std-GLKM & \underline{22.74} & \underline{0.757} & \underline{0.215} & 26.07 & 0.797 & 0.240 & \underline{31.18} & \underline{0.903} & \underline{0.129} & 17.10 & 0.639 & 0.357 & \textbf{27.70} & \textbf{0.930} & \underline{0.065} & \underline{24.96} & \underline{0.805} & \underline{0.201} \\ 
        VDIP-Extreme-GLKM & 21.93 & 0.714 & 0.262 & \underline{26.25} & \underline{0.812} & \underline{0.227} & 29.01 & 0.837 & 0.214 & \underline{17.14} & \underline{0.643} & \underline{0.341} & \underline{27.59} & \underline{0.927} & \textbf{0.061} & 24.38 & 0.787 & 0.223 \\
        VDIP-Sparse-GLKM & 21.99 & 0.732 & 0.240 & 25.21 & 0.771 & 0.274 & 29.88 & 0.870 & 0.161 & \textbf{17.38} & \textbf{0.667} & \textbf{0.335} & 27.42 & 0.926 & 0.067 & 24.38 & 0.793 & 0.215 \\ 
        \Xhline{0.8pt}
    \end{tabular}
\end{table*}

%% file: table/real_lai.tex
\begin{table}[!t]
    \centering
    \caption{Quantitative comparisons of various methods on Lai et al.'s real dataset. The best
    and second best results are highlighted in \textbf{bold} and \underline{underline}, respectively.}
    \label{tab:real_lai}
    \scriptsize
    \begin{tabular}{@{}C{2.5cm}@{}|@{}C{1.4cm}@{}|@{}C{1.4cm}@{}|@{}C{1.4cm}@{}}
        \Xhline{0.8pt}
        Method & NIQE$\downarrow$ & BRISQUE$\downarrow$ & PIQE$\downarrow$ \\
        \hline
        Xu\cite{xu2013unnatural} & 4.9905 & 35.2359 & 35.1043 \\ 
        Dong\cite{dong2017blind} & 5.2532 & 39.7628 & 61.1299 \\ 
        Pan-DCP\cite{pan2017deblurring} & 5.9916 & 43.9319 & 65.3105 \\ 
        Pan-$l_0$\cite{pan2016l_0} & 5.6031 & 45.7093 & 66.5607 \\
        Kupyn\cite{kupyn2019deblurgan} & 4.7872 & 33.7445 & 38.6812 \\ 
        Kaufman\cite{kaufman2020deblurring} & 5.0278 & 39.3698 & 44.6920 \\ 
        Cho\cite{cho2021rethinking} & 5.3996 & 38.7433 & 46.4865 \\ 
        MPRNet\cite{zamir2021multi} & 5.6590 & 42.6111 & 50.8554 \\ 
        Restormer\cite{zamir2022restormer} & 5.7355 & 42.4397 & 49.1491 \\ 
        Zhang\cite{zhang2024cross} & 5.2151 & 29.6184 & 35.4604 \\ 
        \hline \hline
        DIP\cite{ren2020neural} & 4.8498 & 31.1910 & 34.0446 \\ 
        VDIP-Std\cite{huo2023blind} & 5.2058 & 34.0376 & 42.3929 \\ 
        VDIP-Extreme\cite{huo2023blind} & 5.3782 & 30.5701 & 31.7622 \\ 
        VDIP-Sparse\cite{huo2023blind} & 4.5591 & 31.4036 & 35.8809 \\ 
        \hline \hline 
        DIP-GLKM & \underline{4.5222} & \textbf{20.4111} & \textbf{22.5669} \\ 
        VDIP-Std-GLKM & 4.7659 & 25.7389 & 31.1532 \\ 
        VDIP-Extreme-GLKM & 4.8132 & 26.2979 & 31.1755 \\ 
        VDIP-Sparse-GLKM & \textbf{4.4231} & \underline{22.9516} & \underline{23.5313} \\ 
        \Xhline{0.8pt}
    \end{tabular}
\end{table}

%% file: fig/uniform_synthetic.tex
\begin{figure*}[!t]
    \setlength\tabcolsep{1pt}
	\renewcommand{\arraystretch}{0.618}
    \centering
    \scriptsize
    \begin{tabular}{cccccc}
        \includegraphics[width=0.16\textwidth]{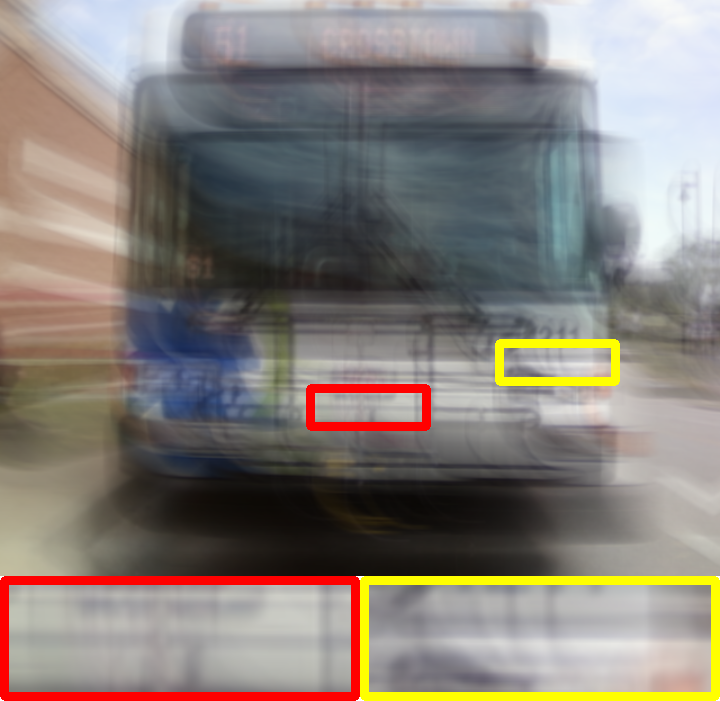}&
        \includegraphics[width=0.16\textwidth]{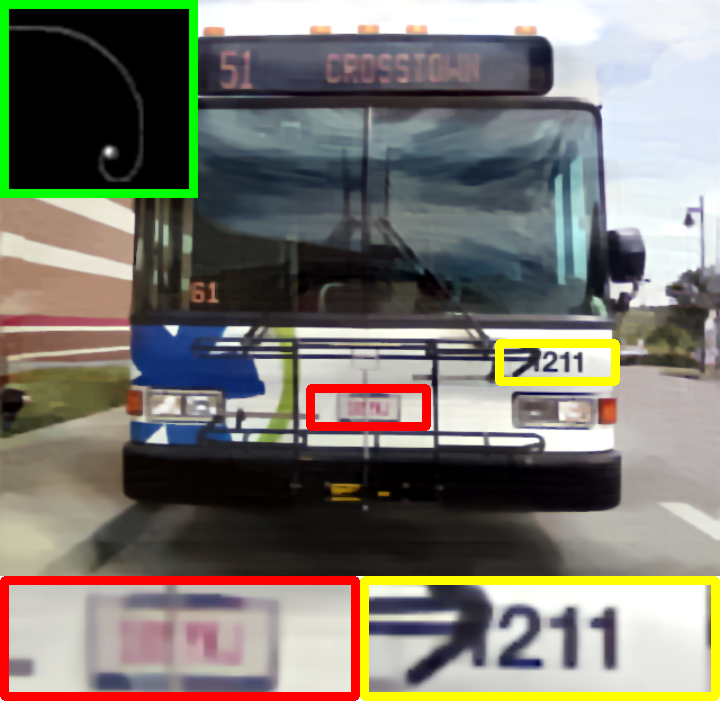}&
        \includegraphics[width=0.16\textwidth]{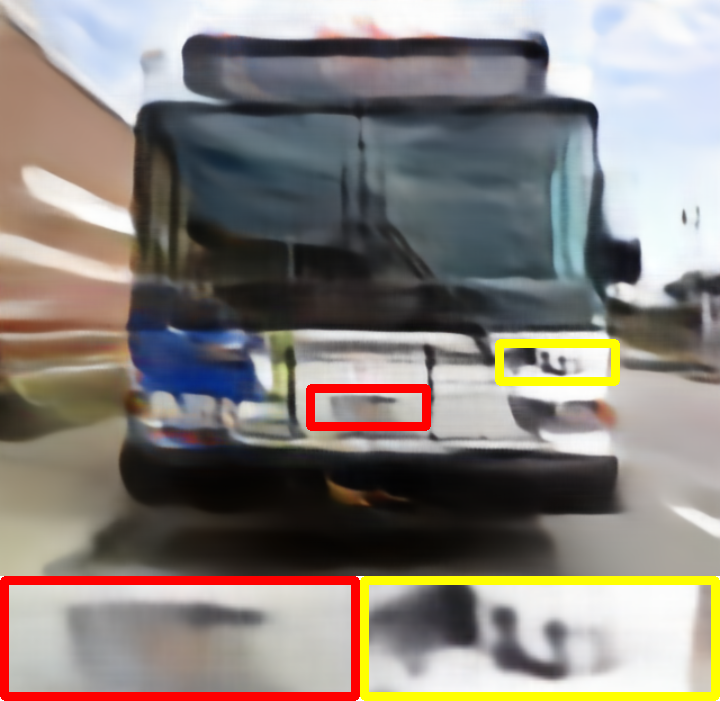}&
        \includegraphics[width=0.16\textwidth]{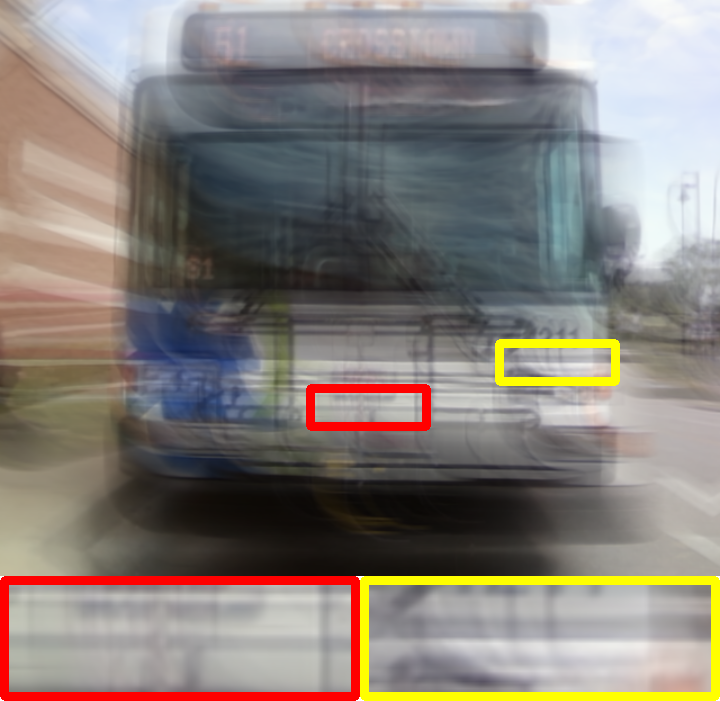}&
        \includegraphics[width=0.16\textwidth]{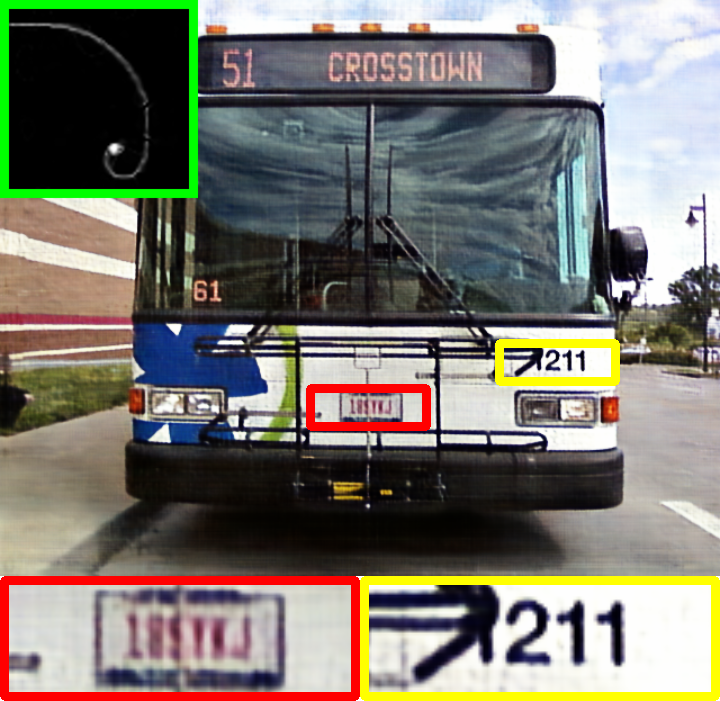}&
        \includegraphics[width=0.16\textwidth]{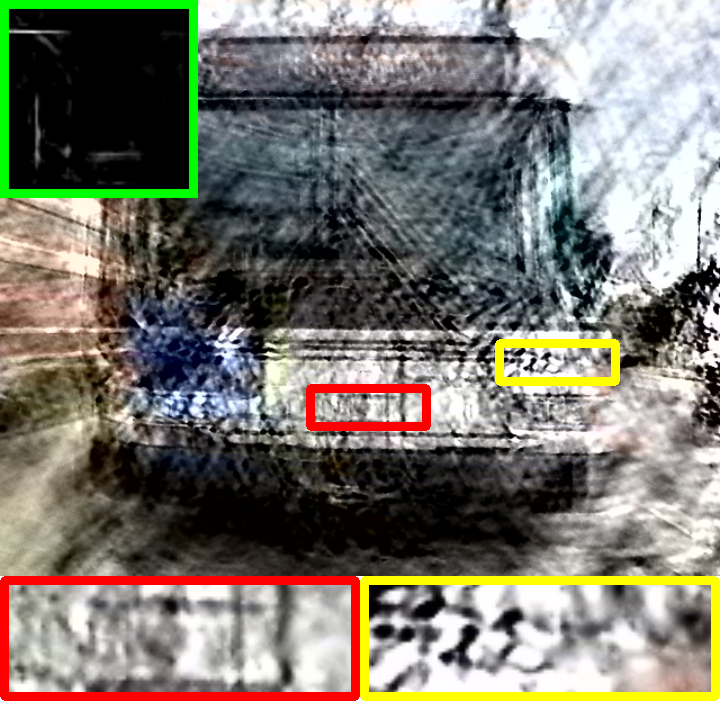}\\
        Blurred & Pan-DCP \cite{pan2017deblurring} & Kaufman \cite{kaufman2020deblurring} & MPRNet \cite{zamir2021multi} & Zhang\cite{zhang2024cross} & DIP\cite{ren2020neural}\\
        \includegraphics[width=0.16\textwidth]{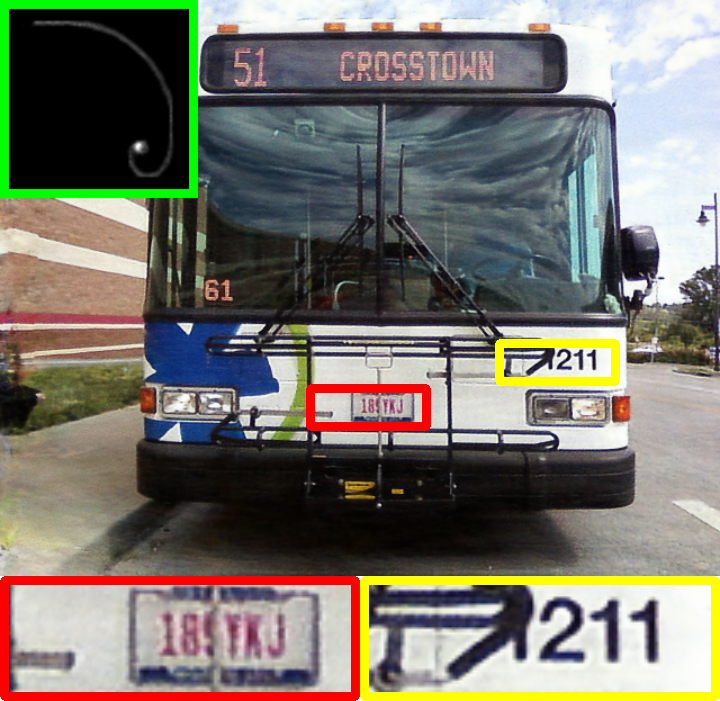}&
        \includegraphics[width=0.16\textwidth]{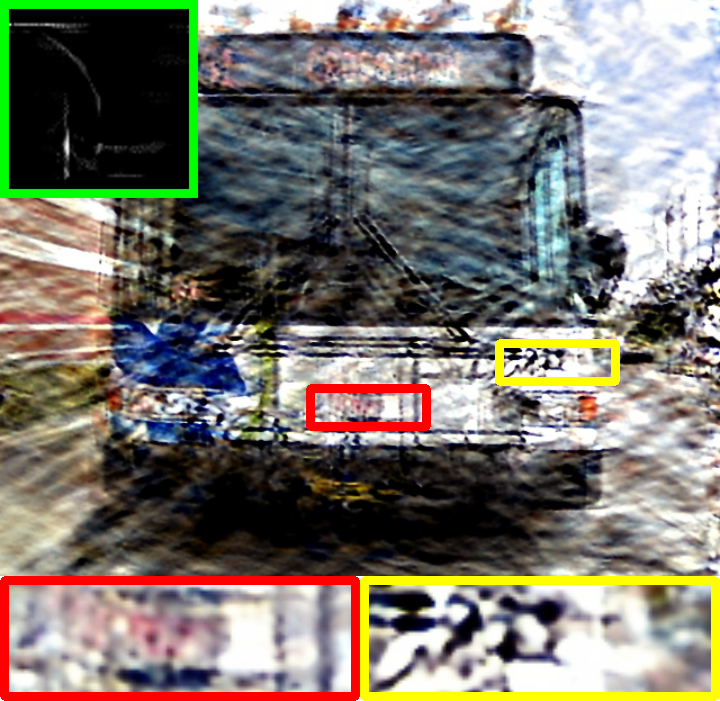}&
        \includegraphics[width=0.16\textwidth]{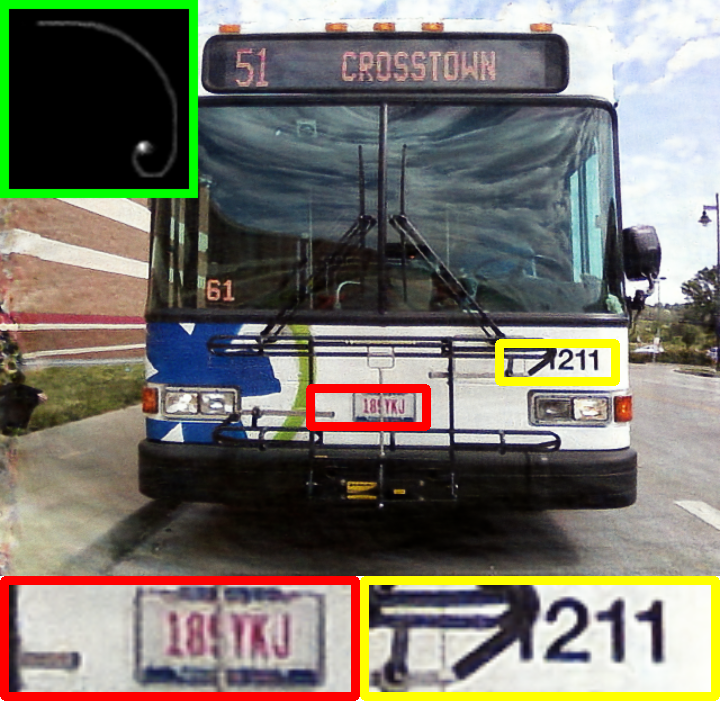}&
        \includegraphics[width=0.16\textwidth]{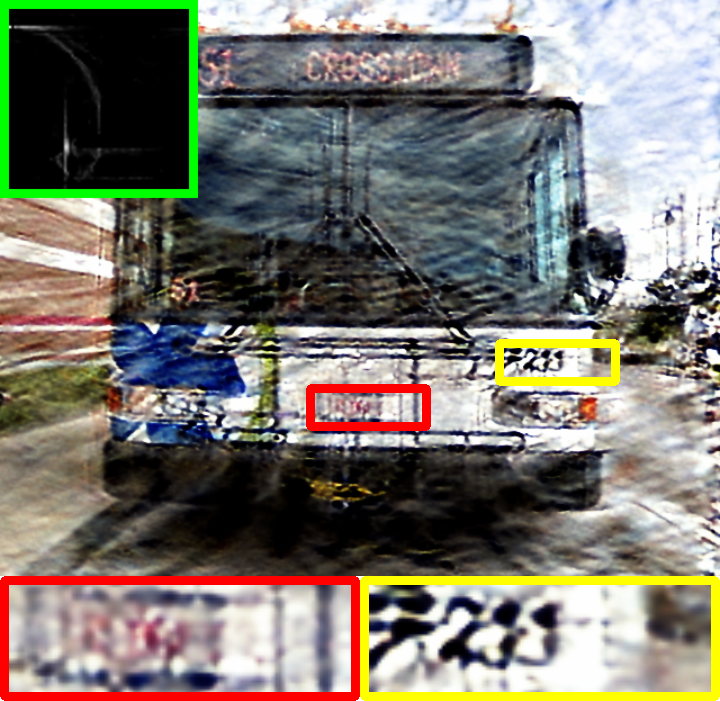}&
        \includegraphics[width=0.16\textwidth]{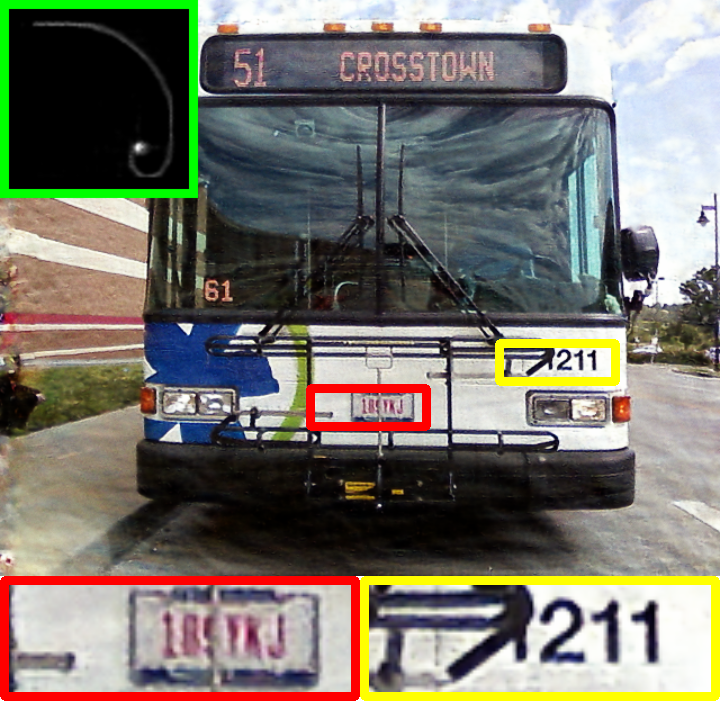}&
        \includegraphics[width=0.16\textwidth]{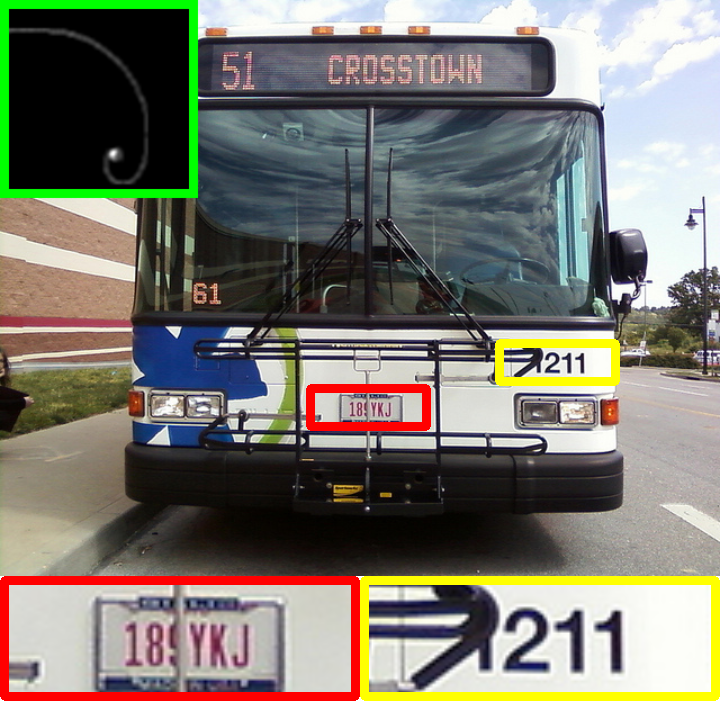}\\
        DIP-GLKM & VDIP-Std\cite{huo2023blind} & VDIP-Std-GLKM & VDIP-Sparse\cite{huo2023blind} & VDIP-Sparse-GLKM & Ground truth
    \end{tabular}
    \caption{Visual comparisons on our synthetic dataset. The estimated blur kernels are pasted at the top-left corners if available. More comparisons can be found in the supplementary material.}
    \label{fig:uniform_synthetic}
\end{figure*}

%% file: fig/uniform_lai.tex
\begin{figure*}[!t]
    \setlength\tabcolsep{1pt}
	\renewcommand{\arraystretch}{0.618}
    \centering
    \scriptsize
    \begin{tabular}{ccccccc}
        \includegraphics[width=0.16\textwidth]{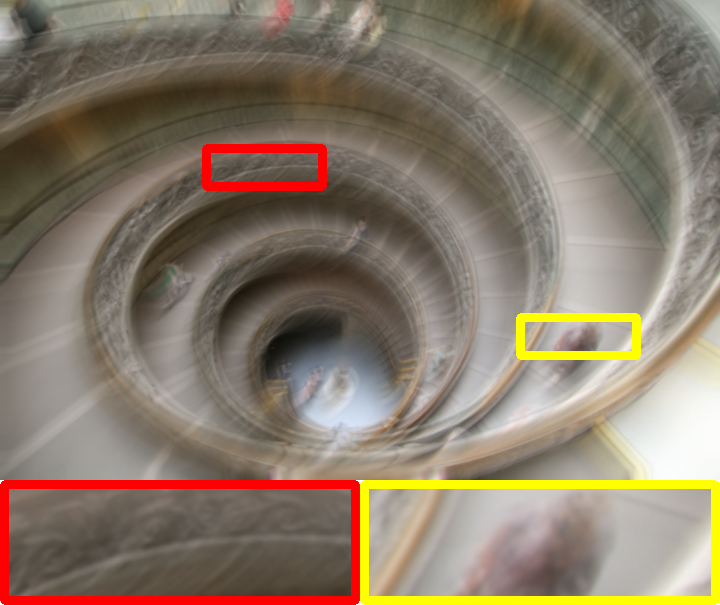}&
        \includegraphics[width=0.16\textwidth]{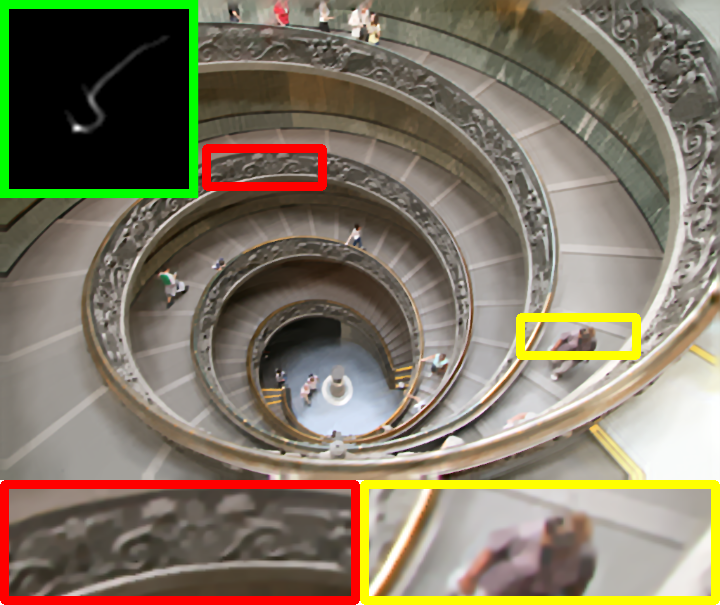}&
        \includegraphics[width=0.16\textwidth]{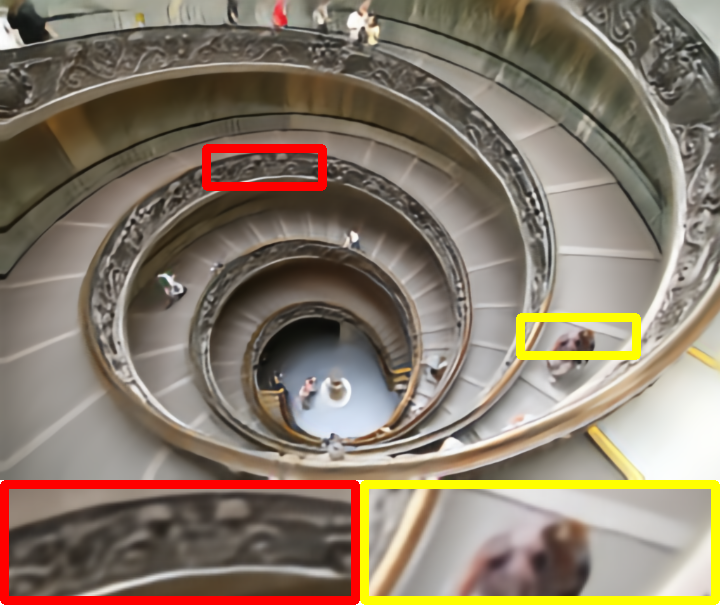}&
        \includegraphics[width=0.16\textwidth]{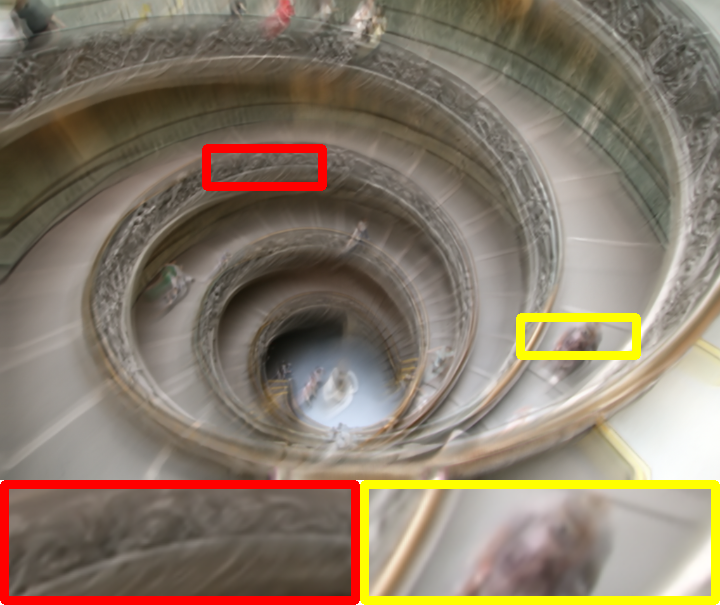}&
        \includegraphics[width=0.16\textwidth]{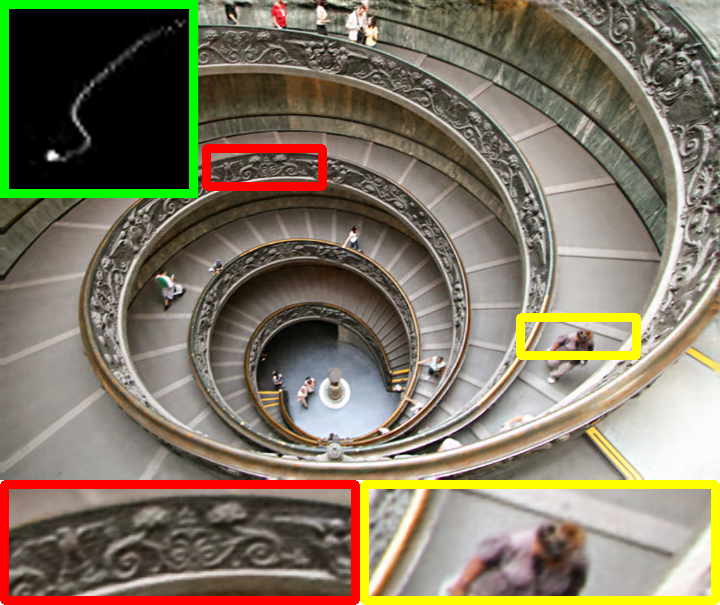}&
        \includegraphics[width=0.16\textwidth]{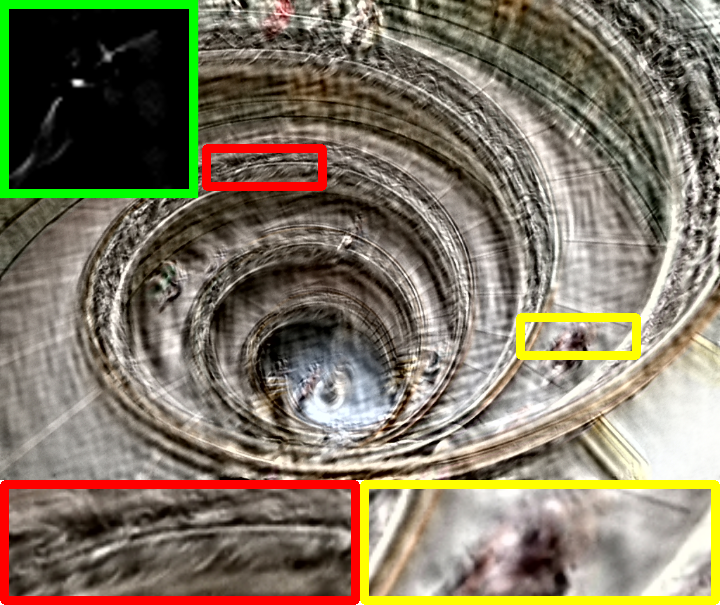}\\
        Blurred & Pan-DCP \cite{pan2017deblurring} & Kaufman \cite{kaufman2020deblurring} & MPRNet \cite{zamir2021multi} & Zhang\cite{zhang2024cross} & DIP\cite{ren2020neural}\\
        \includegraphics[width=0.16\textwidth]{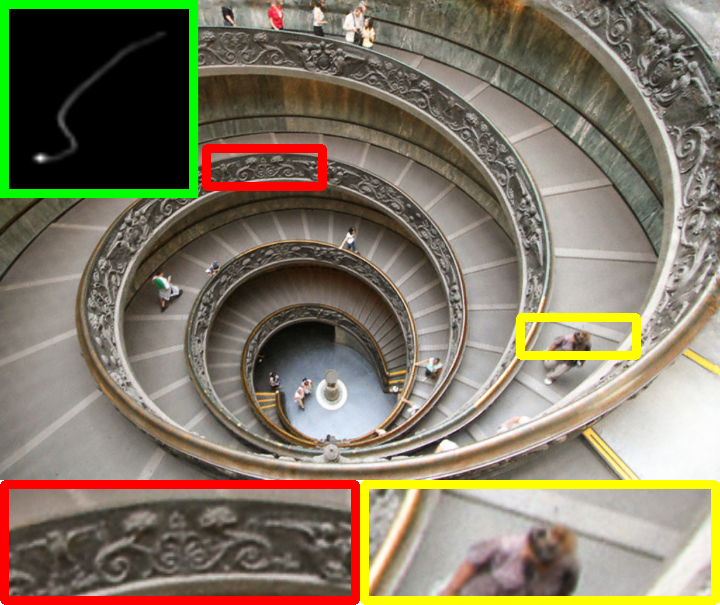}&
        \includegraphics[width=0.16\textwidth]{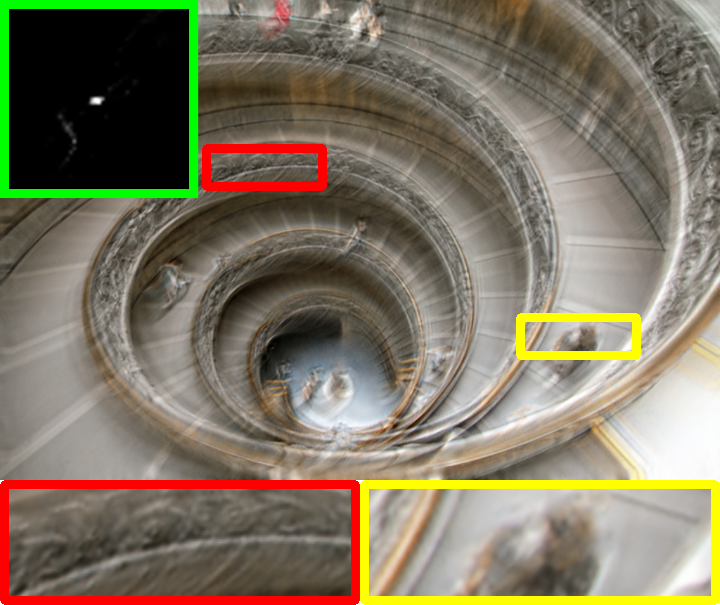}&
        \includegraphics[width=0.16\textwidth]{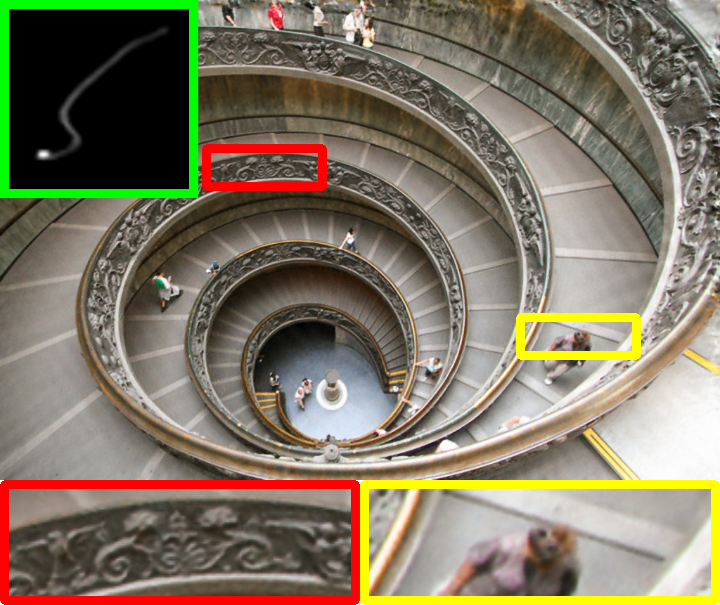}&
        \includegraphics[width=0.16\textwidth]{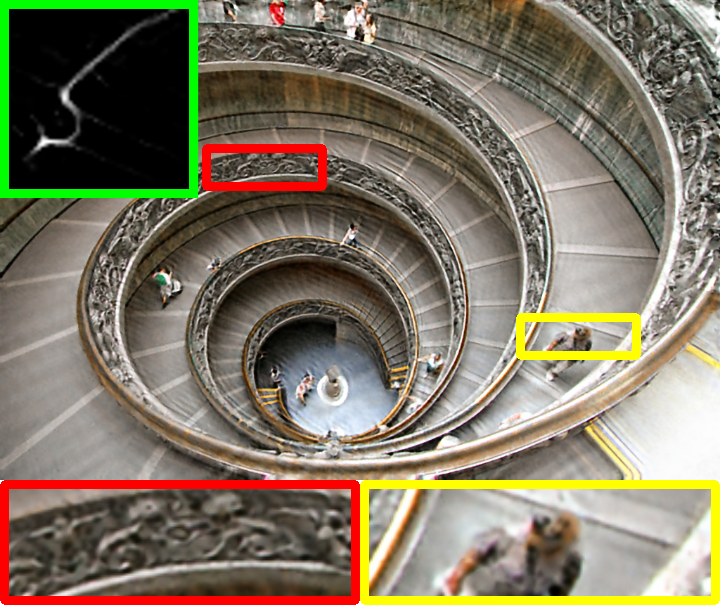}&
        \includegraphics[width=0.16\textwidth]{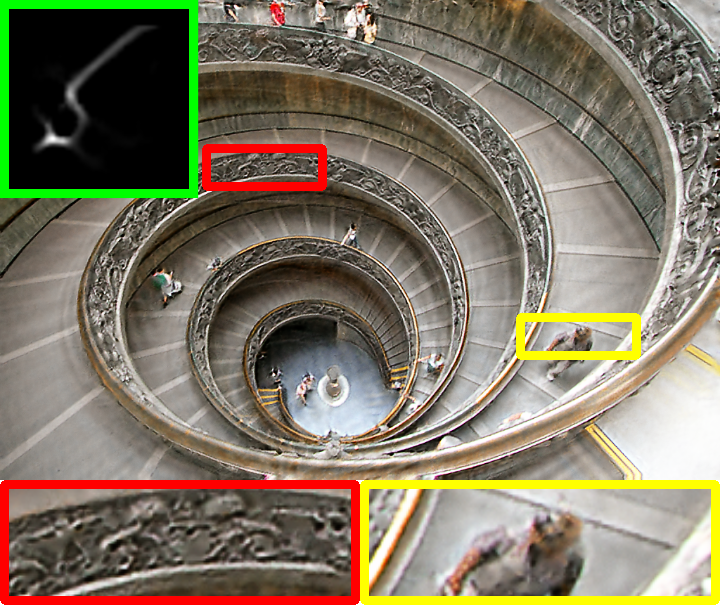}&
        \includegraphics[width=0.16\textwidth]{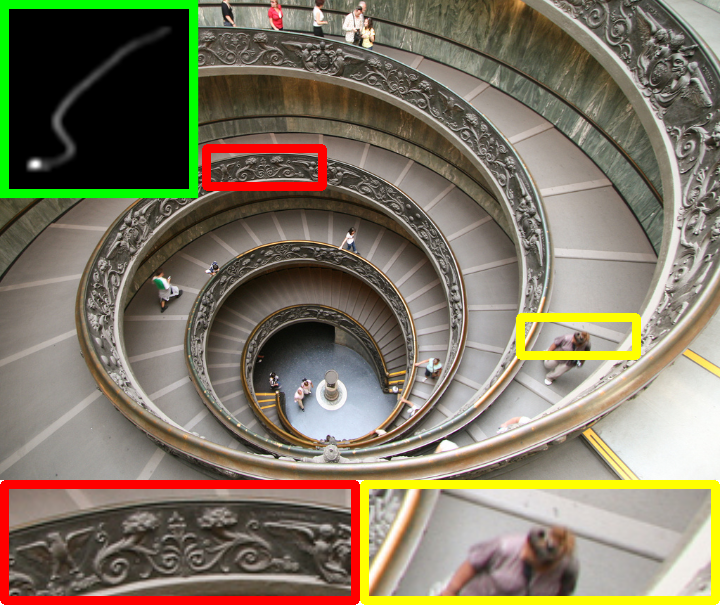}\\
        DIP-GLKM & VDIP-Std\cite{huo2023blind} & VDIP-Std-GLKM & VDIP-Sparse\cite{huo2023blind} & VDIP-Sparse-GLKM & Ground truth
    \end{tabular}
    \caption{Visual comparisons on Lai \MakeLowercase{\textit{et al.}}'s synthetic dataset. The estimated blur kernels are pasted at the top-left corners if available. More comparisons can be found in the supplementary material.}
    \label{fig:uniform_lai}
\end{figure*}

%% file: fig/real_lai.tex
\begin{figure*}[!t]
    \setlength\tabcolsep{1pt}
	\renewcommand{\arraystretch}{0.618}
    \centering
    \scriptsize
    \begin{tabular}{cccccc}
        \includegraphics[width=0.16\textwidth]{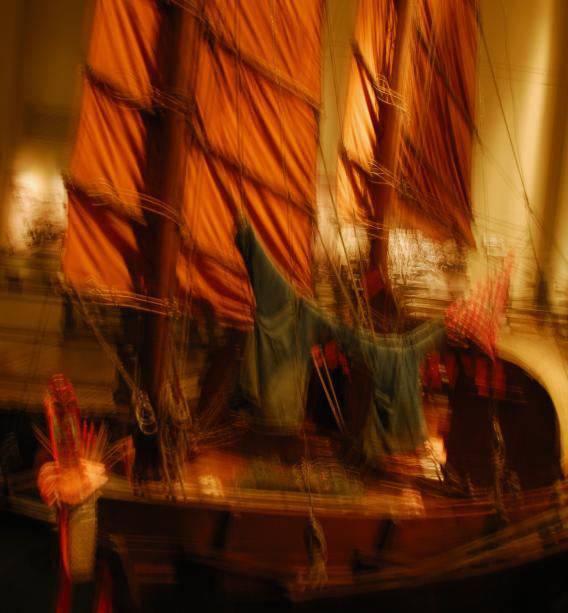}&
        \includegraphics[width=0.16\textwidth]{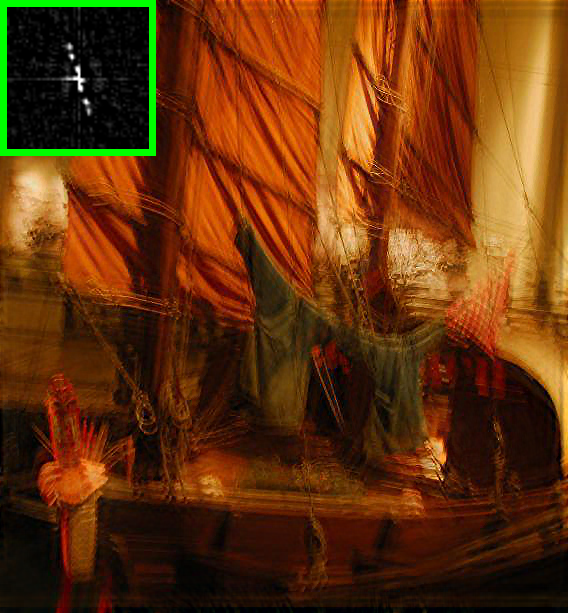}&
        \includegraphics[width=0.16\textwidth]{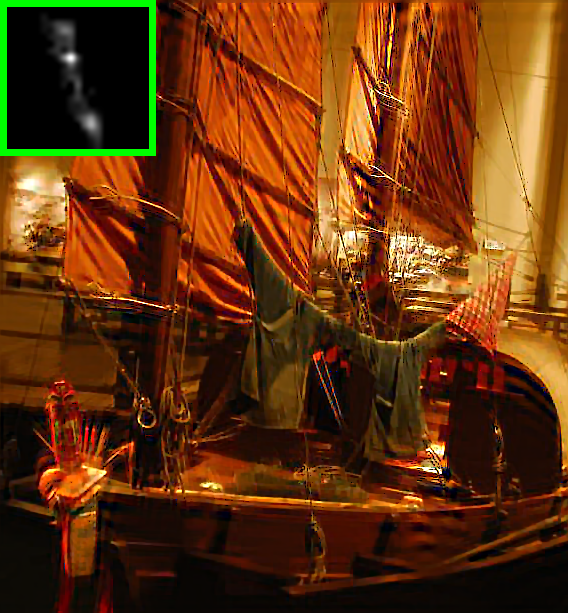}&
        \includegraphics[width=0.16\textwidth]{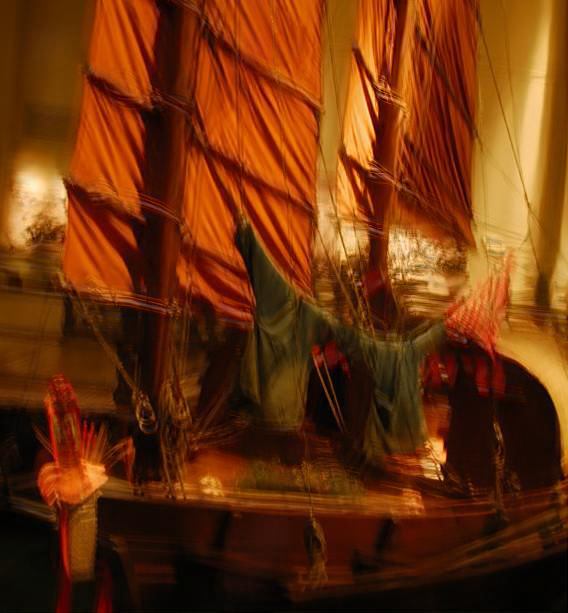}&
        \includegraphics[width=0.16\textwidth]{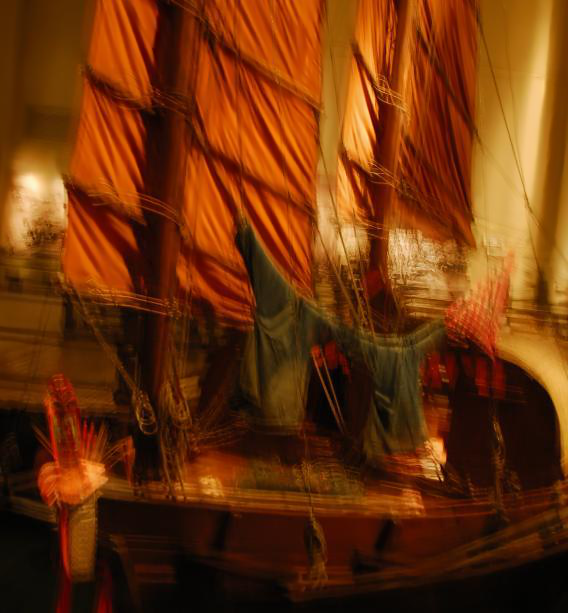}&
        \includegraphics[width=0.16\textwidth]{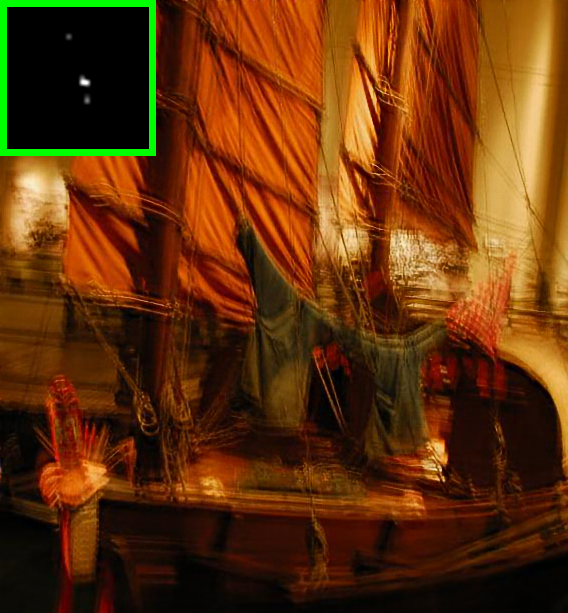}\\
        Blurred&Xu&Pan-DCP\cite{pan2017deblurring}&Kupyn\cite{kupyn2019deblurgan}&MPRNet\cite{zamir2021multi}&Zhang\cite{zhang2024cross}\\
        \includegraphics[width=0.16\textwidth]{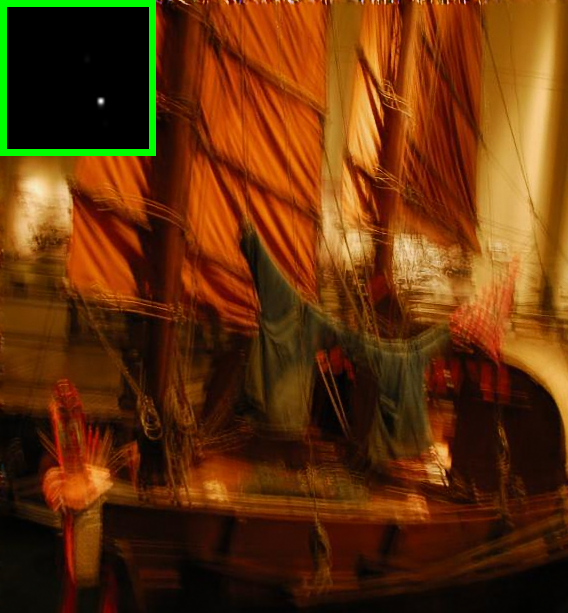}&
        \includegraphics[width=0.16\textwidth]{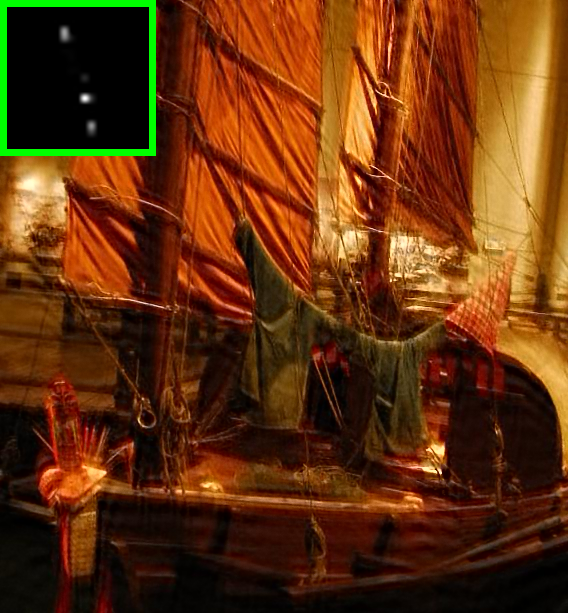}&
        \includegraphics[width=0.16\textwidth]{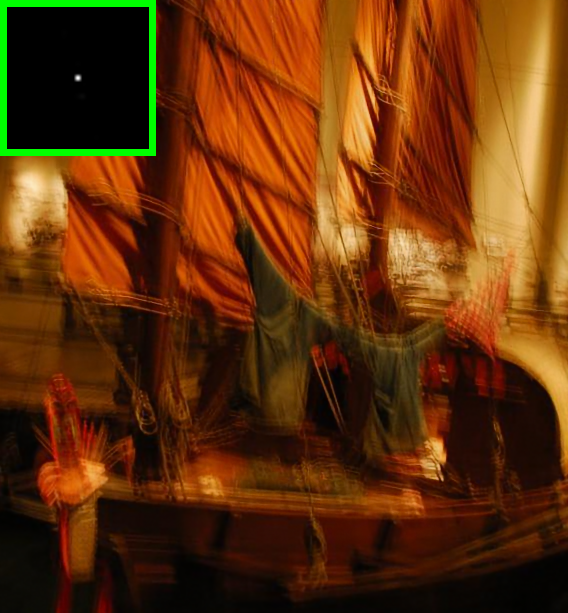}&
        \includegraphics[width=0.16\textwidth]{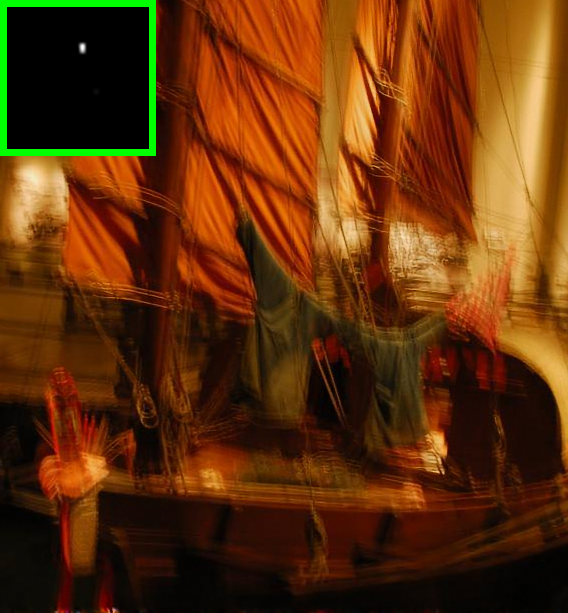}&
        \includegraphics[width=0.16\textwidth]{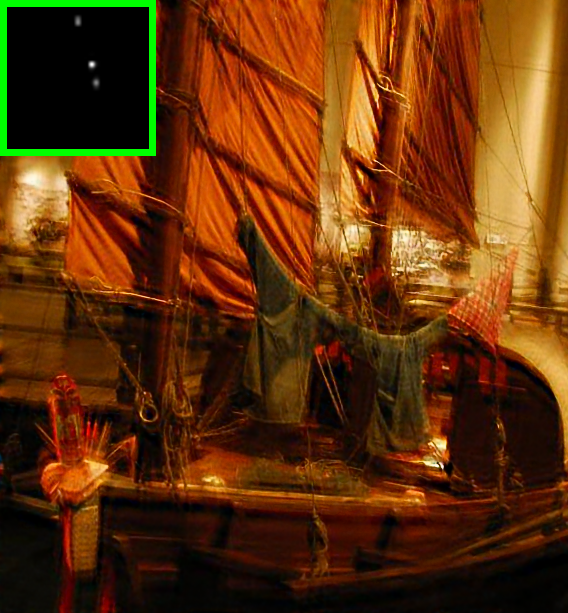}&
        \includegraphics[width=0.16\textwidth]{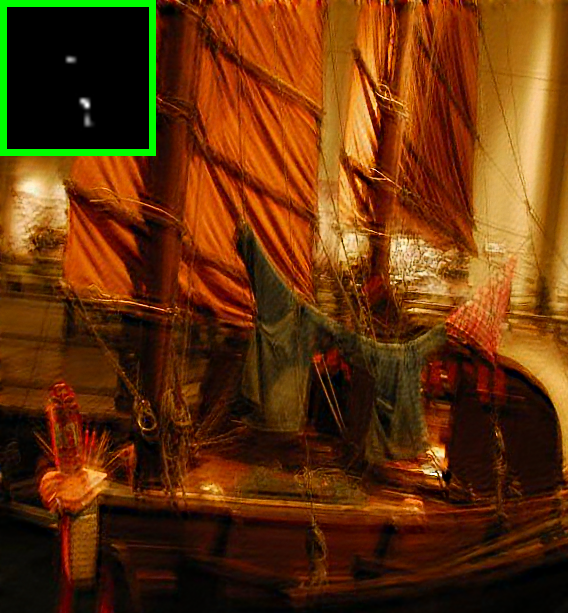}\\
        DIP\cite{ren2020neural}&DIP-GLKM&VDIP-Std\cite{huo2023blind}&VDIP-Std-GLKM&VDIP-Sparse\cite{huo2023blind}&VDIP-Sparse-GLKM
    \end{tabular}
    \caption{Visual results on Lai \MakeLowercase{\textit{et al.}}'s real dataset. The estimated blur kernels are pasted at the top-left corners if available. More comparisons can be found in the supplementary material.}
    \label{fig_real}
\end{figure*}

%% file: table/uniform_diffusion.tex
\begin{table*}[!t]
    \centering
    \caption{Quantitative results of various methods on AFHQ-Dog\cite{choi2020stargan}, CelebAHQ\cite{liu2018large}, and ImageNet\cite{deng2009imagenet}. The best
    and second best results are highlighted in \textbf{bold} and \underline{underline}, respectively.}
    \label{tab:uniform_diffusion}
    \scriptsize
    \begin{tabular}{@{}C{2cm}@{}|@{}C{1cm}@{}|@{}C{1cm}@{}|@{}C{1cm}@{}|@{}C{1cm}@{}|@{}C{1cm}@{}|@{}C{1cm}@{}|@{}C{1cm}@{}|@{}C{1cm}@{}|@{}C{1cm}@{}}
        \Xhline{0.8pt}
        \multirow{2}{*}{Method} & \multicolumn{3}{c|}{\textbf{AFHQ-Dog}\cite{choi2020stargan}} & \multicolumn{3}{c|}{\textbf{CelebAHQ}\cite{liu2018large}} & \multicolumn{3}{c}{\textbf{ImageNet}\cite{deng2009imagenet}} \\
        \cline{2-10}
        & PSNR$\uparrow$ & SSIM$\uparrow$ & LPIPS$\downarrow$ & PSNR$\uparrow$ & SSIM$\uparrow$ & LPIPS$\downarrow$ & PSNR$\uparrow$ & SSIM$\uparrow$ & LPIPS$\downarrow$ \\
        \hline
        Pan-DCP\cite{pan2017deblurring} & 25.66 & 0.730 & 0.322 & 25.81 & 0.753 & 0.263 & 23.22 & 0.683 & 0.382 \\
        Kupyn\cite{kupyn2019deblurgan} & 23.83 & 0.668 & 0.356 & 25.07&0.732&0.273&22.52&0.640&0.407\\
        MPRNet\cite{zamir2021multi} & 21.96 & 0.601 & 0.374 & 23.94 & 0.663 & 0.259 & 21.96 & 0.620 & 0.409 \\
        Restormer\cite{zamir2022restormer}&22.13	&0.612	&0.356&25.73&	0.735&	0.237&22.13	&0.622&0.409\\
        GibbsDDRM\cite{murata2023gibbsddrm}&21.99 &0.601 &0.244 & 21.77&0.587 &0.272 &17.68 & 0.418&0.510 \\
        \hline \hline
        DIP\cite{ren2020neural} & 18.09 & 0.381 & 0.393 & 18.81 & 0.437 & 0.340 & 16.59 & 0.363 & 0.443 \\
        BIRD\cite{chihaoui2024blind} & 16.68 & 0.423 & 0.337 & 16.85 & 0.442 & 0.325 & 14.90 & 0.349 & 0.618 \\
        BlindDPS\cite{chung2023parallel} & 25.68 & 0.730 & 0.133 & 25.58 & 0.752 & 0.130 & 22.41 & 0.631 & 0.326 \\
        \hline \hline
        DIP-GLKM & \textbf{31.59} & \textbf{0.885} & \textbf{0.095} & \textbf{32.45} & \textbf{0.920} & \textbf{0.078} & \textbf{29.93} & \textbf{0.869} & \textbf{0.123} \\
        BIRD-GLKM & 26.60 & 0.755 & 0.181 & 26.04 & 0.750 & 0.159 & 22.64 & 0.643 & 0.354 \\
        BlindDPS-GLKM & \underline{26.92} & \underline{0.774} & \underline{0.118} & \underline{26.88} & \underline{0.794} & \underline{0.112} & \underline{23.65} & \underline{0.687} & \underline{0.303} \\
        \Xhline{0.8pt}
    \end{tabular}
\end{table*}

%% file: fig/uniform_diffusion.tex
\begin{figure*}[!t]
    \setlength\tabcolsep{1pt}
	\renewcommand{\arraystretch}{0.618}
    \centering
    \scriptsize
    \begin{tabular}{cccccccc}
        \includegraphics[width=0.12\textwidth]{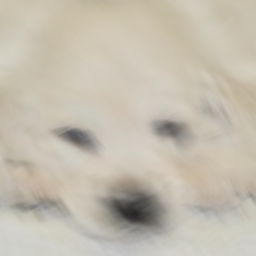}&
        \includegraphics[width=0.12\textwidth]{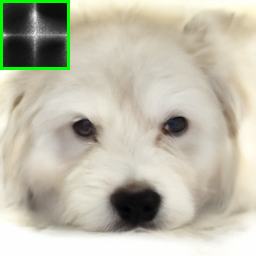}&
        \includegraphics[width=0.12\textwidth]{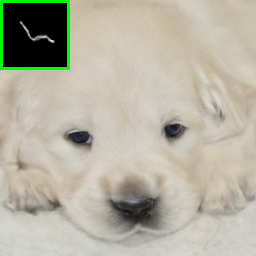}&
        \includegraphics[width=0.12\textwidth]{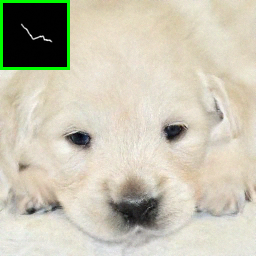}&
        \includegraphics[width=0.12\textwidth]{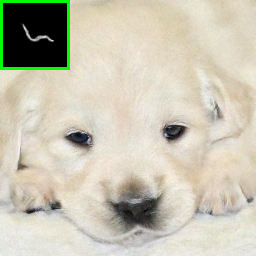}&
        \includegraphics[width=0.12\textwidth]{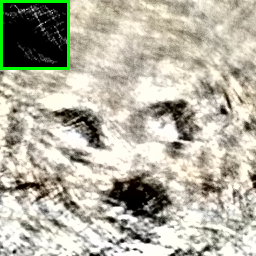}&
        \includegraphics[width=0.12\textwidth]{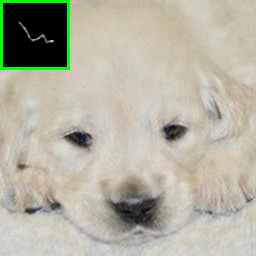}&
        \includegraphics[width=0.12\textwidth]{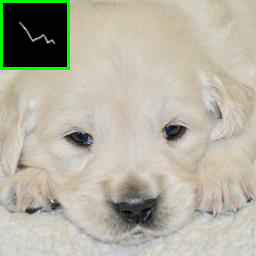}\\
        \includegraphics[width=0.12\textwidth]{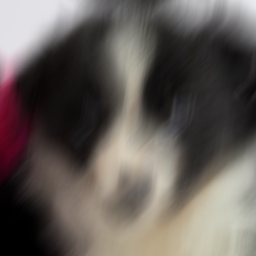}&
        \includegraphics[width=0.12\textwidth]{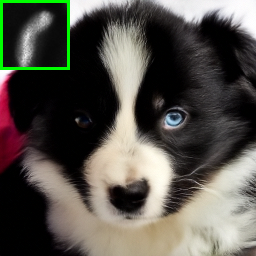}&
        \includegraphics[width=0.12\textwidth]{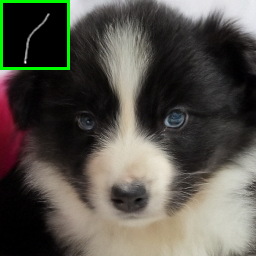}&
        \includegraphics[width=0.12\textwidth]{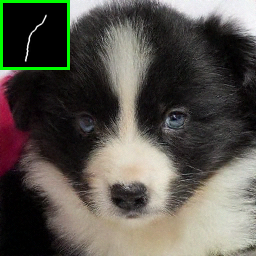}&
        \includegraphics[width=0.12\textwidth]{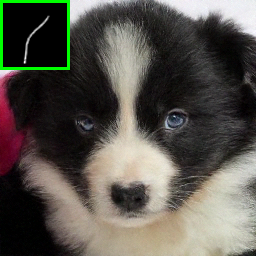}&
        \includegraphics[width=0.12\textwidth]{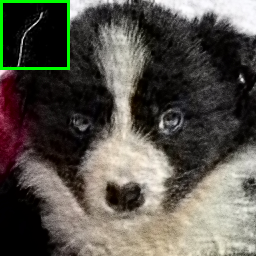}&
        \includegraphics[width=0.12\textwidth]{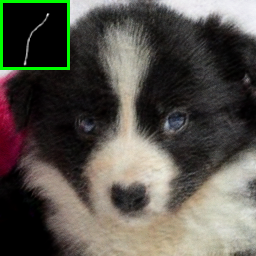}&
        \includegraphics[width=0.12\textwidth]{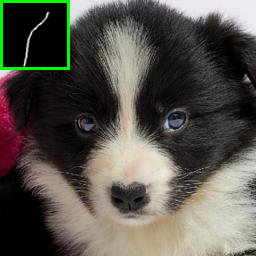}\\
        Blurred&BIRD\cite{chihaoui2024blind}&BIRD-GLKM&BlindDPS\cite{chung2023parallel}&BlindDPS-GLKM&DIP\cite{ren2020neural}&DIP-GLKM&Ground truth   
    \end{tabular}
    \caption{Visual comparison of different BMD methods on AFHQ-Dog~\cite{choi2020stargan} dataset. The estimated blur kernels are pasted at the top-left corners if available.}
    \label{fig:afhq}
\end{figure*}
\begin{figure*}[!t]
    \setlength\tabcolsep{1pt}
	\renewcommand{\arraystretch}{0.618}
    \centering
    \scriptsize
    \begin{tabular}{cccccccc}
        \includegraphics[width=0.12\textwidth]{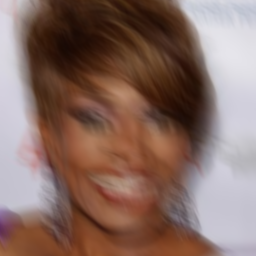}&
        \includegraphics[width=0.12\textwidth]{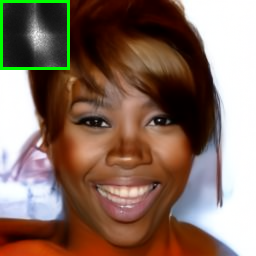}&
        \includegraphics[width=0.12\textwidth]{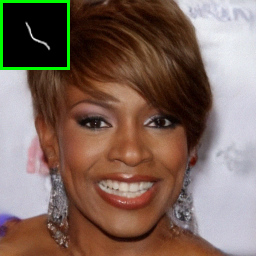}&
        \includegraphics[width=0.12\textwidth]{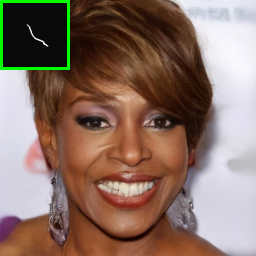}&
        \includegraphics[width=0.12\textwidth]{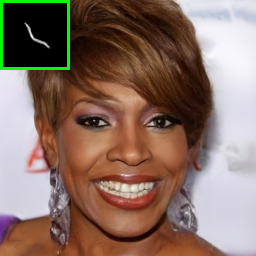}&
        \includegraphics[width=0.12\textwidth]{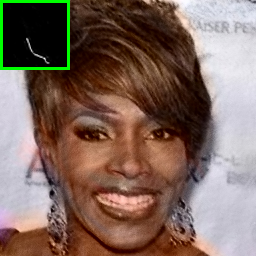}&
        \includegraphics[width=0.12\textwidth]{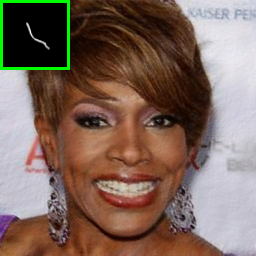}&
        \includegraphics[width=0.12\textwidth]{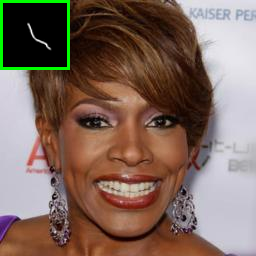}\\
        \includegraphics[width=0.12\textwidth]{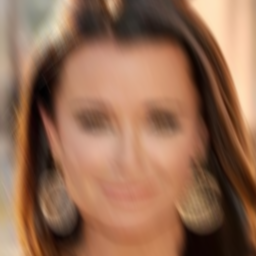}&
        \includegraphics[width=0.12\textwidth]{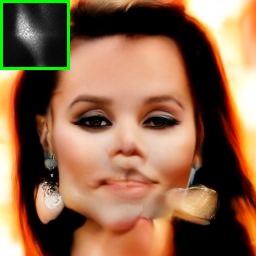}&
        \includegraphics[width=0.12\textwidth]{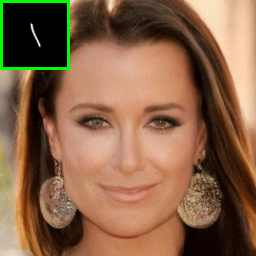}&
        \includegraphics[width=0.12\textwidth]{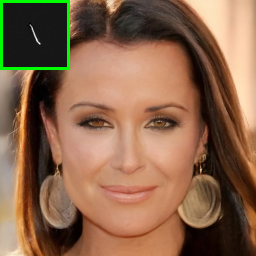}&
        \includegraphics[width=0.12\textwidth]{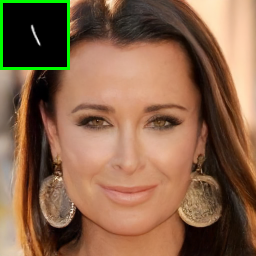}&
        \includegraphics[width=0.12\textwidth]{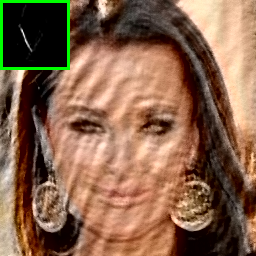}&
        \includegraphics[width=0.12\textwidth]{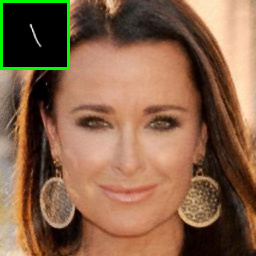}&
        \includegraphics[width=0.12\textwidth]{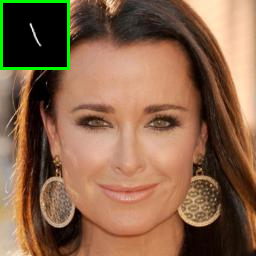}\\
        Blurred&BIRD\cite{chihaoui2024blind}&BIRD-GLKM&BlindDPS\cite{chung2023parallel}&BlindDPS-GLKM&DIP\cite{ren2020neural}&DIP-GLKM&Ground truth
    \end{tabular}
    \caption{Visual comparison of different BMD methods on CelebAHQ~\cite{liu2018large}dataset. The estimated blur kernels are pasted at the top-left corners if available.}
    \label{fig:celebahq}
\end{figure*}

\begin{figure*}[!t]
    \setlength\tabcolsep{1pt}
	\renewcommand{\arraystretch}{0.618}
    \centering
    \scriptsize
    \begin{tabular}{cccccccc}
        \includegraphics[width=0.12\textwidth]{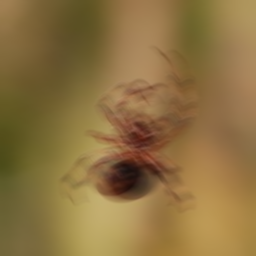}&
        \includegraphics[width=0.12\textwidth]{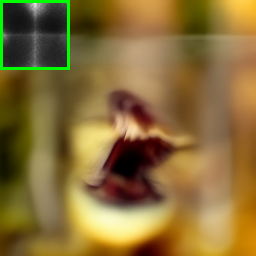}&
        \includegraphics[width=0.12\textwidth]{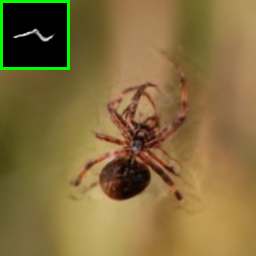}&
        \includegraphics[width=0.12\textwidth]{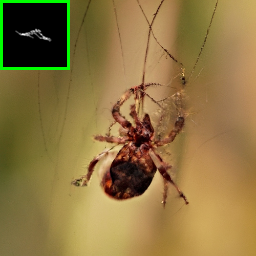}&
        \includegraphics[width=0.12\textwidth]{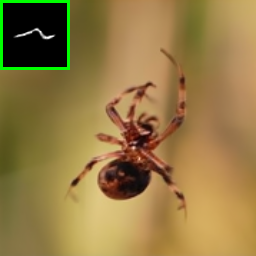}&
        \includegraphics[width=0.12\textwidth]{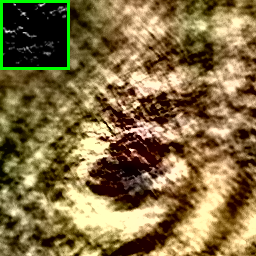}&
        \includegraphics[width=0.12\textwidth]{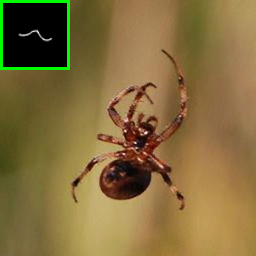}&
        \includegraphics[width=0.12\textwidth]{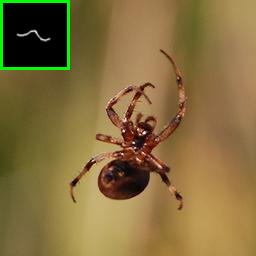}\\
        \includegraphics[width=0.12\textwidth]{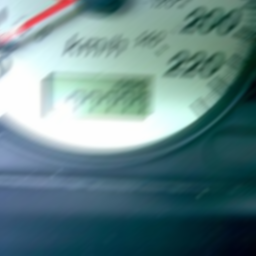}&
        \includegraphics[width=0.12\textwidth]{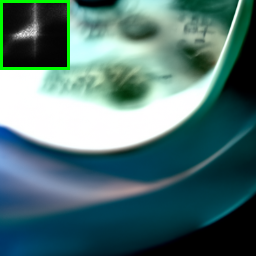}&
        \includegraphics[width=0.12\textwidth]{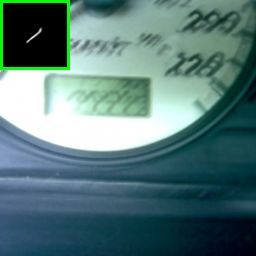}&
        \includegraphics[width=0.12\textwidth]{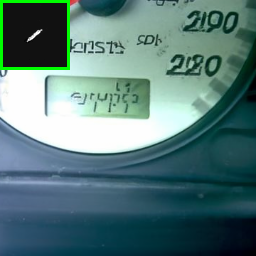}&
        \includegraphics[width=0.12\textwidth]{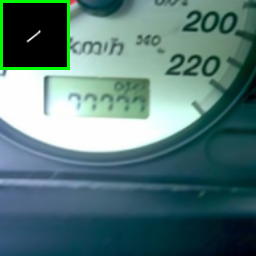}&
        \includegraphics[width=0.12\textwidth]{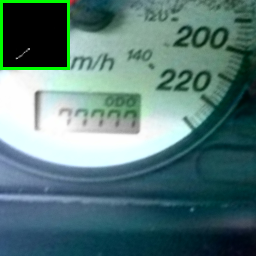}&
        \includegraphics[width=0.12\textwidth]{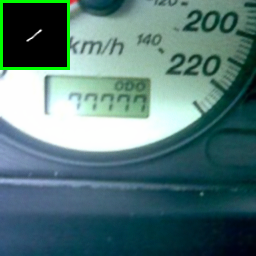}&
        \includegraphics[width=0.12\textwidth]{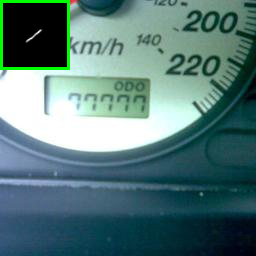}\\
        Blurred&BIRD\cite{chihaoui2024blind}&BIRD-GLKM&BlindDPS\cite{chung2023parallel}&BlindDPS-GLKM&DIP\cite{ren2020neural}&DIP-GLKM&Ground truth
    \end{tabular}
    \caption{Visual comparison of different BMD methods on Imagenet~\cite{deng2009imagenet} dataset. The estimated blur kernels are pasted at the top-left corners if available.}
    \label{fig:imagenet}
\end{figure*}

%% file: table/nonuniform_synthetic.tex

\begin{table}[!tbp]
    \centering
    \caption{Quantitative comparisons of various methods on our non-uniform synthetic dataset. The best
    and second best results are highlighted in \textbf{bold} and \underline{underline}, respectively.}
    \label{tab:nonuniform_synthetic}
    \scriptsize
    \begin{tabular}{@{}C{2.5cm}@{}|@{}C{1.5cm}@{}|@{}C{1.5cm}@{}|@{}C{1.5cm}@{}}
        \Xhline{0.8pt}
        Method & PSNR$\uparrow$ & SSIM$\uparrow$ & LPIPS$\downarrow$ \\
        \hline
        Xu\cite{xu2013unnatural} & 23.05 & 0.625 & 0.334 \\ 
        Whyte\cite{whyte2012non} & 22.88 & 0.614 & 0.310\\
        Kupyn\cite{kupyn2019deblurgan} & 24.75 & 0.698 & 0.313 \\ 
        Cho\cite{cho2021rethinking} & 23.75 & 0.648 & 0.392 \\ 
        MPRNet\cite{zamir2021multi} & 25.05 & 0.724 & 0.328 \\ 
        Restormer\cite{zamir2022restormer} & 25.18 & 0.718 & 0.328 \\ 
        Zhang\cite{zhang2023event} & 23.36 & 0.642 & 0.433 \\ 
        Fang\cite{fang2023self} & 25.54 & 0.738 & 0.325 \\ 
        Li\cite{li2023self} & 24.88 & 0.709 & 0.303 \\
        \hline \hline
        DIP\cite{ren2020neural} & 12.32 & 0.124 & 0.702 \\ 
        VDIP-Std\cite{huo2023blind} & 14.11 & 0.256 & 0.544 \\ 
        VDIP-Extreme\cite{huo2023blind} & 14.30 & 0.253 & 0.538 \\ 
        VDIP-Sparse\cite{huo2023blind} & 13.76 & 0.234 & 0.566 \\ 
        \hline \hline
        DIP-GLKM & \textbf{27.02} & \textbf{0.807} & \underline{0.264} \\ 
        VDIP-Std-GLKM & \underline{26.04} & \underline{0.769} & \textbf{0.222} \\ 
        VDIP-Extreme-GLKM & 25.32 & 0.370 & 0.287 \\ 
        VDIP-Sparse-GLKM & 23.35 & 0.584 & 0.287 \\ 
        \Xhline{0.8pt}
    \end{tabular}
\end{table}

%% file: table/nonuniform_lai.tex
\begin{table*}[!tbp]
    \centering
    \caption{Quantitative results of various methods on Lai's non-uniform blur dataset. The best
    and second best results are highlighted in \textbf{bold} and \underline{underline}, respectively.}
    \label{tab:nonuniform_lai}
    \scriptsize
    \begin{tabular}{@{}C{2.4cm}@{}|@{}C{0.85cm}@{}|@{}C{0.85cm}@{}|@{}C{0.85cm}@{}|@{}C{0.85cm}@{}|@{}C{0.85cm}@{}|@{}C{0.85cm}@{}|@{}C{0.85cm}@{}|@{}C{0.85cm}@{}|@{}C{0.85cm}@{}|@{}C{0.85cm}@{}|@{}C{0.85cm}@{}|@{}C{0.85cm}@{}|@{}C{0.85cm}@{}|@{}C{0.85cm}@{}|@{}C{0.85cm}@{}|@{}C{0.85cm}@{}|@{}C{0.85cm}@{}|@{}C{0.85cm}@{}}
        \Xhline{0.8pt}
        \multirow{2}{*}{Method} & \multicolumn{3}{c|}{Manmade} & \multicolumn{3}{c|}{Natural} & \multicolumn{3}{c|}{People} & \multicolumn{3}{c|}{Saturated} & \multicolumn{3}{c|}{Text} & \multicolumn{3}{c@{}}{Average} \\ 
        \cline{2-19}
        & PSNR$\uparrow$ & SSIM$\uparrow$ & LPIPS$\downarrow$ & PSNR$\uparrow$ & SSIM$\uparrow$ & LPIPS$\downarrow$ & PSNR$\uparrow$ & SSIM$\uparrow$ & LPIPS$\downarrow$ & PSNR$\uparrow$ & SSIM$\uparrow$ & LPIPS$\downarrow$ & PSNR$\uparrow$ & SSIM$\uparrow$ & LPIPS$\downarrow$ & PSNR$\uparrow$ & SSIM$\uparrow$ & LPIPS$\downarrow$ \\
        \hline
        Xu\cite{xu2013unnatural} & 17.37 & 0.408 & 0.399 & 20.77 & 0.517 & 0.388 & 23.71 & 0.726 & 0.257 & 16.97 & 0.557 & 0.339 & 17.69 & 0.673 & 0.312 & 19.30 & 0.576 & 0.339  \\ 
        Whyte\cite{whyte2012non} & 17.29 & 0.417 & 0.358 & 20.93 & 0.521 & 0.356 & 23.59 & 0.716 & 0.237 & 16.51 & 0.539 & 0.319 & 17.00 & 0.634 & 0.238 & 19.11 & 0.564 & 0.303  \\ 
        Vasu\cite{vasu2017local} & 17.93 & 0.477 & 0.383 & 21.94 & 0.602 & 0.374 & 25.63 & 0.795 & 0.225 & 17.57 & 0.617 & \textbf{0.276} & 19.19 & 0.762 & 0.180 & 20.45 & 0.650 & 0.288  \\ 
        Kupyn\cite{kupyn2019deblurgan} & 18.73 & 0.521 & 0.362 & 22.24 & 0.617 & 0.402 & 26.72 & 0.823 & 0.200 & \underline{17.90} & 0.625 & 0.314 & 19.06 & 0.780 & 0.233 & 20.93 & 0.673 & 0.302  \\ 
        Cho\cite{cho2021rethinking} & 17.73 & 0.449 & 0.435 & 20.96 & 0.535 & 0.465 & 25.15 & 0.785 & 0.262 & 17.30 & 0.590 & 0.344 & 18.27 & 0.728 & 0.251 & 19.88 & 0.618 & 0.351  \\ 
        MPRNet\cite{zamir2021multi} & 18.72 & 0.536 & 0.376 & 22.57 & 0.644 & 0.397 & 25.80 & 0.814 & 0.236 & 17.59 & 0.620 & 0.319 & 16.73 & 0.674 & 0.292 & 20.28 & 0.658 & 0.324  \\ 
        Restormer\cite{zamir2022restormer} & 19.07 & 0.556 & 0.361 & 22.49 & 0.626 & 0.405 & 26.36 & 0.821 & 0.223 & 17.56 & 0.612 & 0.328 & 18.25 & 0.752 & 0.249 & 20.74 & 0.673 & 0.313  \\ 
        Zhang\cite{zhang2023event} & 17.81 & 0.456 & 0.455 & 21.03 & 0.547 & 0.486 & 25.47 & 0.795 & 0.271 & 17.36 & 0.595 & 0.355 & 18.21 & 0.728 & 0.271 & 19.98 & 0.624 & 0.368  \\ 
        Fang\cite{fang2023self} & 19.28 & 0.575 & 0.360 & \underline{22.85} & 0.661 & 0.401 & 26.85 & 0.837 & 0.227 & 17.85 & \textbf{0.651} & 0.324 & 17.94 & 0.703 & 0.298 & 20.95 & 0.685 & 0.322  \\ 
        Li\cite{li2023self} & 18.83 & 0.564 & 0.259 & 22.84 & 0.669 & \underline{0.253} & 26.79 & 0.826 & \textbf{0.150} & 15.27 & 0.480 & 0.344 & 17.52 & 0.727 & 0.180 & 20.25 & 0.653 & 0.237  \\
        \hline \hline
        DIP\cite{ren2020neural}&11.95&	0.208&	0.989&	13.32&	0.309&	0.999&	12.85&	0.494&	0.845&	10.19&	0.359&	0.772&	10.88&	0.405&	0.834&	11.84&	0.355&	0.888\\ 
        VDIP-Std\cite{huo2023blind} & 10.69 & 0.126 & 0.587 & 13.51 & 0.206 & 0.547 & 14.69 & 0.368 & 0.506 & 11.80 & 0.350 & 0.476 & 13.33 & 0.496 & 0.336 & 12.80 & 0.309 & 0.491  \\ 
        VDIP-Sparse\cite{huo2023blind} & 10.23 & 0.117 & 0.609 & 13.03 & 0.165 & 0.578 & 14.31 & 0.337 & 0.535 & 11.38 & 0.337 & 0.489 & 11.53 & 0.401 & 0.411 & 12.10 & 0.271 & 0.524  \\ 
        VDIP-Extreme\cite{huo2023blind} & 10.62 & 0.137 & 0.576 & 13.12 & 0.178 & 0.557 & 14.50 & 0.359 & 0.512 & 12.25 & 0.360 & 0.465 & 13.13 & 0.483 & 0.332 & 12.72 & 0.303 & 0.488  \\ 
        \hline \hline
        DIP-GLKM & \textbf{20.16} & \textbf{0.648} & \underline{0.250} & \textbf{23.55} & \textbf{0.711} & 0.282 & \textbf{28.22} & \textbf{0.873} & 0.166 & \textbf{17.97} & \underline{0.646} & \underline{0.278} & \textbf{20.67} & \textbf{0.843} & \textbf{0.129} & \textbf{22.11} & \textbf{0.744} & \textbf{0.221}  \\ 
        VDIP-Std-GLKM & \underline{19.30} & \underline{0.600} & \textbf{0.243} & 22.72 & \underline{0.675} & \textbf{0.242} & \underline{27.33} & \underline{0.841} & \underline{0.156} & 17.10 & 0.614 & 0.292 & \underline{19.57} & \underline{0.782} & \underline{0.163} & \underline{21.21} & \underline{0.702} & \underline{0.223}  \\ 
        VDIP-Extreme-GLKM & 18.71 & 0.543 & 0.339 & 21.68 & 0.633 & 0.304 & 23.75 & 0.719 & 0.360 & 17.26 & 0.600 & 0.305 & 19.44 & 0.762 & 0.232 & 20.17 & 0.651 & 0.308  \\ 
        VDIP-Sparse-GLKM & 18.04 & 0.546 & 0.275 & 21.50 & 0.626 & 0.265 & 25.09 & 0.787 & 0.191 & 16.42 & 0.592 & 0.316 & 19.36 & 0.773 & 0.156 & 20.08 & 0.665 & 0.241  \\ 
        \Xhline{0.8pt}
    \end{tabular}
\end{table*}

%% file: fig/nonuniform_synthetic.tex
\begin{figure*}[t]
    \setlength\tabcolsep{1pt}
	\renewcommand{\arraystretch}{0.618}
    \centering
    \scriptsize
    \begin{tabular}{cccccc}
        \includegraphics[width=0.16\textwidth]{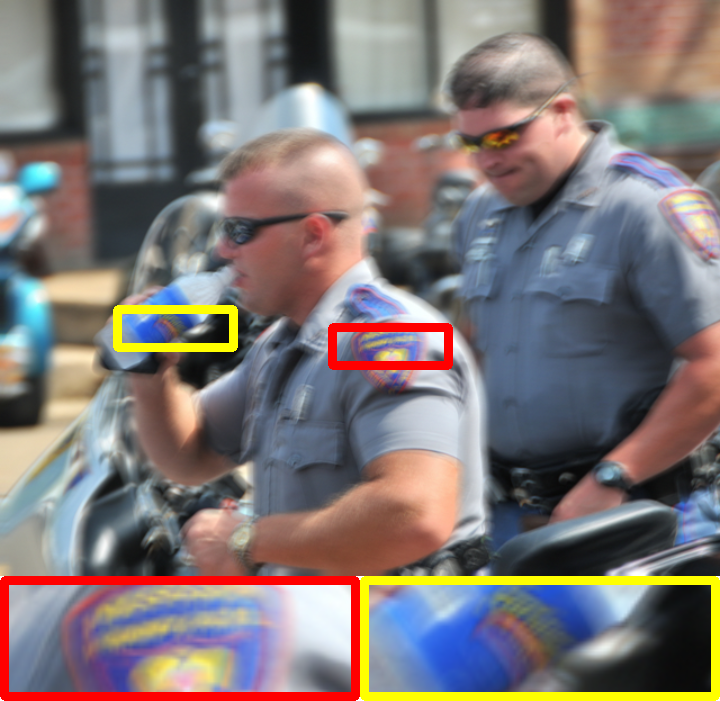}&
        \includegraphics[width=0.16\textwidth]{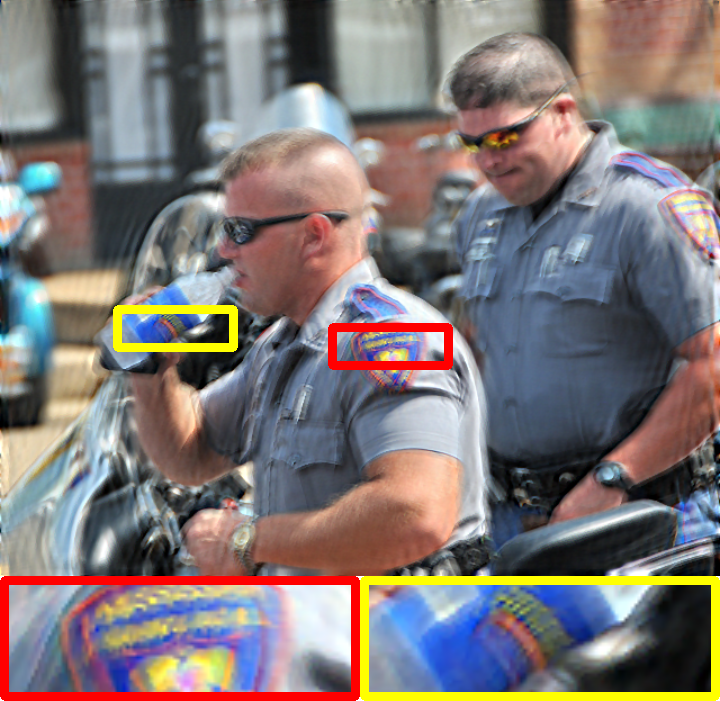}&
        \includegraphics[width=0.16\textwidth]{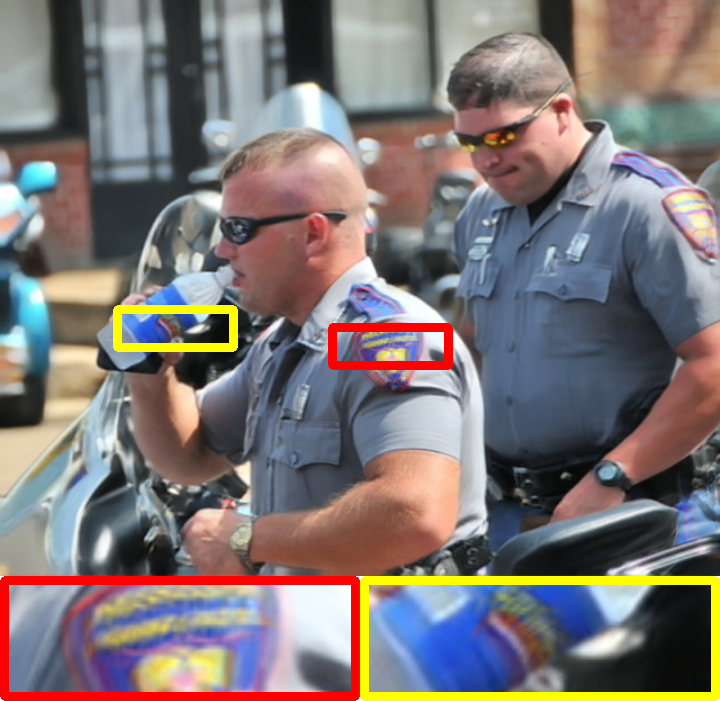}&
        \includegraphics[width=0.16\textwidth]{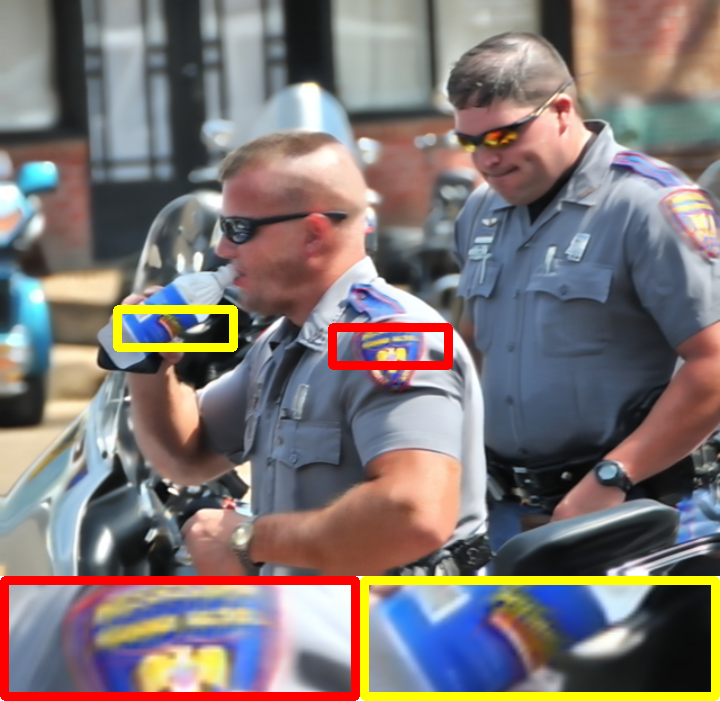}&
        \includegraphics[width=0.16\textwidth]{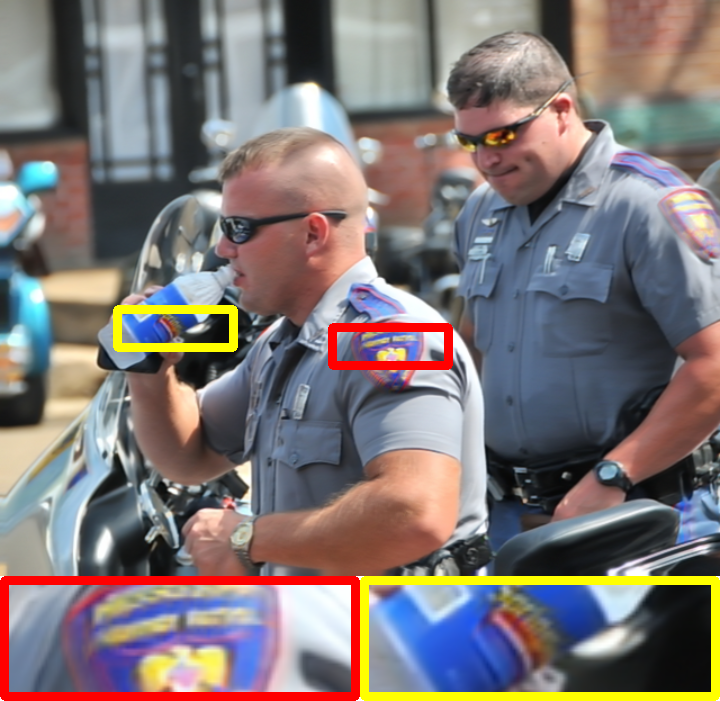}&
        \includegraphics[width=0.16\textwidth]{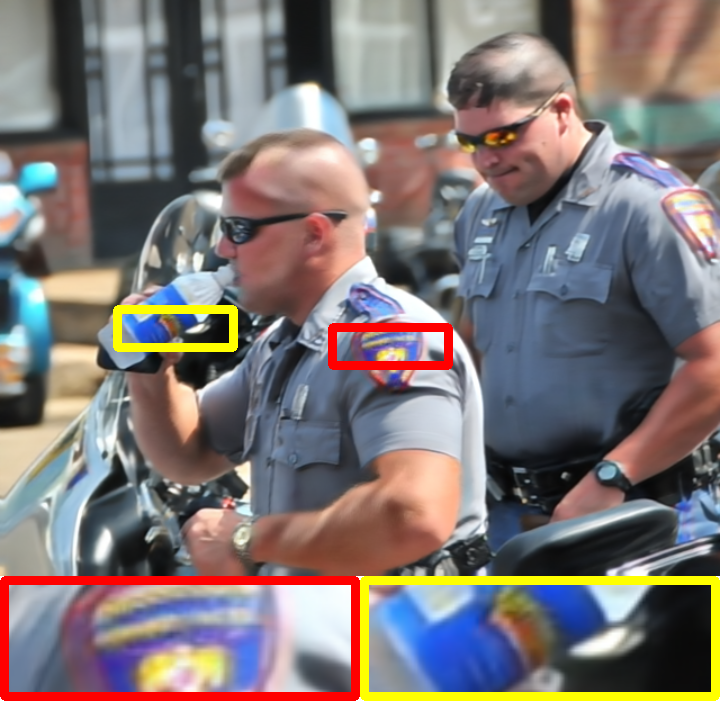}\\
        Blurred & Xu \cite{xu2013unnatural} & Kupyn \cite{kupyn2019deblurgan}  & MPRNet \cite{zamir2021multi} & Restormer \cite{zamir2022restormer} & Zhang\cite{zhang2023event}\\
        \includegraphics[width=0.16\textwidth]{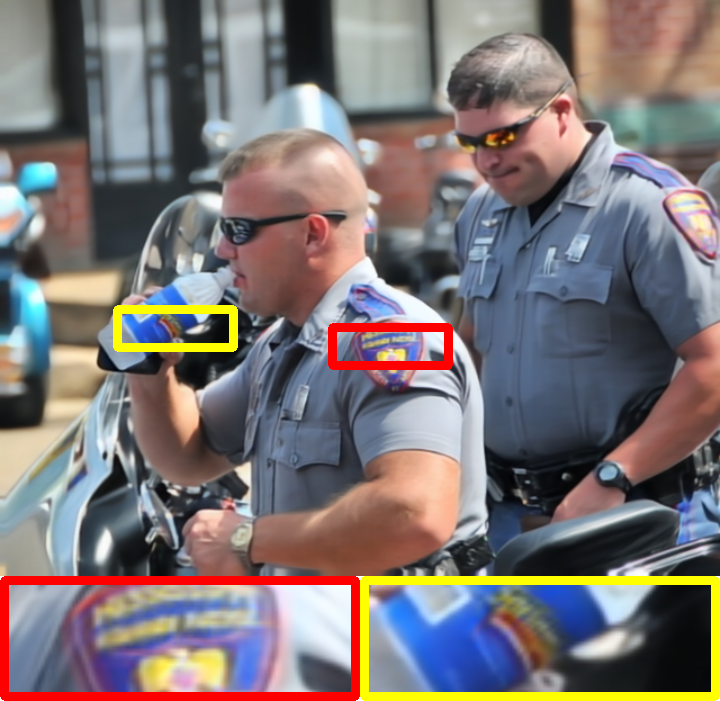}&
        \includegraphics[width=0.16\textwidth]{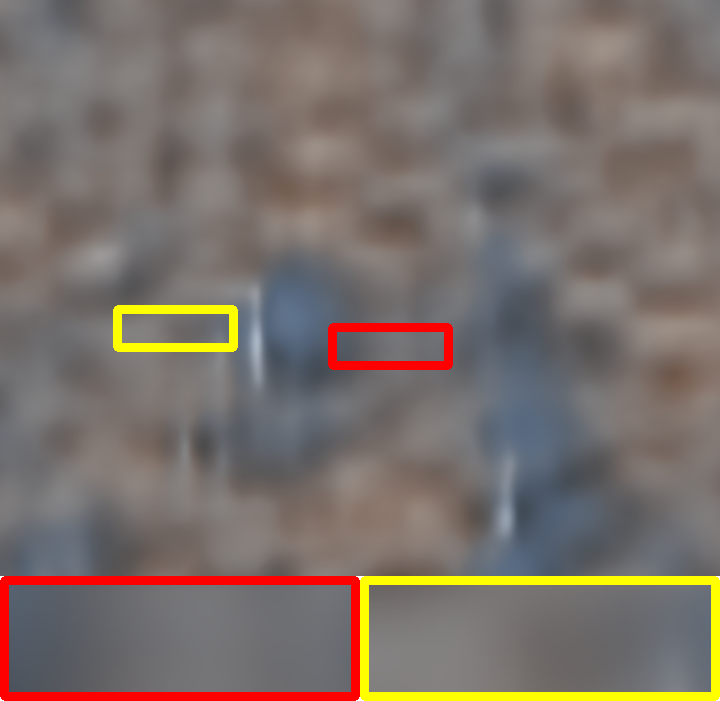}&
        \includegraphics[width=0.16\textwidth]{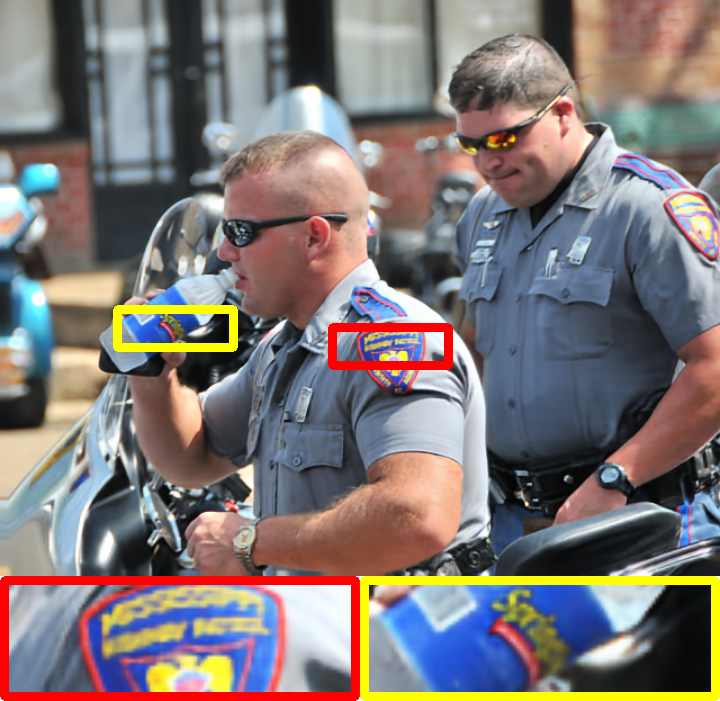}&
        \includegraphics[width=0.16\textwidth]{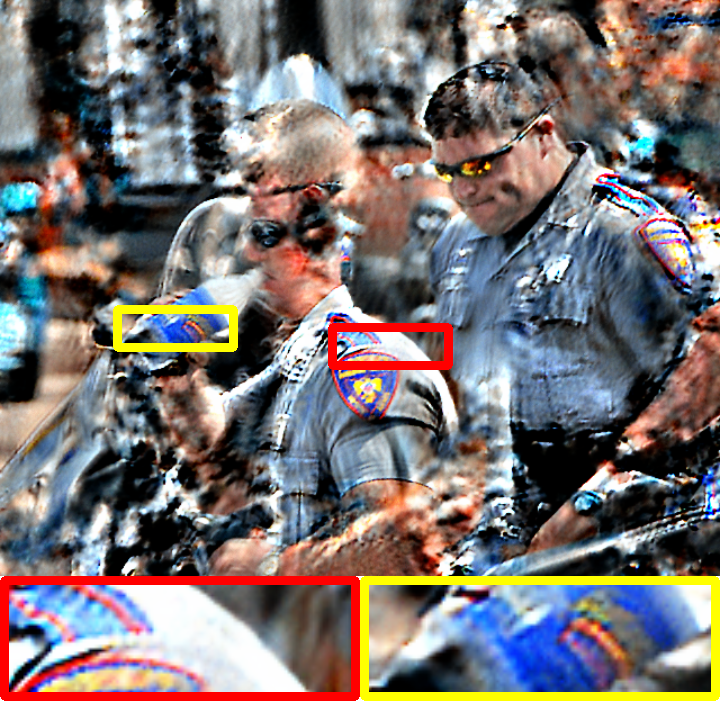}&
        \includegraphics[width=0.16\textwidth]{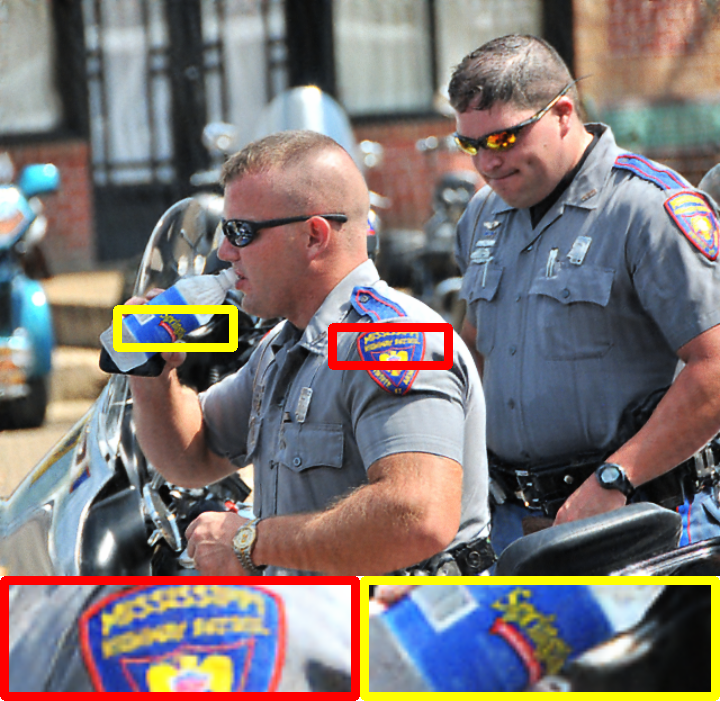}&
        \includegraphics[width=0.16\textwidth]{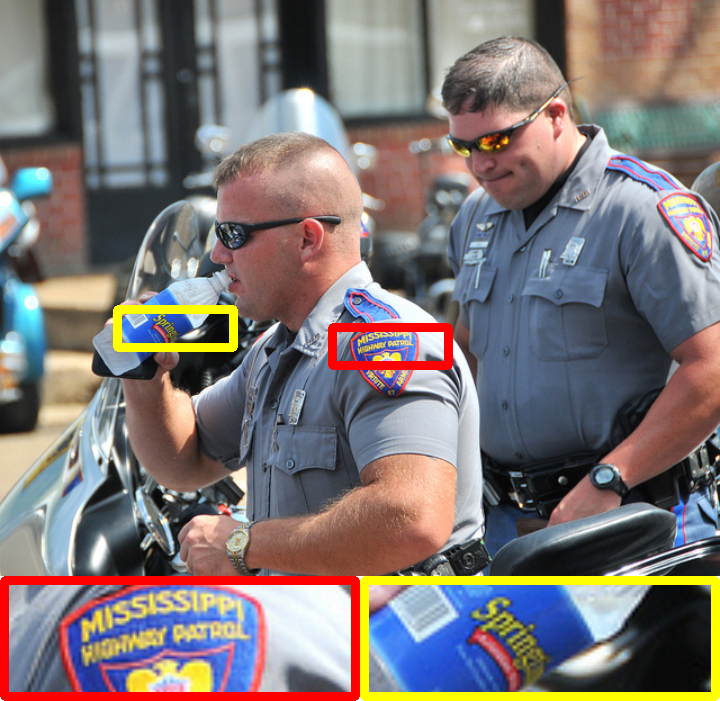}\\
        Fang\cite{fang2023self} & DIP\cite{ren2020neural} & DIP-GLKM & VDIP-Std\cite{huo2023blind} & VDIP-Std-GLKM & Ground truth
    \end{tabular}
    \caption{Visual results on our non-uniform synthetic dataset. More comparisons can be found in the supplementary material.}
    \label{fig:nonuniform_synthetic}
\end{figure*}

%% file: fig/nonuniform_lai.tex
\begin{figure*}[!t]
    \setlength\tabcolsep{1pt}
	\renewcommand{\arraystretch}{0.618}
    \centering
    \scriptsize
    \begin{tabular}{cccccc}
        \includegraphics[width=0.16\textwidth]{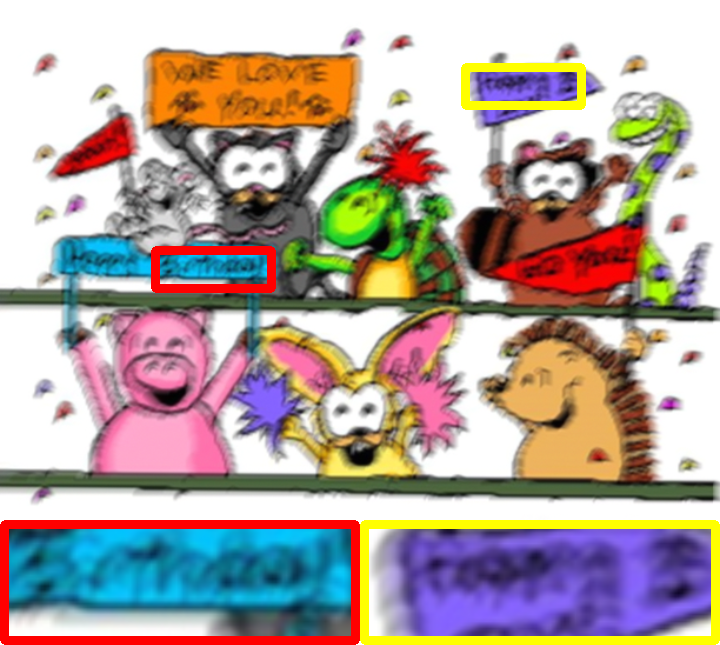}&
        \includegraphics[width=0.16\textwidth]{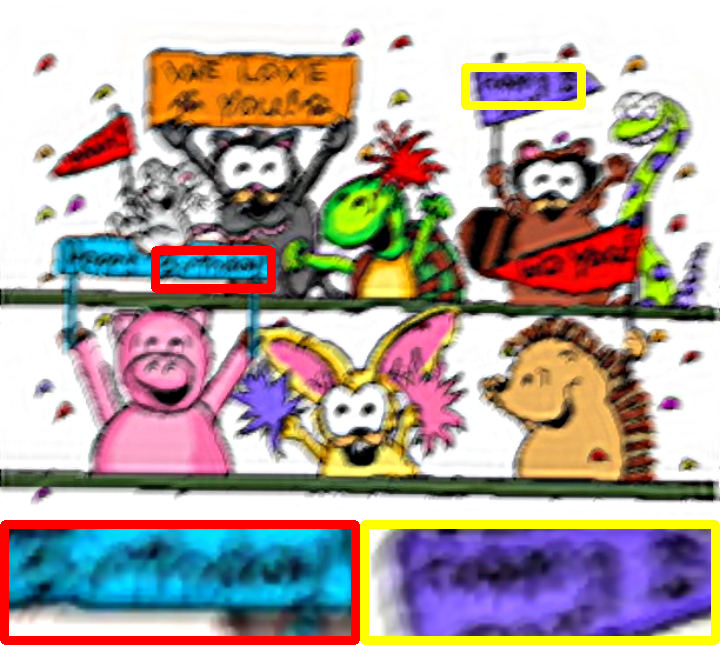}&
        \includegraphics[width=0.16\textwidth]{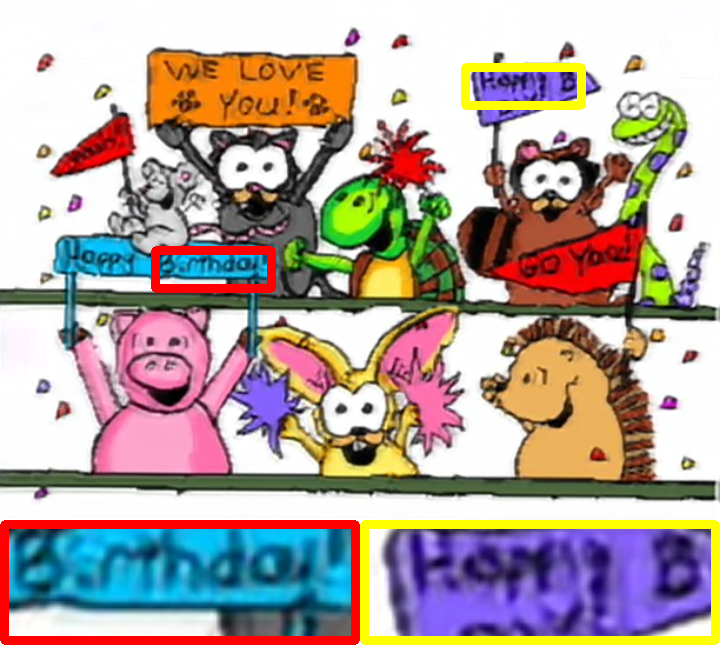}&
        \includegraphics[width=0.16\textwidth]{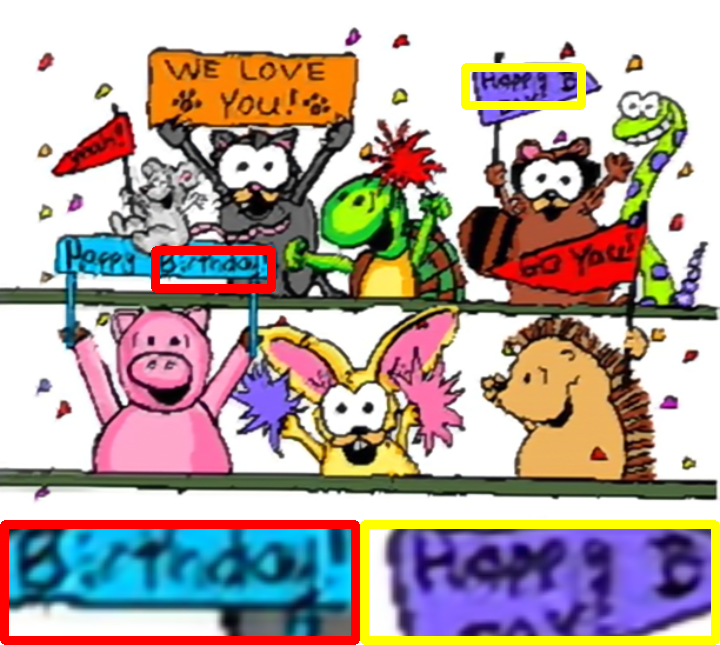}&
        \includegraphics[width=0.16\textwidth]{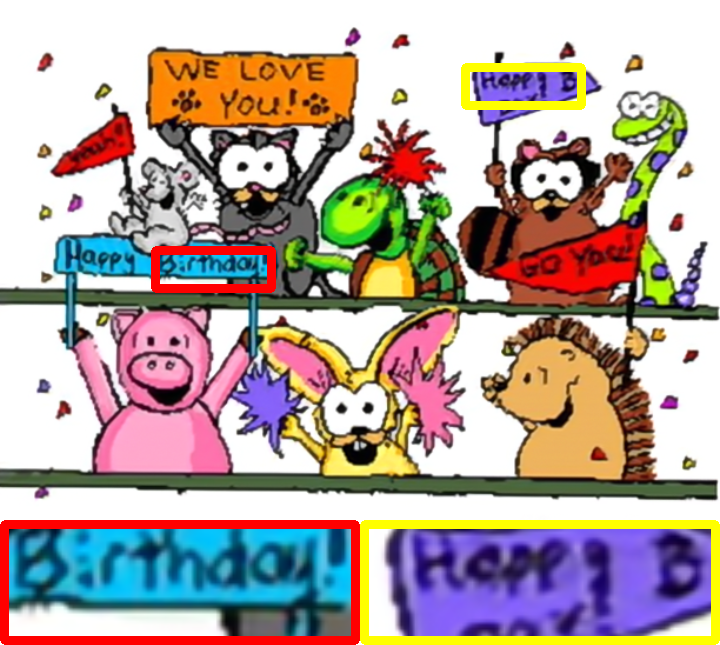}&
        \includegraphics[width=0.16\textwidth]{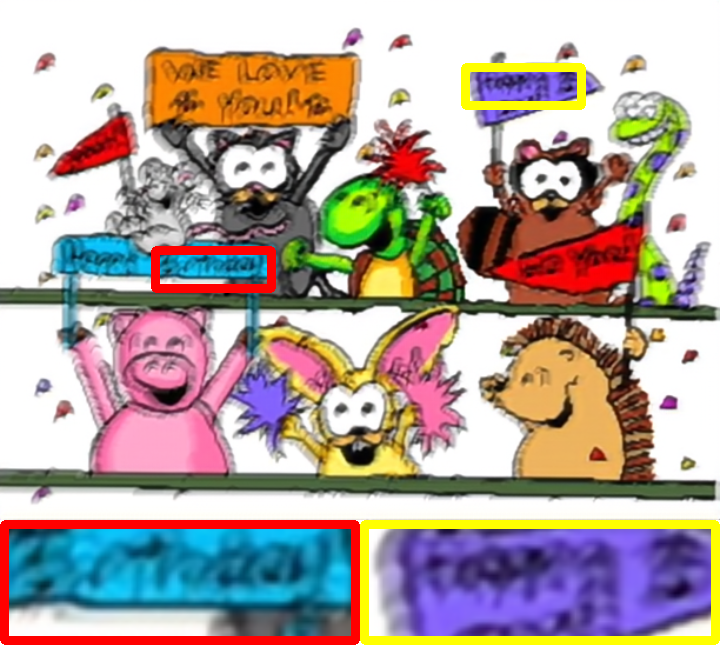}\\
        Blurred & Xu \cite{xu2013unnatural} & Kupyn \cite{kupyn2019deblurgan}  & MPRNet \cite{zamir2021multi} & Restormer \cite{zamir2022restormer} & Zhang\cite{zhang2023event}\\
        \includegraphics[width=0.16\textwidth]{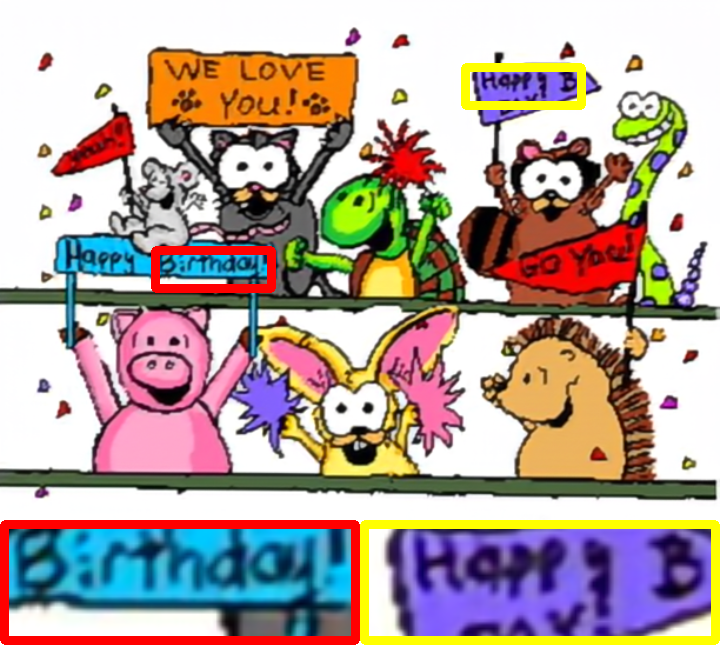}&
        \includegraphics[width=0.16\textwidth]{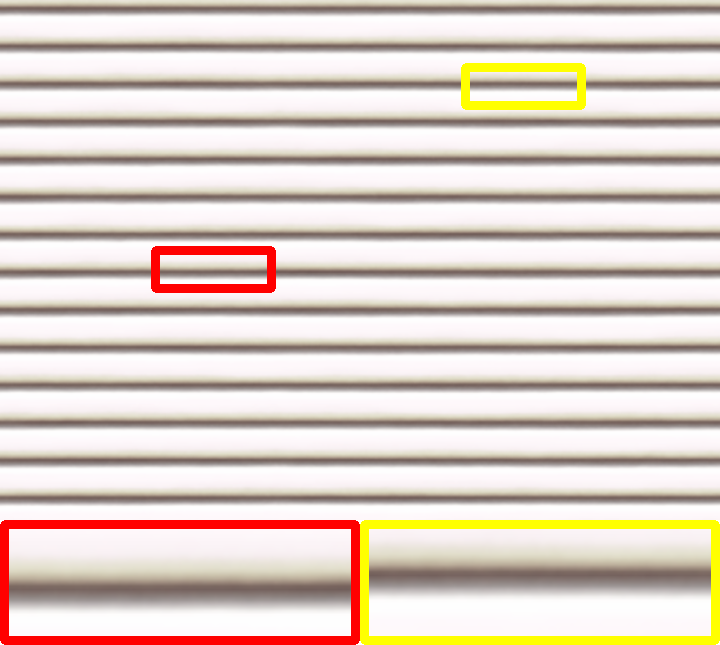}&
        \includegraphics[width=0.16\textwidth]{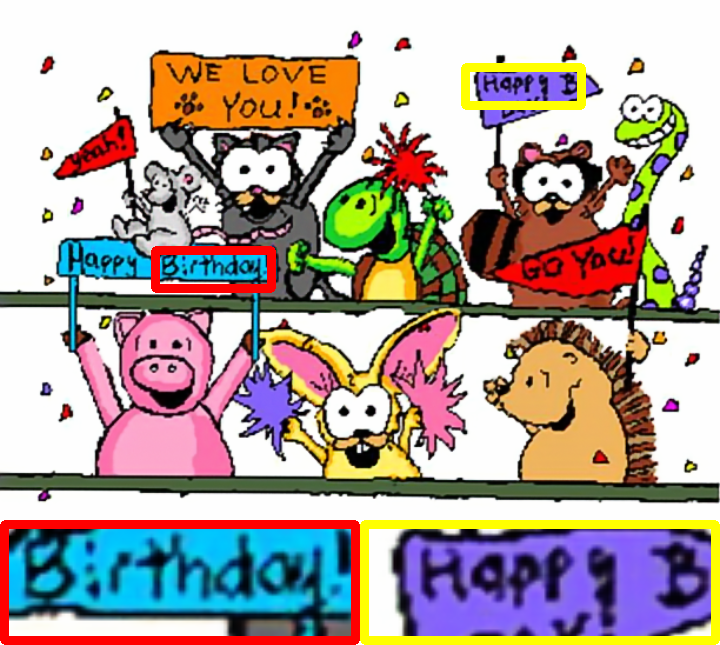}&
        \includegraphics[width=0.16\textwidth]{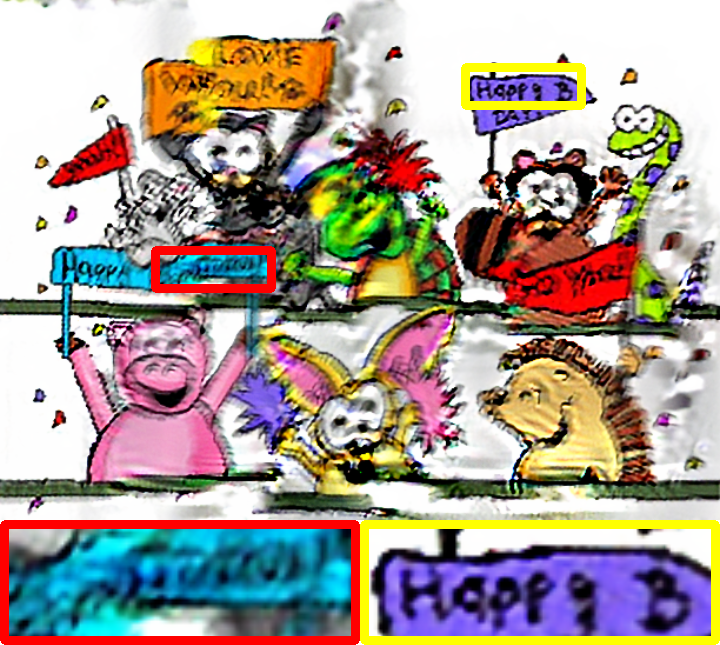}&
        \includegraphics[width=0.16\textwidth]{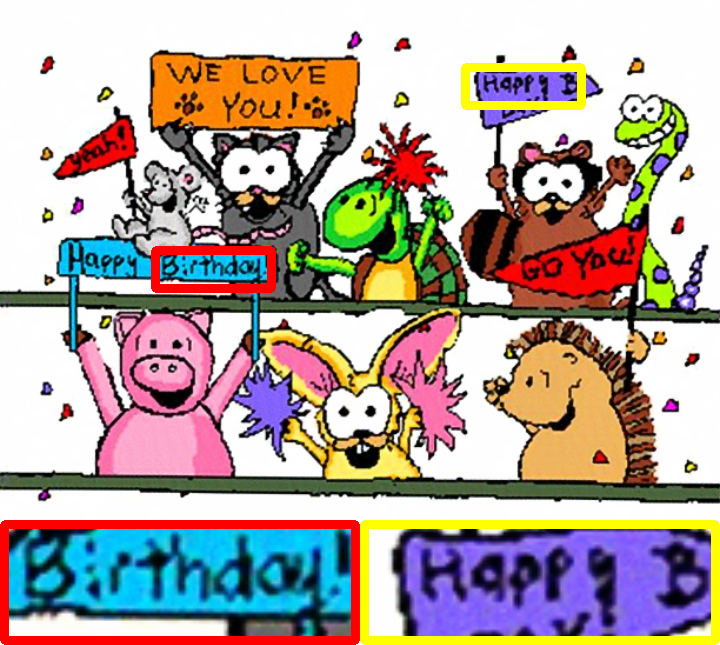}&
        \includegraphics[width=0.16\textwidth]{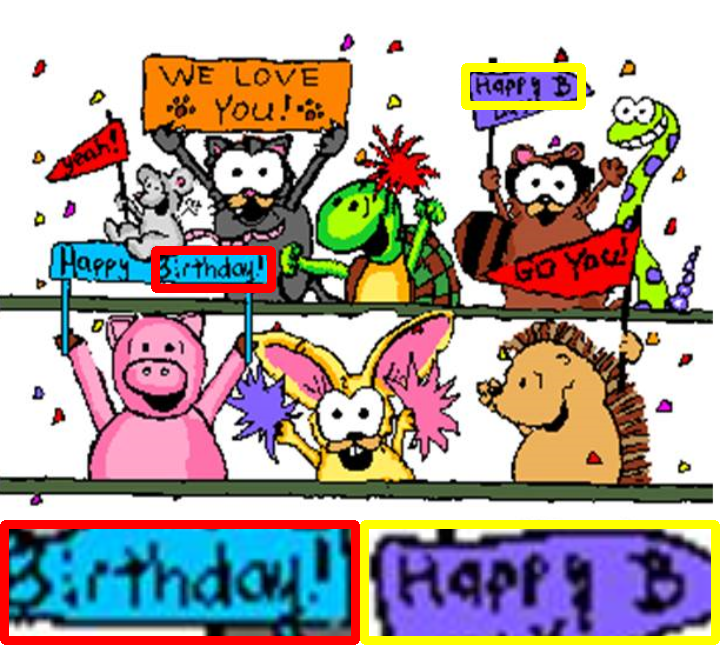}\\
        Fang\cite{fang2023self} & DIP\cite{ren2020neural} & DIP-GLKM & VDIP-Std\cite{huo2023blind} & VDIP-Std-GLKM & Ground truth
    \end{tabular}
    \caption{Visual results on Lai \MakeLowercase{\textit{et al.}}'s non-uniform blurred dataset. More comparisons can be found in the supplementary material.}
    \label{fig:nonuniform_lai}
\end{figure*}

%% file: GLKM.bbl
\begin{thebibliography}{100}
\providecommand{\url}[1]{#1}
\csname url@samestyle\endcsname
\providecommand{\newblock}{\relax}
\providecommand{\bibinfo}[2]{#2}
\providecommand{\BIBentrySTDinterwordspacing}{\spaceskip=0pt\relax}
\providecommand{\BIBentryALTinterwordstretchfactor}{4}
\providecommand{\BIBentryALTinterwordspacing}{\spaceskip=\fontdimen2\font plus
\BIBentryALTinterwordstretchfactor\fontdimen3\font minus \fontdimen4\font\relax}
\providecommand{\BIBforeignlanguage}[2]{{%
\expandafter\ifx\csname l@#1\endcsname\relax
\typeout{** WARNING: IEEEtran.bst: No hyphenation pattern has been}%
\typeout{** loaded for the language `#1'. Using the pattern for}%
\typeout{** the default language instead.}%
\else
\language=\csname l@#1\endcsname
\fi
#2}}
\providecommand{\BIBdecl}{\relax}
\BIBdecl

\bibitem{krishnan2009fast}
D.~Krishnan and R.~Fergus, ``Fast image deconvolution using hyper-laplacian priors,'' in \emph{Advances in Neural Information Processing Systems}, Y.~Bengio, D.~Schuurmans, J.~Lafferty, C.~Williams, and A.~Culotta, Eds., vol.~22.\hskip 1em plus 0.5em minus 0.4em\relax Curran Associates, Inc., 2009.

\bibitem{krishnan2011blind}
D.~Krishnan, T.~Tay, and R.~Fergus, ``Blind deconvolution using a normalized sparsity measure,'' in \emph{Proceedings of the IEEE/CVF Conference on Computer Vision and Pattern Recognition (CVPR)}.\hskip 1em plus 0.5em minus 0.4em\relax IEEE, 2011, pp. 233--240.

\bibitem{tog2008_shan_high_quality}
Q.~Shan, J.~Jia, and A.~Agarwala, ``High-quality motion deblurring from a single image,'' \emph{ACM Transactions on Graphics (TOG)}, vol.~27, no.~3, p. 1–10, 2008.

\bibitem{xu2013unnatural}
L.~Xu, S.~Zheng, and J.~Jia, ``Unnatural l0 sparse representation for natural image deblurring,'' in \emph{Proceedings of the IEEE/CVF Conference on Computer Vision and Pattern Recognition (CVPR)}, 2013, pp. 1107--1114.

\bibitem{pan2014deblurring}
J.~Pan, Z.~Hu, Z.~Su, and M.-H. Yang, ``Deblurring text images via $l_0$-regularized intensity and gradient prior,'' in \emph{Proceedings of the IEEE Conference on Computer Vision and Pattern Recognition (CVPR)}, 2014, pp. 2901--2908.

\bibitem{pan2016l_0}
------, ``$ l_0 $-regularized intensity and gradient prior for deblurring text images and beyond,'' \emph{IEEE Transactions on Pattern Analysis and Machine Intelligence}, vol.~39, no.~2, pp. 342--355, 2016.

\bibitem{michaeli2014blind}
T.~Michaeli and M.~Irani, ``Blind deblurring using internal patch recurrence,'' in \emph{Proceedings of the European Conference on Computer Vision (ECCV)}.\hskip 1em plus 0.5em minus 0.4em\relax Springer, 2014, pp. 783--798.

\bibitem{pan2017deblurring}
J.~Pan, D.~Sun, H.~Pfister, and M.-H. Yang, ``Deblurring images via dark channel prior,'' \emph{IEEE Transactions on Pattern Analysis and Machine Intelligence}, vol.~40, no.~10, pp. 2315--2328, 2017.

\bibitem{fergus2006removing}
R.~Fergus, B.~Singh, A.~Hertzmann, S.~T. Roweis, and W.~T. Freeman, ``Removing camera shake from a single photograph,'' in \emph{Proceedings of ACM SIGGRAPH Conference}, 2006, pp. 787--794.

\bibitem{tip2015_zhou_variational_dirichlet_deblur}
X.~Zhou, J.~Mateos, F.~Zhou, R.~Molina, and A.~K. Katsaggelos, ``Variational dirichlet blur kernel estimation,'' \emph{IEEE Transactions on Image Processing (TIP)}, vol.~24, no.~12, pp. 5127--5139, 2015.

\bibitem{book2017_bishop_review}
T.~E. Bishop, S.~D. Babacan, B.~Amizic, A.~K. Katsaggelos, T.~Chan, and R.~Molina, ``Blind image deconvolution: problem formulation and existing approaches,'' in \emph{Blind Image Deconvolution}.\hskip 1em plus 0.5em minus 0.4em\relax CRC press, 2017, pp. 21--62.

\bibitem{yue2024deep}
Z.~Yue, H.~Yong, Q.~Zhao, L.~Zhang, D.~Meng, and K.-Y.~K. Wong, ``Deep variational network toward blind image restoration,'' \emph{IEEE Transactions on Pattern Analysis and Machine Intelligence}, vol.~46, no.~11, pp. 7011--7026, 2024.

\bibitem{cho2009fast}
S.~Cho and S.~Lee, ``Fast motion deblurring,'' in \emph{ACM SIGGRAPH Asia 2009 papers}, 2009, pp. 1--8.

\bibitem{chan1998total}
T.~F. Chan and C.-K. Wong, ``Total variation blind deconvolution,'' \emph{IEEE Transactions on Image Processing (TIP)}, vol.~7, no.~3, pp. 370--375, 1998.

\bibitem{yan2017image}
Y.~Yan, W.~Ren, Y.~Guo, R.~Wang, and X.~Cao, ``Image deblurring via extreme channels prior,'' in \emph{Proceedings of the IEEE/CVF Conference on Computer Vision and Pattern Recognition (CVPR)}, 2017, pp. 4003--4011.

\bibitem{icassp1997_molina_dirichlet_deblur}
R.~Molina, A.~K. Katsaggelos, J.~Abad, and J.~Mateos, ``A bayesian approach to blind deconvolution based on dirichlet distributions,'' in \emph{IEEE International Conference on Acoustics, Speech, and Signal Processing (ICASSP)}, vol.~4.\hskip 1em plus 0.5em minus 0.4em\relax IEEE, 1997, pp. 2809--2812.

\bibitem{levin2009understanding}
A.~Levin, Y.~Weiss, F.~Durand, and W.~T. Freeman, ``Understanding and evaluating blind deconvolution algorithms,'' in \emph{Proceedings of the IEEE/CVF Conference on Computer Vision and Pattern Recognition (CVPR)}.\hskip 1em plus 0.5em minus 0.4em\relax IEEE, 2009, pp. 1964--1971.

\bibitem{nah2017deep}
S.~Nah, T.~Hyun~Kim, and K.~Mu~Lee, ``Deep multi-scale convolutional neural network for dynamic scene deblurring,'' in \emph{Proceedings of the IEEE/CVF Conference on Computer Vision and Pattern Recognition (CVPR)}, 2017, pp. 3883--3891.

\bibitem{tao2018scale}
X.~Tao, H.~Gao, X.~Shen, J.~Wang, and J.~Jia, ``Scale-recurrent network for deep image deblurring,'' in \emph{Proceedings of the IEEE/CVF Conference on Computer Vision and Pattern Recognition (CVPR)}, 2018, pp. 8174--8182.

\bibitem{kupyn2018deblurgan}
O.~Kupyn, V.~Budzan, M.~Mykhailych, D.~Mishkin, and J.~Matas, ``Deblurgan: Blind motion deblurring using conditional adversarial networks,'' in \emph{Proceedings of the IEEE/CVF Conference on Computer Vision and Pattern Recognition (CVPR)}, 2018, pp. 8183--8192.

\bibitem{kupyn2019deblurgan}
O.~Kupyn, T.~Martyniuk, J.~Wu, and Z.~Wang, ``Deblurgan-v2: Deblurring (orders-of-magnitude) faster and better,'' in \emph{Proceedings of the IEEE/CVF International Conference on Computer Vision (ICCV)}, 2019, pp. 8878--8887.

\bibitem{zamir2021multi}
S.~W. Zamir, A.~Arora, S.~Khan, M.~Hayat, F.~S. Khan, M.-H. Yang, and L.~Shao, ``Multi-stage progressive image restoration,'' in \emph{Proceedings of the IEEE/CVF Conference on Computer Vision and Pattern Recognition (CVPR)}, 2021, pp. 14\,821--14\,831.

\bibitem{zamir2022restormer}
S.~W. Zamir, A.~Arora, S.~Khan, M.~Hayat, F.~S. Khan, and M.-H. Yang, ``Restormer: Efficient transformer for high-resolution image restoration,'' in \emph{Proceedings of the IEEE/CVF Conference on Computer Vision and Pattern Recognition (CVPR)}, 2022, pp. 5728--5739.

\bibitem{ulyanov2018deep}
D.~Ulyanov, A.~Vedaldi, and V.~Lempitsky, ``Deep image prior,'' in \emph{Proceedings of the IEEE/CVF Conference on Computer Vision and Pattern Recognition (CVPR)}, 2018, pp. 9446--9454.

\bibitem{ren2020neural}
D.~Ren, K.~Zhang, Q.~Wang, Q.~Hu, and W.~Zuo, ``Neural blind deconvolution using deep priors,'' in \emph{Proceedings of the IEEE/CVF Conference on Computer Vision and Pattern Recognition (CVPR)}, 2020, pp. 3341--3350.

\bibitem{huo2023blind}
D.~Huo, A.~Masoumzadeh, R.~Kushol, and Y.-H. Yang, ``Blind image deconvolution using variational deep image prior,'' \emph{IEEE Transactions on Pattern Analysis and Machine Intelligence}, vol.~45, no.~10, pp. 11\,472--11\,483, 2023.

\bibitem{li2023self}
J.~Li, W.~Wang, Y.~Nan, and H.~Ji, ``Self-supervised blind motion deblurring with deep expectation maximization,'' in \emph{Proceedings of the IEEE/CVF Conference on Computer Vision and Pattern Recognition (CVPR)}, 2023, pp. 13\,986--13\,996.

\bibitem{ho2020denoising}
J.~Ho, A.~Jain, and P.~Abbeel, ``Denoising diffusion probabilistic models,'' \emph{Advances in Neural Information Processing Systems}, vol.~33, pp. 6840--6851, 2020.

\bibitem{song2021scorebased}
\BIBentryALTinterwordspacing
Y.~Song, J.~Sohl-Dickstein, D.~P. Kingma, A.~Kumar, S.~Ermon, and B.~Poole, ``Score-based generative modeling through stochastic differential equations,'' in \emph{International Conference on Learning Representations}, 2021. [Online]. Available: \url{https://openreview.net/forum?id=PxTIG12RRHS}
\BIBentrySTDinterwordspacing

\bibitem{chung2023parallel}
H.~Chung, J.~Kim, S.~Kim, and J.~C. Ye, ``Parallel diffusion models of operator and image for blind inverse problems,'' in \emph{Proceedings of the IEEE/CVF Conference on Computer Vision and Pattern Recognition (CVPR)}, 2023, pp. 6059--6069.

\bibitem{chihaoui2024blind}
H.~Chihaoui, A.~Lemkhenter, and P.~Favaro, ``Blind image restoration via fast diffusion inversion,'' in \emph{Advances in Neural Information Processing Systems}, vol.~37.\hskip 1em plus 0.5em minus 0.4em\relax Curran Associates, Inc., 2024, pp. 34\,513--34\,532.

\bibitem{pan2021exploiting}
X.~Pan, X.~Zhan, B.~Dai, D.~Lin, C.~C. Loy, and P.~Luo, ``Exploiting deep generative prior for versatile image restoration and manipulation,'' \emph{IEEE Transactions on Pattern Analysis and Machine Intelligence}, vol.~44, no.~11, pp. 7474--7489, 2021.

\bibitem{chan2021glean}
K.~C. Chan, X.~Wang, X.~Xu, J.~Gu, and C.~C. Loy, ``Glean: Generative latent bank for large-factor image super-resolution,'' in \emph{Proceedings of the IEEE/CVF Conference on Computer Vision and Pattern Recognition (CVPR)}, 2021, pp. 14\,245--14\,254.

\bibitem{yang2021gan}
T.~Yang, P.~Ren, X.~Xie, and L.~Zhang, ``Gan prior embedded network for blind face restoration in the wild,'' in \emph{Proceedings of the IEEE/CVF Conference on Computer Vision and Pattern Recognition (CVPR)}, 2021, pp. 672--681.

\bibitem{wang2021towards}
X.~Wang, Y.~Li, H.~Zhang, and Y.~Shan, ``Towards real-world blind face restoration with generative facial prior,'' in \emph{Proceedings of the IEEE/CVF Conference on Computer Vision and Pattern Recognition (CVPR)}, 2021, pp. 9168--9178.

\bibitem{goodfellow2014generative}
I.~J. Goodfellow, J.~Pouget-Abadie, M.~Mirza, B.~Xu, D.~Warde-Farley, S.~Ozair, A.~Courville, and Y.~Bengio, ``Generative adversarial nets,'' \emph{Advances in Neural Information Processing Systems}, vol.~27, 2014.

\bibitem{zhang2024blind}
J.~Zhang, Z.~Yue, H.~Wang, Q.~Zhao, and D.~Meng, ``Blind image deconvolution by generative-based kernel prior and initializer via latent encoding,'' in \emph{Proceedings of the European Conference on Computer Vision (ECCV)}.\hskip 1em plus 0.5em minus 0.4em\relax Springer, 2024, pp. 73--92.

\bibitem{lai2016comparative}
W.-S. Lai, J.-B. Huang, Z.~Hu, N.~Ahuja, and M.-H. Yang, ``A comparative study for single image blind deblurring,'' in \emph{Proceedings of the IEEE/CVF Conference on Computer Vision and Pattern Recognition (CVPR)}, 2016, pp. 1701--1709.

\bibitem{cai2009blind}
J.-F. Cai, H.~Ji, C.~Liu, and Z.~Shen, ``Blind motion deblurring from a single image using sparse approximation,'' in \emph{Proceedings of the IEEE/CVF Conference on Computer Vision and Pattern Recognition (CVPR)}.\hskip 1em plus 0.5em minus 0.4em\relax IEEE, 2009, pp. 104--111.

\bibitem{sun2013edge}
L.~Sun, S.~Cho, J.~Wang, and J.~Hays, ``Edge-based blur kernel estimation using patch priors,'' in \emph{IEEE international conference on computational photography (ICCP)}.\hskip 1em plus 0.5em minus 0.4em\relax IEEE, 2013, pp. 1--8.

\bibitem{ren2016image}
W.~Ren, X.~Cao, J.~Pan, X.~Guo, W.~Zuo, and M.-H. Yang, ``Image deblurring via enhanced low-rank prior,'' \emph{IEEE Transactions on Image Processing (TIP)}, vol.~25, no.~7, pp. 3426--3437, 2016.

\bibitem{liu2014blind}
G.~Liu, S.~Chang, and Y.~Ma, ``Blind image deblurring using spectral properties of convolution operators,'' \emph{IEEE Transactions on Image Processing (TIP)}, vol.~23, no.~12, pp. 5047--5056, 2014.

\bibitem{perrone2014total}
D.~Perrone and P.~Favaro, ``Total variation blind deconvolution: The devil is in the details,'' in \emph{Proceedings of the IEEE/CVF Conference on Computer Vision and Pattern Recognition (CVPR)}, 2014, pp. 2909--2916.

\bibitem{zuo2016learning}
W.~Zuo, D.~Ren, D.~Zhang, S.~Gu, and L.~Zhang, ``Learning iteration-wise generalized shrinkage--thresholding operators for blind deconvolution,'' \emph{IEEE Transactions on Image Processing (TIP)}, vol.~25, no.~4, pp. 1751--1764, 2016.

\bibitem{sun2015learning}
J.~Sun, W.~Cao, Z.~Xu, and J.~Ponce, ``Learning a convolutional neural network for non-uniform motion blur removal,'' in \emph{Proceedings of the IEEE/CVF Conference on Computer Vision and Pattern Recognition (CVPR)}, 2015, pp. 769--777.

\bibitem{chakrabarti2016neural}
A.~Chakrabarti, ``A neural approach to blind motion deblurring,'' in \emph{Proceedings of the European Conference on Computer Vision (ECCV)}.\hskip 1em plus 0.5em minus 0.4em\relax Springer, 2016, pp. 221--235.

\bibitem{gong2017motion}
D.~Gong, J.~Yang, L.~Liu, Y.~Zhang, I.~Reid, C.~Shen, A.~Van Den~Hengel, and Q.~Shi, ``From motion blur to motion flow: A deep learning solution for removing heterogeneous motion blur,'' in \emph{Proceedings of the IEEE/CVF Conference on Computer Vision and Pattern Recognition (CVPR)}, 2017, pp. 2319--2328.

\bibitem{pan2020physics}
J.~Pan, J.~Dong, Y.~Liu, J.~Zhang, J.~Ren, J.~Tang, Y.-W. Tai, and M.-H. Yang, ``Physics-based generative adversarial models for image restoration and beyond,'' \emph{IEEE Transactions on Pattern Analysis and Machine Intelligence}, vol.~43, no.~7, pp. 2449--2462, 2020.

\bibitem{cai2020dark}
J.~Cai, W.~Zuo, and L.~Zhang, ``Dark and bright channel prior embedded network for dynamic scene deblurring,'' \emph{IEEE Transactions on Image Processing (TIP)}, vol.~29, pp. 6885--6897, 2020.

\bibitem{cho2021rethinking}
S.-J. Cho, S.-W. Ji, J.-P. Hong, S.-W. Jung, and S.-J. Ko, ``Rethinking coarse-to-fine approach in single image deblurring,'' in \emph{Proceedings of the IEEE/CVF International Conference on Computer Vision (ICCV)}, 2021, pp. 4641--4650.

\bibitem{chen2022simple}
L.~Chen, X.~Chu, X.~Zhang, and J.~Sun, ``Simple baselines for image restoration,'' in \emph{Proceedings of the European Conference on Computer Vision (ECCV)}.\hskip 1em plus 0.5em minus 0.4em\relax Springer, 2022, pp. 17--33.

\bibitem{kong2023efficient}
L.~Kong, J.~Dong, J.~Ge, M.~Li, and J.~Pan, ``Efficient frequency domain-based transformers for high-quality image deblurring,'' in \emph{Proceedings of the IEEE/CVF Conference on Computer Vision and Pattern Recognition (CVPR)}, 2023, pp. 5886--5895.

\bibitem{pan2023cascaded}
J.~Pan, B.~Xu, H.~Bai, J.~Tang, and M.-H. Yang, ``Cascaded deep video deblurring using temporal sharpness prior and non-local spatial-temporal similarity,'' \emph{IEEE Transactions on Pattern Analysis and Machine Intelligence}, vol.~45, no.~8, pp. 9411--9425, 2023.

\bibitem{tang2024residual}
X.~Tang, X.~Hu, X.~Gu, and J.~Sun, ``Residual-conditioned optimal transport: towards structure-preserving unpaired and paired image restoration,'' in \emph{International Conference on Machine Learning}.\hskip 1em plus 0.5em minus 0.4em\relax PMLR, 2024, pp. 47\,757--47\,777.

\bibitem{pan2025learning}
J.~Pan, L.~Sun, B.~Xu, J.~Dong, and J.~Tang, ``Learning efficient deep discriminative spatial and temporal networks for video deblurring,'' \emph{IEEE Transactions on Pattern Analysis and Machine Intelligence}, 2025.

\bibitem{tang2025degradation}
X.~Tang, X.~Gu, X.~He, X.~Hu, and J.~Sun, ``Degradation-aware residual-conditioned optimal transport for unified image restoration,'' \emph{IEEE Transactions on Pattern Analysis and Machine Intelligence}, 2025.

\bibitem{rim2020real}
J.~Rim, H.~Lee, J.~Won, and S.~Cho, ``Real-world blur dataset for learning and benchmarking deblurring algorithms,'' in \emph{Proceedings of the European Conference on Computer Vision (ECCV)}.\hskip 1em plus 0.5em minus 0.4em\relax Springer, 2020, pp. 184--201.

\bibitem{tang2023uncertainty}
X.~Tang, X.~Zhao, J.~Liu, J.~Wang, Y.~Miao, and T.~Zeng, ``Uncertainty-aware unsupervised image deblurring with deep residual prior,'' in \emph{Proceedings of the IEEE/CVF Conference on Computer Vision and Pattern Recognition (CVPR)}, 2023, pp. 9883--9892.

\bibitem{asim2020blind}
M.~Asim, F.~Shamshad, and A.~Ahmed, ``Blind image deconvolution using deep generative priors,'' \emph{IEEE Transactions on Computational Imaging}, vol.~6, pp. 1493--1506, 2020.

\bibitem{liang2021flow}
J.~Liang, K.~Zhang, S.~Gu, L.~Van~Gool, and R.~Timofte, ``Flow-based kernel prior with application to blind super-resolution,'' in \emph{Proceedings of the IEEE/CVF Conference on Computer Vision and Pattern Recognition (CVPR)}, 2021, pp. 10\,601--10\,610.

\bibitem{yue2022blind}
Z.~Yue, Q.~Zhao, J.~Xie, L.~Zhang, D.~Meng, and K.-Y.~K. Wong, ``Blind image super-resolution with elaborate degradation modeling on noise and kernel,'' in \emph{Proceedings of the IEEE/CVF Conference on Computer Vision and Pattern Recognition (CVPR)}, 2022, pp. 2128--2138.

\bibitem{gandelsman2019double}
Y.~Gandelsman, A.~Shocher, and M.~Irani, ``" double-dip": unsupervised image decomposition via coupled deep-image-priors,'' in \emph{Proceedings of the IEEE/CVF Conference on Computer Vision and Pattern Recognition (CVPR)}, 2019, pp. 11\,026--11\,035.

\bibitem{zhao2021retinexdip}
Z.~Zhao, B.~Xiong, L.~Wang, Q.~Ou, L.~Yu, and F.~Kuang, ``Retinexdip: A unified deep framework for low-light image enhancement,'' \emph{IEEE Transactions on Circuits and Systems for Video Technology}, vol.~32, no.~3, pp. 1076--1088, 2021.

\bibitem{gong2018pet}
K.~Gong, C.~Catana, J.~Qi, and Q.~Li, ``Pet image reconstruction using deep image prior,'' \emph{IEEE Transactions on Medical Imaging}, vol.~38, no.~7, pp. 1655--1665, 2018.

\bibitem{luo2021hyperspectral}
Y.-S. Luo, X.-L. Zhao, T.-X. Jiang, Y.-B. Zheng, and Y.~Chang, ``Hyperspectral mixed noise removal via spatial-spectral constrained unsupervised deep image prior,'' \emph{IEEE Journal of Selected Topics in Applied Earth Observations and Remote Sensing}, vol.~14, pp. 9435--9449, 2021.

\bibitem{menon2020pulse}
S.~Menon, A.~Damian, S.~Hu, N.~Ravi, and C.~Rudin, ``Pulse: Self-supervised photo upsampling via latent space exploration of generative models,'' in \emph{Proceedings of the IEEE/CVF Conference on Computer Vision and Pattern Recognition (CVPR)}, 2020, pp. 2437--2445.

\bibitem{chihaoui2025diffusionimageprior}
\BIBentryALTinterwordspacing
H.~Chihaoui and P.~Favaro, ``Diffusion image prior,'' 2025. [Online]. Available: \url{https://arxiv.org/abs/2503.21410}
\BIBentrySTDinterwordspacing

\bibitem{murata2023gibbsddrm}
N.~Murata, K.~Saito, C.-H. Lai, Y.~Takida, T.~Uesaka, Y.~Mitsufuji, and S.~Ermon, ``Gibbsddrm: A partially collapsed gibbs sampler for solving blind inverse problems with denoising diffusion restoration,'' in \emph{International Conference on Machine Learning}.\hskip 1em plus 0.5em minus 0.4em\relax PMLR, 2023, pp. 25\,501--25\,522.

\bibitem{xia2022gan}
W.~Xia, Y.~Zhang, Y.~Yang, J.-H. Xue, B.~Zhou, and M.-H. Yang, ``Gan inversion: A survey,'' \emph{IEEE Transactions on Pattern Analysis and Machine Intelligence}, vol.~45, no.~3, pp. 3121--3138, 2022.

\bibitem{wang2025navigating}
S.~Wang, N.~Zheng, J.~Huang, and F.~Zhao, ``Navigating image restoration with var's distribution alignment prior,'' in \emph{Proceedings of the IEEE/CVF Conference on Computer Vision and Pattern Recognition (CVPR)}, 2025, pp. 7559--7569.

\bibitem{shah2025join}
\BIBentryALTinterwordspacing
V.~Shah, S.~Lazebnik, and J.~Philip, ``Jo{IN}: Joint {GAN}s inversion for intrinsic image decomposition,'' \emph{Transactions on Machine Learning Research}, 2025. [Online]. Available: \url{https://openreview.net/forum?id=JEHIVfjmOf}
\BIBentrySTDinterwordspacing

\bibitem{kingma2013auto}
D.~P. Kingma, M.~Welling \emph{et~al.}, ``Auto-encoding variational bayes,'' 2013.

\bibitem{dinh2014nice}
L.~Dinh, D.~Krueger, and Y.~Bengio, ``Nice: Non-linear independent components estimation,'' \emph{arXiv preprint arXiv:1410.8516}, 2014.

\bibitem{dinh2017density}
\BIBentryALTinterwordspacing
L.~Dinh, J.~Sohl-Dickstein, and S.~Bengio, ``Density estimation using real {NVP},'' in \emph{International Conference on Learning Representations}, 2017. [Online]. Available: \url{https://openreview.net/forum?id=HkpbnH9lx}
\BIBentrySTDinterwordspacing

\bibitem{kingma2018glow}
D.~P. Kingma and P.~Dhariwal, ``Glow: Generative flow with invertible 1x1 convolutions,'' \emph{Advances in Neural Information Processing Systems}, vol.~31, 2018.

\bibitem{radford2015unsupervised}
A.~Radford, L.~Metz, and S.~Chintala, ``Unsupervised representation learning with deep convolutional generative adversarial networks,'' \emph{arXiv preprint arXiv:1511.06434}, 2015.

\bibitem{bredell2023explicitly}
\BIBentryALTinterwordspacing
G.~Bredell, K.~Flouris, K.~Chaitanya, E.~Erdil, and E.~Konukoglu, ``Explicitly minimizing the blur error of variational autoencoders,'' in \emph{The Eleventh International Conference on Learning Representations}, 2023. [Online]. Available: \url{https://openreview.net/forum?id=9krnQ-ue9M}
\BIBentrySTDinterwordspacing

\bibitem{guan2020collaborative}
S.~Guan, Y.~Tai, B.~Ni, F.~Zhu, F.~Huang, and X.~Yang, ``Collaborative learning for faster stylegan embedding,'' \emph{arXiv preprint arXiv:2007.01758}, 2020.

\bibitem{karras2019style}
T.~Karras, S.~Laine, and T.~Aila, ``A style-based generator architecture for generative adversarial networks,'' in \emph{Proceedings of the IEEE/CVF Conference on Computer Vision and Pattern Recognition (CVPR)}, 2019, pp. 4401--4410.

\bibitem{song2021denoising}
\BIBentryALTinterwordspacing
J.~Song, C.~Meng, and S.~Ermon, ``Denoising diffusion implicit models,'' in \emph{International Conference on Learning Representations}, 2021. [Online]. Available: \url{https://openreview.net/forum?id=St1giarCHLP}
\BIBentrySTDinterwordspacing

\bibitem{harmeling2010space}
S.~Harmeling, H.~Michael, and B.~Sch{\"o}lkopf, ``Space-variant single-image blind deconvolution for removing camera shake,'' \emph{Advances in Neural Information Processing Systems}, vol.~23, 2010.

\bibitem{hirsch2010efficient}
M.~Hirsch, S.~Sra, B.~Sch{\"o}lkopf, and S.~Harmeling, ``Efficient filter flow for space-variant multiframe blind deconvolution,'' in \emph{Proceedings of the IEEE/CVF Conference on Computer Vision and Pattern Recognition (CVPR)}.\hskip 1em plus 0.5em minus 0.4em\relax IEEE, 2010, pp. 607--614.

\bibitem{joshi2010image}
N.~Joshi, S.~B. Kang, C.~L. Zitnick, and R.~Szeliski, ``Image deblurring using inertial measurement sensors,'' \emph{ACM Transactions on Graphics (TOG)}, vol.~29, no.~4, pp. 1--9, 2010.

\bibitem{carbajal2023blind}
G.~Carbajal, P.~Vitoria, J.~Lezama, and P.~Mus{\'e}, ``Blind motion deblurring with pixel-wise kernel estimation via kernel prediction networks,'' \emph{IEEE Transactions on Computational Imaging}, vol.~9, pp. 928--943, 2023.

\bibitem{dong2017blind}
J.~Dong, J.~Pan, Z.~Su, and M.-H. Yang, ``Blind image deblurring with outlier handling,'' in \emph{Proceedings of the IEEE/CVF International Conference on Computer Vision (ICCV)}, 2017, pp. 2478--2486.

\bibitem{kaufman2020deblurring}
A.~Kaufman and R.~Fattal, ``Deblurring using analysis-synthesis networks pair,'' in \emph{Proceedings of the IEEE/CVF Conference on Computer Vision and Pattern Recognition (CVPR)}, 2020, pp. 5811--5820.

\bibitem{zhang2024cross}
T.~Zhang, Y.~Quan, and H.~Ji, ``Cross-scale self-supervised blind image deblurring via implicit neural representation,'' in \emph{The Thirty-eighth Annual Conference on Neural Information Processing Systems}, 2024.

\bibitem{he2016deep}
K.~He, X.~Zhang, S.~Ren, and J.~Sun, ``Deep residual learning for image recognition,'' in \emph{Proceedings of the IEEE/CVF Conference on Computer Vision and Pattern Recognition (CVPR)}, 2016, pp. 770--778.

\bibitem{kuznetsova2020open}
A.~Kuznetsova, H.~Rom, N.~Alldrin, J.~Uijlings, I.~Krasin, J.~Pont-Tuset, S.~Kamali, S.~Popov, M.~Malloci, A.~Kolesnikov \emph{et~al.}, ``The open images dataset v4: Unified image classification, object detection, and visual relationship detection at scale,'' \emph{International Journal of Computer Vision}, vol. 128, no.~7, pp. 1956--1981, 2020.

\bibitem{kingma2014adam}
D.~P. Kingma and J.~Ba, ``Adam: A method for stochastic optimization,'' \emph{arXiv preprint arXiv:1412.6980}, 2014.

\bibitem{zhang2018unreasonable}
R.~Zhang, P.~Isola, A.~A. Efros, E.~Shechtman, and O.~Wang, ``The unreasonable effectiveness of deep features as a perceptual metric,'' in \emph{Proceedings of the IEEE/CVF Conference on Computer Vision and Pattern Recognition (CVPR)}, 2018, pp. 586--595.

\bibitem{mittal2012making}
A.~Mittal, R.~Soundararajan, and A.~C. Bovik, ``Making a “completely blind” image quality analyzer,'' \emph{IEEE Signal Processing Letters}, vol.~20, no.~3, pp. 209--212, 2012.

\bibitem{mittal2011blind}
A.~Mittal, A.~K. Moorthy, and A.~C. Bovik, ``Blind/referenceless image spatial quality evaluator,'' in \emph{2011 conference record of the forty fifth asilomar conference on signals, systems and computers (ASILOMAR)}.\hskip 1em plus 0.5em minus 0.4em\relax IEEE, 2011, pp. 723--727.

\bibitem{venkatanath2015blind}
N.~Venkatanath, D.~Praneeth, M.~C. Bh, S.~S. Channappayya, and S.~S. Medasani, ``Blind image quality evaluation using perception based features,'' in \emph{2015 twenty first national conference on communications (NCC)}.\hskip 1em plus 0.5em minus 0.4em\relax IEEE, 2015, pp. 1--6.

\bibitem{lin2014microsoft}
T.-Y. Lin, M.~Maire, S.~Belongie, J.~Hays, P.~Perona, D.~Ramanan, P.~Doll{\'a}r, and C.~L. Zitnick, ``Microsoft coco: Common objects in context,'' in \emph{Proceedings of the European Conference on Computer Vision (ECCV)}.\hskip 1em plus 0.5em minus 0.4em\relax Springer, 2014, pp. 740--755.

\bibitem{choi2020stargan}
Y.~Choi, Y.~Uh, J.~Yoo, and J.-W. Ha, ``Stargan v2: Diverse image synthesis for multiple domains,'' in \emph{Proceedings of the IEEE/CVF Conference on Computer Vision and Pattern Recognition (CVPR)}, 2020, pp. 8188--8197.

\bibitem{liu2018large}
Z.~Liu, P.~Luo, X.~Wang, and X.~Tang, ``Large-scale celebfaces attributes (celeba) dataset,'' \emph{Retrieved August}, vol.~15, no. 2018, p.~11, 2018.

\bibitem{deng2009imagenet}
J.~Deng, W.~Dong, R.~Socher, L.-J. Li, K.~Li, and L.~Fei-Fei, ``Imagenet: A large-scale hierarchical image database,'' in \emph{Proceedings of the IEEE/CVF Conference on Computer Vision and Pattern Recognition (CVPR)}.\hskip 1em plus 0.5em minus 0.4em\relax Ieee, 2009, pp. 248--255.

\bibitem{whyte2012non}
O.~Whyte, J.~Sivic, A.~Zisserman, and J.~Ponce, ``Non-uniform deblurring for shaken images,'' \emph{International Journal of Computer Vision}, vol.~98, pp. 168--186, 2012.

\bibitem{zhang2023event}
H.~Zhang, L.~Zhang, Y.~Dai, H.~Li, and P.~Koniusz, ``Event-guided multi-patch network with self-supervision for non-uniform motion deblurring,'' \emph{International Journal of Computer Vision}, vol. 131, no.~2, pp. 453--470, 2023.

\bibitem{fang2023self}
Z.~Fang, F.~Wu, W.~Dong, X.~Li, J.~Wu, and G.~Shi, ``Self-supervised non-uniform kernel estimation with flow-based motion prior for blind image deblurring,'' in \emph{Proceedings of the IEEE/CVF Conference on Computer Vision and Pattern Recognition (CVPR)}, 2023, pp. 18\,105--18\,114.

\bibitem{vasu2017local}
S.~Vasu and A.~Rajagopalan, ``From local to global: Edge profiles to camera motion in blurred images,'' in \emph{Proceedings of the IEEE/CVF Conference on Computer Vision and Pattern Recognition (CVPR)}, 2017, pp. 4447--4456.

\end{thebibliography}
